\documentclass[10pt,twocolumn,letterpaper]{article}

\usepackage{cvpr}              % To produce the CAMERA-READY version
\definecolor{cvprblue}{rgb}{0.21,0.49,0.74}
\usepackage[pagebackref,breaklinks,colorlinks,allcolors=cvprblue]{hyperref}

\def\paperID{6022} % *** Enter the Paper ID here
\def\confName{CVPR}
\def\confYear{2026}

\title{LD-ViCE: Latent Diffusion Model for Video Counterfactual Explanations}

\author{
Payal Varshney\textsuperscript{1,2}\thanks{Corresponding author. Email: payal.varshney@dfki.de} \quad
Adriano Lucieri\textsuperscript{1,2} \quad
Christoph Balada\textsuperscript{1,2} \quad
Sheraz Ahmed\textsuperscript{2} \quad
Andreas Dengel\textsuperscript{1,2} \\
\textsuperscript{1}Rheinland-Pfälzische Technische Universität Kaiserslautern-Landau, Kaiserslautern, Germany \\
\textsuperscript{2}German Research Center for Artificial Intelligence (DFKI), Kaiserslautern, Germany \\
{\tt\small firstname.lastname@dfki.de}
}

\newcommand{\frameworkName}{Latent Diffusion for Video Counterfactual Explanations}
\newcommand{\frameworkAcronym}{LD-ViCE}

\begin{document}

\twocolumn[{%
\maketitle
\centering
\includegraphics[width=\linewidth]{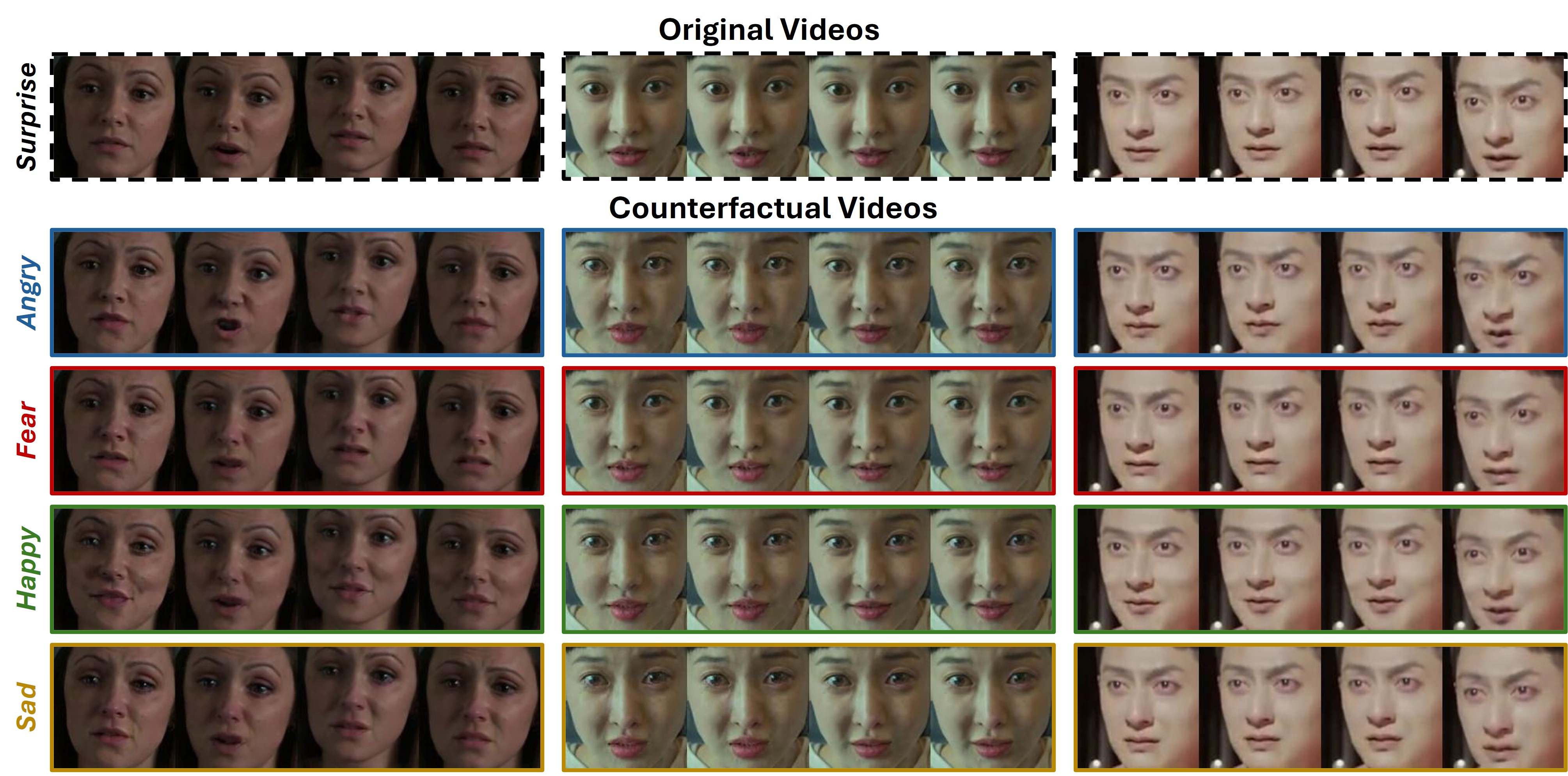}
\captionof{figure}{Qualitative counterfactual results generated by \frameworkAcronym\ on the FERV39k dataset. The first row shows four frames from three original videos predicted as \textit{Surprise}. The second, third, fourth, and fifth rows display counterfactuals generated for the target emotion classes \textit{Angry}, \textit{Fear}, \textit{Happy}, and \textit{Sad}, respectively. The generated counterfactuals exhibit distinct and class-consistent facial dynamics corresponding to the desired emotional categories. }
\label{fig:ferv39k}
\vspace{1em} % optional: small breathing room
}]
\begin{abstract}
Video-based AI systems are increasingly adopted in safety-critical domains such as autonomous driving and healthcare.
However, interpreting their decisions remains challenging due to the inherent spatiotemporal complexity of video data and the opacity of deep learning models.
Existing explanation techniques often suffer from limited temporal coherence and a lack of actionable causal insights.
Current counterfactual explanation methods typically do not incorporate guidance from the target model, reducing semantic fidelity and practical utility.
We introduce \frameworkName~(\frameworkAcronym), a novel framework designed to explain the behavior of video-based AI models.
Compared to previous approaches, \frameworkAcronym\ reduces the computational costs of generating explanations by operating in latent space using a state-of-the-art diffusion model, while producing realistic and interpretable counterfactuals through an additional refinement step.
Experiments on three diverse video datasets—EchoNet-Dynamic (cardiac ultrasound), FERV39k (facial expression), and Something-Something V2 (action recognition) with multiple target models covering both classification and regression tasks, demonstrate that \frameworkAcronym\ generalizes well and achieves state-of-the-art performance.
On the EchoNet-Dynamic dataset, \frameworkAcronym\ achieves significantly higher regression accuracy than prior methods and exhibits high temporal consistency, while the refinement stage further improves perceptual quality.
Qualitative analyses confirm that \frameworkAcronym\ produces semantically meaningful and temporally coherent explanations, providing actionable insights into model behavior.
\frameworkAcronym\ advances the trustworthiness and interpretability of video-based AI systems through visually coherent counterfactual explanations.
\end{abstract}    
\section{Introduction}\label{sec:introduction}

Video-based artificial intelligence (AI) systems are rapidly gaining prominence across high-stakes domains such as autonomous driving~\cite{yurtsever2020survey}, urban surveillance~\cite{ullah2023comprehensive}, and clinical video diagnostics~\cite{farhad2023review}.
This growing adoption is supported by recent advances in large-scale spatiotemporal foundation models, which are capable of processing minutes-long clips to produce temporally coherent predictions~\cite{tang2025video}.
Despite these advancements, video understanding remains significantly more challenging than static image recognition due to the intricate temporal dynamics, long-range dependencies, and causal interactions across frames.
For instance, detecting suspicious behavior in surveillance footage requires interpreting motion and intent over time, not just spatial patterns within individual frames~\cite{arunnehru2023deep}.
This increased complexity means that errors in video-based predictions are difficult to diagnose and have far-reaching consequences in safety-critical settings.
Misclassifications in medical ultrasound videos or missed anomalies in autonomous driving scenarios can lead to severe consequences. 
While explainability may not always prevent such failures, it is crucial for accountability and post-hoc analyses, offering insights into why specific decisions were made.
Thus, explainable AI (XAI) is essential for fostering trust, regulatory compliance, and supporting effective human oversight~\cite{adadi2018peeking}. 
Despite these needs, explainability in the video domain remains substantially underexplored relative to the image domain. 

Most existing explainability methods for video models adapt image-based techniques to the spatiotemporal domain.
Feature attribution methods highlight regions influencing predictions~\cite{stergiou2019saliency, li2021towards} but typically produce pixel‑level saliency maps that lack semantic depth, temporal coherence, and robustness to noise~\cite{adebayo2018sanity, arun2021assessing}.
Concept-based methods link predictions to human-interpretable concepts~\cite{ji2023spatial, saha2024exploring}, yet often rely on static clustering or predefined concept sets, which limits their ability to capture dynamic motion patterns critical for video understanding.
While feature attribution reveals where a model attends and concept‑based methods indicate what it focuses on, neither clarifies how the input must change to alter the prediction, making them non‑actionable.
For example, a saliency map may highlight an object, and a concept‑based explanation may confirm its relevance.
However, neither specifies whether the object should be added, removed, resized, or recolored to change the outcome.
In contrast, counterfactual explanations (CEs)~\cite{wachter2017counterfactual} explicitly identify the minimal, targeted changes required to flip a model’s decision, offering actionable and causal insights essential for video analysis.

Recently, generative approaches have been explored for producing video CEs.
Existing methods~\cite{reynaud2023feature, zong2025text} utilize diffusion models guided by textual prompts, without incorporating feedback from the target video model during generation. 
Instead, they follow a model‑agnostic strategy, using the target model only for post‑hoc evaluation rather than steering the generation process.
Consequently, although these approaches can produce visually plausible CEs, they provide limited insight into the model’s decision-making process and may even yield unreliable or misleading explanations.
Moreover, both approaches operate entirely in pixel space, resulting in high computational costs and poor scalability for longer sequences or higher-resolution videos.

To overcome these limitations, a novel method, \frameworkName~(\frameworkAcronym) is proposed, which generates counterfactual video explanations using a latent diffusion model explicitly guided by the target model.
By operating in latent space, \frameworkAcronym\ significantly reduces computational cost while maintaining temporal coherence and visual fidelity.
\frameworkAcronym\ integrates model feedback throughout generation, ensuring causal alignment with the model’s predictions.
The main contributions of this paper are summarized as follows:
\begin{itemize}
    \item This work introduces \frameworkAcronym, a latent diffusion–based formulation for counterfactual video explanation that integrates explicit target model guidance, filling a gap left by existing text-guided approaches.
    \item A denoising refinement step is proposed within \frameworkAcronym\ to mitigate residual noise introduced by the diffusion denoising process, enhancing visual realism and stability.
    \item The effectiveness of \frameworkAcronym\ is demonstrated across diverse publicly available datasets, including EchoNet-Dynamic (cardiac ultrasound), FERV39k (facial expression recognition), and Something-Something V2 (action recognition), and multiple target models covering both classification and regression tasks.
  \end{itemize}

%%%%%%%%%%%%%%%%%%%%%%%%%
%%% Related Work      %%%
%%%%%%%%%%%%%%%%%%%%%%%%%

\section{Related Work}\label{sec:relatedwork}

Understanding the decision-making process of video models is essential due to their complex spatiotemporal architectures and increasing deployment in high-stakes applications.
Explainability in the video domain remains far less developed than in the image domain, where post‑hoc interpretability methods have advanced rapidly. 
Existing methods for explaining video-based models adapt techniques initially designed for 2D image classification, which limits their effectiveness for models that rely on rich temporal dynamics and higher-dimensional inputs~\cite{hiley2020explaining}. 
\citet{kolarik2023explainability} provide a comprehensive survey of interpretability techniques in medical video analysis, underscoring their clinical relevance while highlighting critical gaps: the absence of standardized evaluation protocols and the limited temporal reasoning capabilities of current methods. 
This section discusses existing counterfactual video explanation methods and their key limitations.

Early works in counterfactual video explanation have focused on selecting existing evidence that flips a model’s prediction. 
~\citet{kanehira2019multimodal} present one of the earliest frameworks for generating multimodal explanations, identifying spatiotemporal tubes and corresponding attributes whose presence would alter the classifier’s decision.
While insightful, the method is limited to masking or retrieving existing regions and cannot synthesize novel visual content, restricting its expressiveness in generating counterfactuals. 

To move beyond region selection, generative methods have been explored to produce counterfactual video samples.
\citet{reynaud2022d} leverage a conditional GAN to synthesize cardiac ultrasound videos conditioned on user-specified clinical factors, enabling actionable visual counterfactuals for model interpretation.
However, the method inherits common GAN limitations, such as training instability, mode collapse, and reduced temporal consistency for longer sequences.
More recently, diffusion models have emerged as a stable, higher-fidelity alternative for generating counterfactual videos.
\citet{reynaud2023feature} propose a cascaded diffusion framework that generates high‑resolution echocardiogram videos conditioned on clinical attributes and a randomly selected reference frame. 
This method processes only a single reference frame, rather than the entire sequence, resulting in poor temporal coherence, and it does not incorporate feedback from the target model.
Similarly, \citet{zong2025text} introduce a multimodal debiasing pipeline using a text-guided diffusion model to generate fine-grained counterfactuals for video-based fake news detection.
Although semantic control is achieved through textual prompts, the approach lacks conditioning on the target classifier, reducing its explanatory power.
In both cases, the target model is used only post‑hoc for evaluation rather than guiding the counterfactual generation.
Consequently, the resulting explanations may not capture the model’s actual decision process and, in some cases, may lead to misleading or spurious interpretations.
Furthermore, both approaches operate in pixel space, incurring high computational cost and limiting scalability to longer or higher‑resolution videos.

While classifier guidance has been explored in image‑based counterfactual generation~\cite{jeanneret2022diffusion, augustin2022diffusion, varshney2024generating}, it remains largely absent from video explanation methods.
Notably, to the best of our knowledge, no prior work combines classifier-guided diffusion models with counterfactual generation for video explanations.
To address this research gap, \frameworkAcronym\ is introduced, a framework for generating counterfactual video explanations via a latent diffusion process explicitly guided by the target model.
By incorporating classifier feedback throughout generation, \frameworkAcronym\ ensures alignment with the model’s decision boundaries, while operating entirely in the latent space preserves spatiotemporal coherence and significantly reduces computational cost compared to pixel‑space approaches.
This design enables scalable, causally valid, and temporally coherent counterfactual explanations for videos. %higher‑resolution
%%%%%%%%%%%%%%%%%%%%%%%%%
%%% Methodology     %%%
%%%%%%%%%%%%%%%%%%%%%%%%%

\begin{figure*}[t]
\centering
\includegraphics[width=\linewidth]{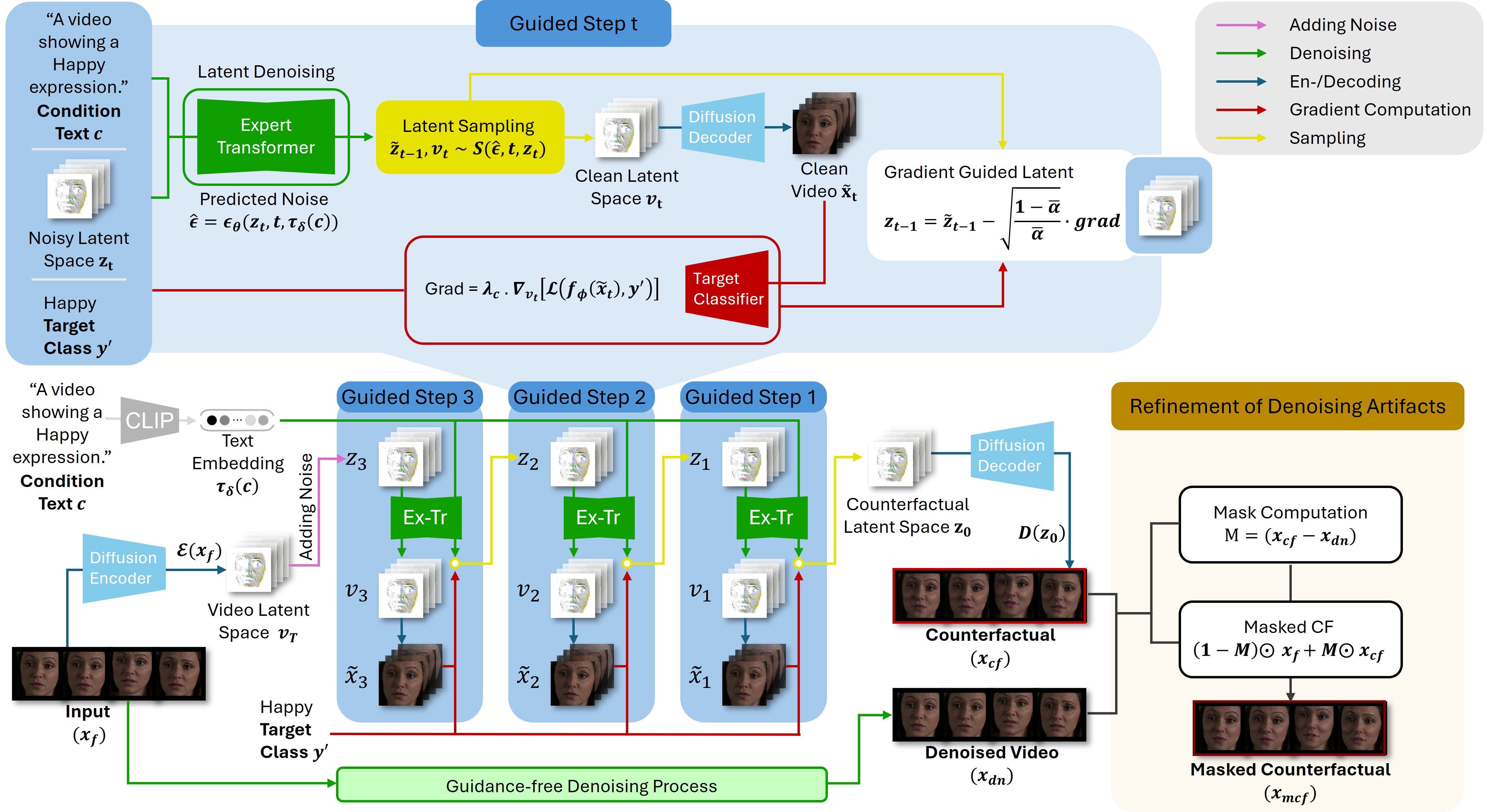}

\caption{
Overview of the \frameworkAcronym\ counterfactual generation process.
The factual video $x_f$ is encoded and perturbed to obtain the noisy latent $z_T$ (here, $T=3$), while the conditional text prompt $c$ is embedded via the text encoder $\tau_\delta(c)$. At each guided denoising step $t$, the latent $z_t$ and embedding $\tau_\delta(c)$ are provided to the diffusion model.
The denoising model (e.g., Expert Transformer (Ex-Tr)) predicts the noise $\hat{\epsilon}$, which is used in the sampling process to compute the clean latent $v_t$ and the less noisy latent $\tilde{z}_{t-1}$. 
The clean latent $v_t$ is decoded to produce $\tilde{x}_t$, which is used to estimate classifier gradients, scaled by $\lambda_c$, to compute the updated latent $z_{t-1}$.
After the final step, $z_0$ is decoded into the counterfactual video $x_{cf}$.
A refinement stage then denoises the same latent $z_T$ without guidance to obtain a clean reference video, from which a mask is computed to suppress diffusion artifacts and produce the final masked counterfactual video $x_{mcf}$.
}
  
\label{fig:ld-vice}
\end{figure*}

\section{Methodology} \label{sec:methodology}

We propose \frameworkName~(\frameworkAcronym), a novel framework for generating counterfactual video explanations by integrating classifier guidance into a text-to-video latent diffusion process.
Although conceptually related to a latent diffusion–based counterfactual image generation method~\cite{varshney2024generating}, \frameworkAcronym~extends counterfactual reasoning to the spatiotemporal domain, where both spatial realism and temporal coherence must be preserved.
\frameworkAcronym~leverages a video diffusion model to implicitly maintain temporal consistency across frames, while gradient-based conditioning from the target classifier steers the generation towards the desired class.
To further enhance the visual fidelity of the explanations, a novel masking‑based post‑processing technique is introduced that selectively preserves unchanged regions of the original video while integrating only the classifier‑relevant modifications from the generated counterfactual.
\cref{fig:ld-vice} provides an overview of the counterfactual generation pipeline.

\subsection{Problem Formulation}
Let \(f_{\phi}: \mathcal{X} \rightarrow \mathcal{Y}\) denote a pretrained target model to be explained that maps a video \(x \in \mathcal{X}\) to a class label or regression value \(y \in \mathcal{Y}\). 
Given a factual video  \(x_{\mathrm{f}} \in \mathbb{R}^{F \times H \times W \times C}\) with $F$ frames and model prediction \(y = f_{\phi}(x_{\mathrm{f}})\), the objective is to generate a counterfactual video \(x_{\mathrm{cf}} \in \mathbb{R}^{F \times H \times W \times C}\) with target \(y'\) that is visually plausible and semantically minimal, satisfying

\begin{equation}  
    f_{\phi}(x_{\mathrm{cf}}) = y', \quad
    \text{where } y' \in \mathcal{Y},\; y' \neq y.
     \label{eq:problem_formulation}
\end{equation}

Among all plausible alternatives, \(x_{\mathrm{cf}}\) should capture the minimal set of changes needed to alter the model’s decision, while preserving spatiotemporal coherence.
To achieve this, the generation process is guided along trajectories aligned with the decision boundary of \(f_{\phi}\), using classifier gradients to ensure that modifications are causally relevant. 

\subsection{Latent Diffusion-Based Video Counterfactual Generation}
\frameworkAcronym\ generates counterfactual explanations for video models, ranging from lightweight networks to transformer-based architectures, and is also task-agnostic, generalizing effectively across various tasks, including regression and classification.
In contrast to prior work~\cite{reynaud2023feature}, \frameworkAcronym\ leverages a text-to-video latent diffusion model (LDM), whose operation in a compressed latent space enables significantly lower computational cost, faster sampling, and reduced memory usage, while still generating high-quality, temporally coherent counterfactual videos.

\subsubsection{Forward Diffusion Process}

The factual video \( x_{\text{f}} \) is transformed to a spatiotemporal latent representation by the encoder \(\mathcal{E} \) of the 3D causal VAE~\cite{yang2024cogvideox}. 
This encoder captures both spatial and temporal dependencies, producing a compact latent space
\begin{equation}  
    \mathcal{E}(x_{\text{f}} ) \in \mathbb{R}^{F'\times H'\times W'\times C'}
     \label{eq:encoding}
\end{equation}
where $F' < F$ denotes the temporally downsampled sequence length, $H'< H$  and  $W'< W$ indicate the compressed spatial dimensions, and $C'< C$  represents the number of latent feature channels retained after encoding.
This latent preserves spatial detail and motion consistency while enabling efficient processing.

The encoded video is then perturbed using the forward diffusion process defined by a variance schedule \( \beta_t \in (0, 1) \), with the cumulative product:
\begin{equation} 
    \bar{\alpha}_t = \prod_{k=1}^{t}(1 - \beta_k), \quad 1 \leq t \leq T
     \label{eq:cumulative_product}
\end{equation}
The noisy latent at step \( T \) is obtained as:
\begin{equation} 
    z_T = \sqrt{\bar{\alpha}_T} \cdot \mathcal{E}(x_{\text{f}} ) + \sqrt{1 - \bar{\alpha}_T} \cdot \epsilon_T,\quad \epsilon_T \sim \mathcal{N}(0, I)
    \label{eq:noisy_latent}
\end{equation}
Simultaneously, a text prompt \( c \), incorporating the counterfactual target label  \( y' \)  in textual form and relevant metadata, is encoded via the text encoder \( \tau_\delta(c) \).

\subsubsection{Reverse Diffusion Process with Causal Guidance}

The reverse denoising process \( z_T \rightarrow z_0 \) is guided by the gradients of the target downsampling task.
The following steps are performed at each denoising step \( t = T, \ldots, 1 \):

\begin{itemize}
    \item The diffusion denoising model \( \epsilon_\theta \) predicts the noise component \( \hat{\epsilon}_t \):
        \begin{equation} 
            \hat{\epsilon}_t = \epsilon_\theta(z_t, \tau_\delta(c), t)
            \label{eq:noisy_predict}
        \end{equation}

    \item DDIM sampling~\cite{song2020denoising} \(\mathcal{S}\) estimates the previous noisy latent \(\tilde{z}_{t-1}\) and the clean latent \(v_t\) at step $t$:
        \begin{equation} 
            \tilde{z}_{t-1}, v_t = \mathcal{S}(z_t, \hat{\epsilon}_t, t)
            \label{eq:sampling}
        \end{equation}
    
    \item The estimated clean latent \( {v}_t \) is decoded into the video:
        \begin{equation} 
            \tilde{x}_t = D({v}_t)
            \label{eq:decoding_latent}
        \end{equation}
    \item The target model \( f_\phi \) is applied to compute the task specific loss \(\mathcal{L}\) with respect to the target value \( y' \) and the gradient is scaled by \(\lambda_c\) as  :
        \begin{equation} 
            \nabla_{v_t} \mathcal{L}_{\text{CE}} = \lambda_c \cdot \nabla_{v_t} \left[ \mathcal{L}\left( f_\phi(\tilde{x}_t), y' \right) \right]
            \label{eq:ce_loss}
        \end{equation}
    \item The noisy latent is updated using the gradient:
        \begin{equation} 
            z_{t-1} = \tilde{z}_{t-1} - \sqrt{\frac{1 -\bar{\alpha}_t}{\bar{\alpha}_t}}
                 \cdot \nabla_{v_t} \mathcal{L}_{\text{CE}}
            \label{eq:noisy_latent_update}
        \end{equation}

\end{itemize}
The updated latent \(z_{t-1} \) is input for the next reverse step. After \( T \) reverse steps, \( z_0 \) is obtained, which is decoded by the decoder \( D \) of the 3D causal VAE to produce the counterfactual video:
\begin{equation} 
    x_{\text{cf}} = D(z_0)
    \label{eq:decoding}
\end{equation}
Similar to the previous method by ~\citet{farid2023latent}, \frameworkAcronym\ leverages DDIM sampling~\cite{song2020denoising} that analytically computes and returns the clean latent representation at every diffusion timestep. The computational complexity is \( \mathcal{O}(T) \), enabling efficient inference while preserving high‑quality counterfactual generation.

\subsection{Refinement of Denoising Artifacts}
The diffusion denoising process often introduces minor visual artifacts that degrade counterfactual fidelity.
To isolate causally relevant changes from these artifacts, the same latent \( z_T \) is denoised without guidance to produce a clean reference \( x_{\mathrm{den}} \).
Although diffusion sampling is inherently stochastic, using a fixed random seed and deterministic scheduler ensures that 
\( x_{\mathrm{den}} \) remains structurally aligned with the guided counterfactual \(x_{\mathrm{cf}} \).

A binary refinement mask \( M \in \{0,1\}^{F \times H \times W} \) is computed by comparing \( x_{\mathrm{cf}} \) with \( x_{\mathrm{den}} \) to suppress artifacts from the diffusion process while retaining semantically meaningful changes:
\begin{equation} 
    \Delta(f, h, w) = \sum_{c=1}^{3} \left| x_{\mathrm{cf}}(f, h, w, c) - x_{\mathrm{den}}(f, h, w, c) \right|
    \label{eq:difference_map}
\end{equation}
\begin{equation} 
    M(f, h, w) = \mathbb{I} \left[ \Delta(f, h, w) > t_{sup} \right]
    \label{eq:mask}
\end{equation}
where \( \mathbb{I}[\cdot] \) denotes the indicator function, and $t_{sup}$ is an empirically chosen threshold, controlling artifact suppression.
The resulting mask \( M \) is broadcast along the channel dimension to construct the refined counterfactual video:
\begin{equation} 
    x_{\mathrm{mask\_cf}} \;=\;
    (1-M)\odot x_{\mathrm{f}} \;+\; M\odot x_{\mathrm{cf}}.
    \label{eq:masked_cf}
\end{equation}
where \( \odot \) denotes element-wise multiplication. This formulation replaces only those voxels marked by \( {M} \), thereby preserving the original content elsewhere and effectively mitigating diffusion-related artifacts.
%%%%%%%%%%%%%%%%%%%%%%%%%%%%%%%%%%%%%%%%%%
%%% Experiments \& Results %%%%%%%%%%%%%%%
%%%%%%%%%%%%%%%%%%%%%%%%%%%%%%%%%%%%%%%%%%

\section{Experiments \& Results}

% This section describes the experimental setup and presents a comprehensive evaluation of the
% % proposed
% \frameworkAcronym{} framework.

\subsection{Datasets and Target Models}
\frameworkAcronym\ is evaluated on three benchmark datasets spanning diverse video domains.
A state-of-the-art network is used as a black-box target model for each dataset, without modifying its parameters.

\textbf{FERV39K} A large-scale facial expression recognition dataset~\cite{wang2022ferv39k} comprising videos annotated with seven emotion categories: Angry, Disgust, Fear, Happy, Neutral, Sad, and Surprise (multiclass classification).
The publicly available Static-to-Dynamic (S2D) model~\cite{chen2024static} is utilized as the black-box target.
This architecture extends a Vision Transformer by incorporating Temporal Modeling Adapters to adapt image features dynamically across time.

\textbf{EchoNet‐Dynamics} An echocardiography video dataset~\cite{ouyang2020video} annotated with continuous left‐ventricular ejection‐fraction (LVEF) values (regression task).
The 3D ResNet variant leveraged by ~\citet{reynaud2023feature} is trained using the provided configuration and adopted as the black-box regressor, since pretrained weights are not publicly available.

\textbf{Something‐Something~V2 (SSv2)} A fine‐grained human‐action dataset~\cite{goyal2017something} with 174 classes (multiclass classification).
The publicly released VideoMAE-Base model~\cite{tong2022videomae} served as black‐box classifier.
It is a spatiotemporal Vision Transformer pre-trained by masked autoencoding and equipped with factorised space–time attention.

\subsection{Fine-tuning of Video Diffusion Model}

\frameworkAcronym~is compatible with any text-to-video latent diffusion model for counterfactual generation. 
In all experiments, CogVideoX~\cite{yang2024cogvideox} serves as the diffusion backbone.
The denoising Expert Transformer is fine-tuned\footnote{Fine-tuning code adapted from \url{https://github.com/huggingface/diffusers/tree/main/examples/cogvideo}.}  independently for each dataset, whereas the 3D causal VAE and text encoder remain frozen and are used solely to obtain latent video representations and text embeddings for conditioning.
Frame sampling and spatial resolution are aligned with the input requirements of the corresponding target model, with 16 frames extracted per video at a sampling rate of 8 frames per second.
Training text prompts are generated for each video by incorporating ground-truth class labels or regression values with available metadata, thereby enriching the conditioning signal during fine-tuning.
The diffusion backbone is fine-tuned for \(50\) epochs with a learning rate of \(1e-3\), enabling the generation of realistic and domain-consistent samples.
Detailed information about the caption generation process is provided in \cref{sec:text-prompt}.

\begin{figure*}[htb]
\centering
\includegraphics[width=\linewidth]{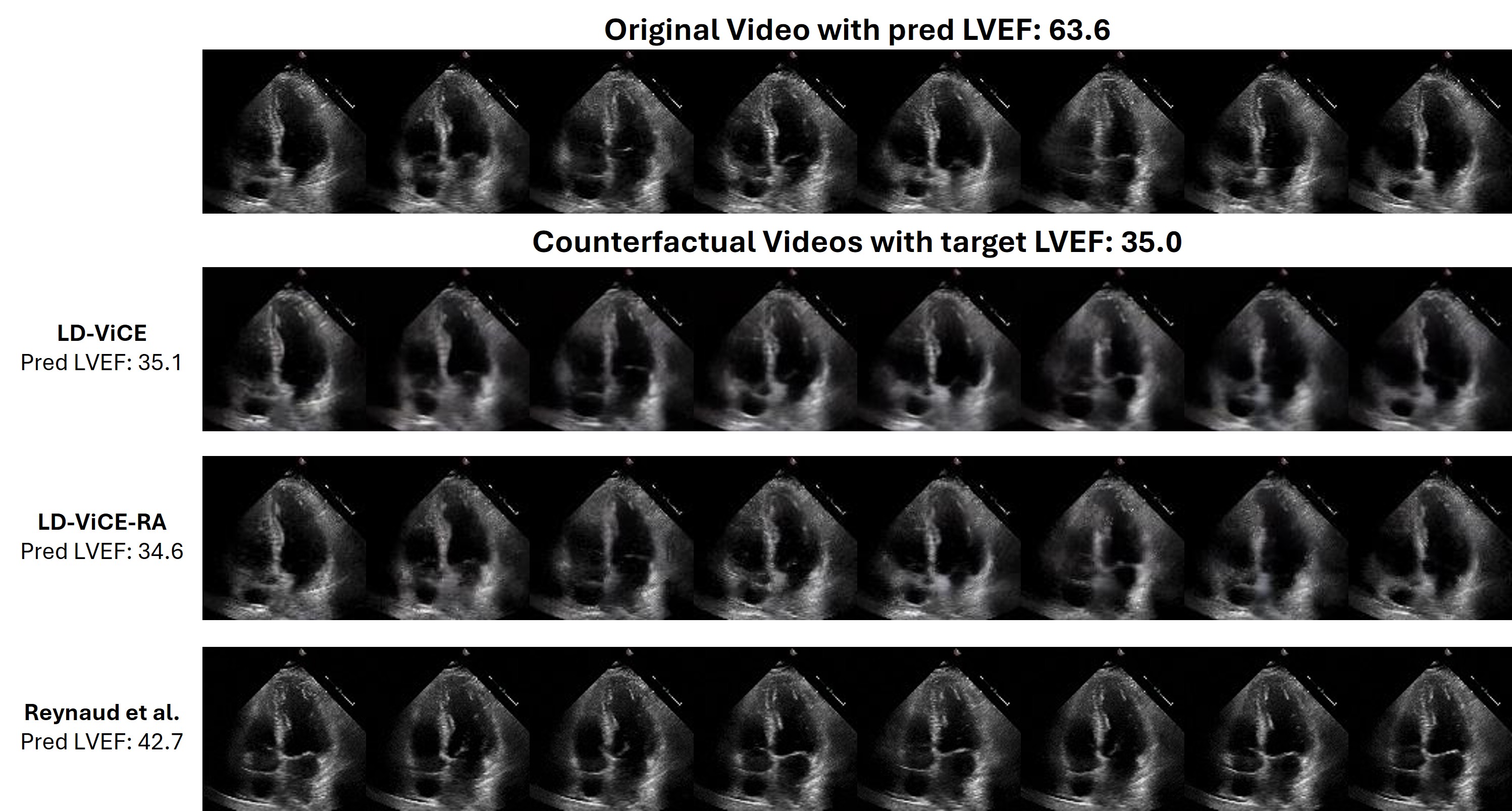}
\caption{
Qualitative comparison of counterfactual explanations on the EchoNet-Dynamics dataset. 
The first row shows eight frames from the original video, while the subsequent rows present counterfactuals generated using \frameworkAcronym{}, \frameworkAcronym{}-RA, and 1SCM~\cite{reynaud2023feature}, respectively. 
Predicted LVEF values are shown on the left. 
The figure illustrates that \frameworkAcronym{} produces visually coherent counterfactuals that more closely match the target regression values compared to prior methods.
}
\label{fig:echonet}
\end{figure*}

\subsection{Video Counterfactual Generation}

The proposed \frameworkAcronym{} framework generates counterfactual videos across all datasets.
Two variants of the framework are considered for evaluation: (i) Gradient Guidance, and (ii) Gradient Guidance with Refinement of Denoising Artifacts (RA).
Counterfactual generation in \frameworkAcronym{} is controlled by three key hyperparameters: the gradient loss scale \((\lambda_c)\), the number of denoising inference steps \((T)\), and the artifact suppression threshold \((t_{sup})\). 
Ablation studies are first conducted using the Gradient Guidance variant to determine optimal values for \(\lambda_c\) and \(T\). 
Based on these selected values, the optimal threshold \((t_{sup})\) for artifact suppression is determined for the RA variant. 
Comprehensive ablation results and parameter analyses are provided in \cref{sec:ablation_Study}.
All experiments are repeated $5$ times, and the mean and standard deviation are reported.
All quantitative results are reported rounded to two decimal places.
% All experiments were repeated $5$ times with mean and standard deviation reported, and all metrics are rounded to $2$ decimal places.

\begin{table*}[htb]
    \centering
    \small
    \caption{Quantitative comparison of regression accuracy, visual quality, temporal consistency (Temp.), and inference time for counterfactuals generated by \frameworkAcronym{}, its variants, and baseline models~\cite{reynaud2022d, reynaud2023feature}, evaluated on EchoNet-Dynamic validation split (1,288 videos). Best values are highlighted in bold. All variants of \frameworkAcronym{} outperform prior methods in regression performance (R², MAE, RMSE) and in perceptual similarity (LPIPS), while the RA variant further enhances visual quality by achieving the highest SSIM and lowest LPIPS.}
    \begin{tabular}{lcccccccc}
        \toprule
        \textbf{Method} & \textbf{Time~$\downarrow$} & \textbf{$\mathbf{R^2}$~$\uparrow$} & \textbf{MAE~$\downarrow$} & \textbf{RMSE~$\downarrow$} & \textbf{SSIM~$\uparrow$} & \textbf{LPIPS~$\downarrow$} & \textbf{Temp.~$\uparrow$}\\
        \midrule
        Dartagnan~\cite{reynaud2022d} & \textbf{1} s & 0.51 & 15.7 & 18.4 & 0.79 & - & -\\
        \midrule
        1SCM~\cite{reynaud2023feature} &62 s & 0.64 & 9.65 & 12.2 & 0.53 & 0.21 & -\\
        2SCM\cite{reynaud2023feature}  & 146 s & 0.89 & 4.81 & 6.69 & 0.53 & 0.24   & - \\
        4SCM~\cite{reynaud2023feature} & 279 s  & 0.93 & 3.77 & 5.26 & 0.48 & 0.25  & -\\
        \midrule
        \frameworkAcronym{} (ours) & 7 s &$\mathbf{1.00\pm.00}$ & $0.29\pm.01$  & $0.55\pm.06$ &  $0.75\pm.00$    &  $0.18\pm.00$  & $\mathbf{0.95\pm.00}$ \\
        \frameworkAcronym{}-{wo\_ft} (ours) & 7 s &$\mathbf{1.00\pm.00}$ & $\mathbf{0.28\pm.01}$  & $\mathbf{0.51\pm.05}$ &  $0.75\pm.00$  &  $0.18\pm.00$ & $0.94\pm.00$ \\
        \frameworkAcronym{}-{un\_ft} (ours) & 7 s &$\mathbf{1.00\pm.00}$ & $0.29\pm.01$  & $0.52\pm.04$ &  $0.75\pm.00$  &  $0.18\pm.00$ & $\mathbf{0.95\pm.00}$ \\
        \frameworkAcronym{}-{un\_wo\_ft} (ours)& 7 s &$\mathbf{1.00\pm.00}$ & $0.29\pm.02$  & $0.53\pm.06$ &  $0.75\pm.00$  &  $0.18\pm.00$ & $0.94\pm.00$ \\
        \frameworkAcronym{}-RA (ours) & 8 s & $0.86\pm.00$ & $3.07\pm.05$  & $4.14\pm.06$ &  $\mathbf{0.89\pm.00}$    &   $\mathbf{0.09\pm.00}$ & $0.94\pm.00$ \\
        \bottomrule
    \end{tabular}%
    %}
    \label{tab:echonet_results}
\end{table*}

\subsubsection{Regression Task}
In regression settings, counterfactual generation in \frameworkAcronym{} is guided by the mean squared error (MSE) loss.
Target LVEF values are defined following a clinical rule: predictions with $\text{LVEF} \geq 50\%$ (normal function) are assigned a counterfactual target of $35\%$, indicating reduced ejection fraction, while those with $\text{LVEF} < 50\%$ (impaired function) are assigned $60\%$, reflecting normal performance. %restored normal performance
% This setup enforces counterfactuals that reflect realistic transitions between pathological and healthy cardiac states.

\cref{fig:echonet} compares counterfactuals generated by \frameworkAcronym{} and its RA variant with the baseline~\cite{reynaud2023feature} on the EchoNet-Dynamics dataset for LVEF prediction.
The baseline~\cite{reynaud2023feature} predicts an LVEF of 42.7, well above the target, and produces frames that differ from the original video but remain nearly identical to each other, likely due to its reliance on a single frame and target text without model feedback.
In contrast, \frameworkAcronym{} processes the entire video sequence with gradient guidance, generating temporally coherent CEs that closely align with the target value.
Both variants produce realistic videos that remain visually close to the original while reflecting the intended low-functioning cardiac state.

\cref{tab:echonet_results} reports regression accuracy ($R^2$, MAE, RMSE), structural similarity (SSIM), perceptual quality (LPIPS),  temporal consistency, and sampling time (measured for generating a counterfactual on an NVIDIA RTX~A6000 GPU). 
Temporal consistency is computed by averaging the cosine similarity between the CLIP~\cite{radford2021learning} image embeddings of adjacent video frames.
Details of the evaluation metrics are provided in \cref{sec:evaluation}.
The baseline results for~\citet{reynaud2022d, reynaud2023feature} are taken from their respective papers.
The subscripts \textit{ft} and \textit{wo-ft} denote fine-tuned and pretrained diffusion models, while \textit{un} indicates unconditional counterfactual generation without text conditioning.
Even without fine-tuning or textual prompts, \frameworkAcronym{} maintains strong performance, indicating that the method can operate effectively with any pretrained text-to-video diffusion model and gradient guidance alone suffices for generating meaningful counterfactuals.
Across all configurations, \frameworkAcronym{} achieves near-perfect regression accuracy ($R^2=1.00$) with the lowest MAE and RMSE, high temporal coherence ($\text{Temp.}\approx0.95$), and significantly faster sampling than~\citet{reynaud2023feature}.
FID and FVD values (\cref{sec:ab_echonet}) are not directly comparable due to CogVideoX’s 3D VAE not being fine-tuned on the medical domain, which increases FVD.
% FID and FVD values (\cref{sec:ab_echonet}) are not directly comparable across methods. The quality of the generated counterfactuals depends on the reconstruction of the 3D causal VAE, which is not fine-tuned on the medical domain and removes fine-grained details, leading to increased FVD scores. This reflects a limitation of the pretrained VAE rather than the counterfactuals themselves, and FVD should therefore not be interpreted in the same way as for pixel-space approaches.
The RA variant yields the most realistic counterfactuals, achieving the highest SSIM (0.89) and lowest LPIPS (0.09), with a slight reduction in regression accuracy. 
These results confirm that \frameworkAcronym{} generates visually coherent, semantically aligned counterfactuals that achieve regression targets while improving both efficiency and realism.
% These results confirm that \frameworkAcronym{} generates visually coherent and semantically aligned counterfactuals that achieve the intended regression targets, advancing the state of the art in counterfactual video explanations for medical imaging.

\subsubsection{Classification Task}

\begin{figure*}
\centering
\includegraphics[width=\linewidth]{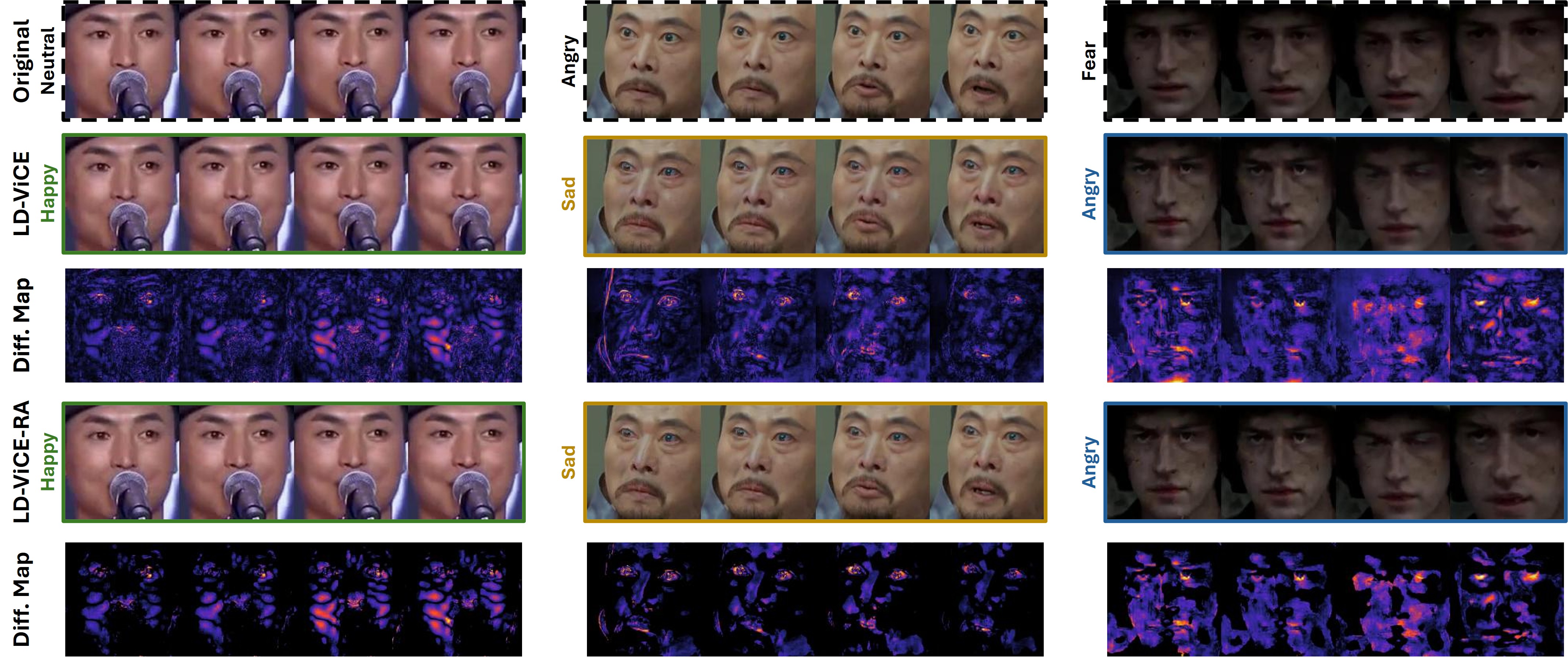}
\caption{
Qualitative comparison of counterfactuals generated by \frameworkAcronym{} and its RA variant on the FERV39K dataset. The figure shows three representative samples, each consisting of four frames from the original video (top row) and the corresponding counterfactuals generated by \frameworkAcronym{} and \frameworkAcronym{}-RA, with their corresponding difference maps. Original and target emotion classes are indicated on the left side of each example. The \frameworkAcronym{}-RA variant focuses more precisely on expression-relevant facial regions while maintaining realistic appearance and temporal coherence.
}
\label{fig:ferv39k_RA}
\end{figure*}

In classification settings, counterfactual generation in \frameworkAcronym{} is guided by the cross-entropy loss, and the target class is randomly selected from the set of non-predicted classes.
Counterfactual results for the FERV39K dataset are discussed below, and results for the Something‑Something V2 dataset are presented in \cref{sec:ssv2}.
\cref{fig:ferv39k} shows counterfactual video explanations generated by \frameworkAcronym{}, capturing plausible spatiotemporal transformations aligned with the target emotions.
% Counterfactuals typically exhibit distinct facial cues: anger with furrowed brows, tightened lips, and pronounced tension in the eye region, fear with widened eyes, slightly raised eyebrows, and a tensed mouth, sadness with downward mouth corners, tearing eyes, and chin contraction, and happiness with an upward mouth curve, cheek elevation, and softening of the eye region.
For instance, counterfactual videos targeting anger exhibit furrowed brows, tightened lips, and pronounced tension in the eye region.
In the case of Fear, video display widened eyes, slightly raised eyebrows, and a tensed mouth. %, capturing typical visual cues of apprehension.
In videos targeting sadness, subtle cues such as redness around the eyes and nose, gentle downward shifts in the mouth corners, and slight chin muscle contraction emerge consistently.
 Conversely, the happy counterfactuals prominently show upward curvature of the mouth, cheek elevation, and softening of the eye region. %, reflecting typical positive affective expressions.
These changes in mouth movement, eye expression, and brow position correspond to the intended emotions, indicating that \frameworkAcronym{} produces realistic and class-discriminative counterfactuals.

\begin{table*}
\centering
\small
\caption{Quantitative comparison of counterfactuals generated by \frameworkAcronym{} and its variants, evaluated on the FERV39K dataset test split (7847 videos). Best values are highlighted in bold. Results illustrate a trade-off between classification effectiveness (FR) and perceptual quality (SSIM, LPIPS), with all variants maintaining high temporal consistency.}
\begin{tabular}{lcccccc}
\toprule
\textbf{Method} & \textbf{FR~$\uparrow$} & \textbf{SSIM~$\uparrow$} & \textbf{LPIPS~$\downarrow$} & \textbf{FID~$\downarrow$} & \textbf{FVD~$\downarrow$} & \textbf{Temp.~$\uparrow$}\\
\midrule
\frameworkAcronym{} & $\mathbf{1.00\pm0.00} $ & $0.83\pm0.00$  & $0.18\pm0.00$ &  $\mathbf{4.19\pm0.02}$     & $28.1\pm0.69$   & $\mathbf{0.96\pm0.00}$   \\
\frameworkAcronym{}-{wo\_ft} & $\mathbf{1.00\pm0.00} $ & $0.83\pm0.00$  & $0.19\pm0.00$ &  $4.76\pm0.05$     & $35.5\pm0.95$   & $\mathbf{0.96\pm0.00}$   \\
\frameworkAcronym{}-{un\_ft} & $\mathbf{1.00\pm0.00} $ & $0.83\pm0.00$  & $0.18\pm0.00$ &  $4.33\pm0.03$     & $29.3\pm0.76$   & $\mathbf{0.96\pm0.00}$   \\
\frameworkAcronym{}-{un\_wo\_ft} & $\mathbf{1.00\pm0.00} $ & $0.83\pm0.00$  & $0.19\pm0.00$ &  $4.75\pm0.06$     & $35.4\pm1.04$   & $\mathbf{0.96\pm0.00}$   \\
\frameworkAcronym{}-RA & $0.79\pm0.00$ & $\mathbf{0.95\pm0.00}$  & $\mathbf{0.10\pm0.00}$ &  $6.45\pm0.05$     & $\mathbf{19.8\pm0.31}$   & $0.94\pm0.00$   \\
\bottomrule
\end{tabular}%
\label{tab:ferv39k_results}
\end{table*}

\cref{tab:ferv39k_results} reports the quantitative results of counterfactuals generated by different \frameworkAcronym{} variants on the FERV39K dataset (see \cref{sec:evaluation} for metric details).
All variants achieve high flip ratio (FR) and temporal consistency, indicating reliable target classification and coherent motion dynamics.
Despite a slightly lower FR, the RA variant achieves the best perceptual quality, characterized by the highest SSIM and the lowest LPIPS and FVD, indicating visually coherent and artifact-free transformations.
The unconditional variants (\frameworkAcronym{}-{un\_ft} and \frameworkAcronym{}-{un\_wo\_ft}) perform comparably to the conditional fine-tuned model, indicating that explicit text conditioning and diffusion fine-tuning contribute marginally to overall performance in this setting.
These findings collectively highlight a trade-off between classification effectiveness and perceptual realism, with each variant offering a distinct balance between fidelity and discriminative alignment.

\cref{fig:ferv39k_RA} illustrates counterfactuals generated by \frameworkAcronym{} and its RA variants in the FERV39K dataset.
Difference maps are obtained by subtracting each factual frame from its corresponding counterfactual frame.
While \frameworkAcronym{} captures key facial regions such as the brows, eyes, and mouth, its difference maps also reveal diffuse activations across non-relevant areas, such as the microphone region, suggesting residual diffusion noise and less targeted edits.
In contrast, the RA variant produces spatially localized, classifier-specific modifications concentrated around expression-relevant regions, effectively reducing background interference (see the corresponding difference maps). Qualitative results visualizing the diffusion-induced changes are provided in the \cref{sec:qual_echonet,sec:qual_ferv39k}.
These focused and semantically meaningful changes demonstrate the model’s ability to generate more interpretable counterfactuals consistent with class-discriminative facial cues.

% \setlength{\tabcolsep}{2mm} 

%%%%%%%%%%%%%%%%%%%%%%%%%%%%%%%%%%%%%%%%%%%%%
%%% Limitations \& Future work %%%%%%%%%%%%%%
%%%%%%%%%%%%%%%%%%%%%%%%%%%%%%%%%%%%%%%%%%%%%
\section{Limitations \& Future work}

While \frameworkAcronym{} demonstrates strong performance across diverse tasks, some aspects require further investigation to improve reliability and interpretability.
Even though \frameworkAcronym{} performs robustly without fine-tuning across datasets, the FVD score is comparatively high for the EchoNet-Dynamic dataset.
This may be due to the pre-trained 3D causal VAE, which was not adapted to the medical domain and removes fine-grained details during reconstruction.
Our experiment confirms that these details are relevant, as the target model’s regression accuracy drops when evaluated on VAE-reconstructed factual videos.
Future work should explore the domain-specific fine-tuning of the VAE to improve visual fidelity.
Although the artifact suppression module improves visual fidelity and counterfactual alignment, it can reduce task-specific predictive accuracy.
This trade-off highlights the importance of developing principled approaches to disentangle causal effects from reconstruction-induced variations.
Expert validation, especially in high-stakes domains such as medical imaging, remains essential to ensure the clinical plausibility and trustworthiness of generated explanations.
% Additionally, while concepts within CEs are generally intuitive when aligned with human-recognizable features, they become more challenging to interpret when counterfactuals reveal novel or unfamiliar semantic patterns.
% Future work should develop automated methods to extract and interpret such concepts, improving reliability and practical utility.

%%%%%%%%%%%%%%%%%%%%%%%%%%%%%
%%% Conclusion %%%%%%%%%%%%%%
%%%%%%%%%%%%%%%%%%%%%%%%%%%%%

\section{Conclusion}

As deep learning advances in video understanding tasks, the need for explainable methods has become increasingly critical, especially in safety-sensitive domains such as healthcare and surveillance.
Counterfactual explanations offer a powerful means of interpreting model behavior by revealing causal relationships along temporal decision boundaries.
We introduce \frameworkAcronym{}, a novel framework for generating counterfactual video explanations using a latent diffusion model.
By operating in latent space and integrating gradient‑based guidance, \frameworkAcronym{} produces visually plausible counterfactuals at a lower computational cost than pixel‑space approaches. 
A masking‑based artifact suppression module refines outputs by isolating causal edits while minimizing modifications in non‑causal regions.
Comprehensive experiments on diverse tasks show consistent improvements over state‑of‑the‑art baselines.
Collectively, these findings establish \frameworkAcronym{} as a robust and versatile paradigm for producing high‑fidelity, temporally consistent, and interpretable video explanations across various tasks.

\section{Multimedia Material}
All videos corresponding to qualitative results shown in the main paper and supplementary document are provided in the accompanying multimedia supplementary material.
\section{Acknowledgments}
The project was funded by the Federal Ministry for Education and Research (BMBF) with grant number 03ZU1202JA.
\clearpage
\setcounter{page}{1}
\maketitlesupplementary
\appendix

\section{Text-Prompt for Guidance}\label{sec:text-prompt}
Text prompts are constructed from dataset-specific labels and metadata to provide semantic guidance to the diffusion model.
These prompts serve as conditioning signals during two stages: (i) finetuning the diffusion model on each dataset, where they encode factual class information, and (ii) counterfactual generation, where the original class description is replaced with a target class description. 
The following subsections describe the prompt-construction procedure for each dataset.

\subsection{FERV39K}

Each video in the FERV39K dataset contains metadata specifying its \emph{scene} (e.g., Social, School, Interview) and its \emph{facial expression label} (e.g., Angry, Happy, Sad). 
Following the reasoning provided in the FERV39K dataset documentation~\cite{wang2022ferv39k}, the raw scene names are grouped into higher-level \emph{scenario categories} that better capture the interaction type and contextual environment associated with each video.
The complete mapping used in the text preprocessing is as follows:
\begin{itemize}
    \item \textbf{Daily Life:} Argue, Social, School, Medicine, Conflict, Daily-Life  
    \item \textbf{Weak-Interactive Show:} Action, Scholar-Reports, Speech, Elegant-Art, Live-Show, Talk-Show    
    \item \textbf{Strong-Interactive Activity:} Business, Experiment, Official-Event, Crime, Interview, Contest    
    \item \textbf{Anomaly Issue:} History, Terror, War, Crisis
\end{itemize}
After mapping each scene to its scenario, a grammatically coherent caption is constructed for every video. 
Articles (\emph{a} vs.\ \emph{an}) are selected automatically based on the initial vowel sound. 
Each prompt follows the template:
\begin{quote}
``A video in a/an \textless scenario\textgreater{} scenario, showing a/an \textless scene\textgreater{} setting with a/an \textless expression\textgreater{} expression.''
\end{quote}
This template produces a unified natural-language description that integrates the scenario context, the visual scene, and the expressed emotion. 
For example:

\begin{itemize}
    \item ``A video in a daily life scenario, showing a social setting with an angry expression.''
    \item ``A video in a weak-interactive show scenario, showing a speech setting with a surprise expression.''
    \item ``A video in a strong-interactive activity scenario, showing an interview setting with a sad expression.''
    \item ``A video in an anomaly issue scenario, showing a crisis setting with a neutral expression.''
\end{itemize}
These text prompts are used during diffusion model finetuning to provide consistent language conditioning. 

For generating counterfactual explanations, the facial expression in the caption is replaced with the \emph{target} expression.  

\begin{quote}
``A video in a/an \textless scenario\textgreater{} scenario, showing a/an \textless scene\textgreater{} setting with a/an \textless target-expression\textgreater{} expression.''
\end{quote}

Two example prompts are shown below:

\begin{itemize}
    \item \textbf{Factual Prompt:} ``A video in a daily life scenario, showing a social setting with an \textbf{angry} expression.''
    \item \textbf{Counterfactual Prompt:} ``A video in a daily life scenario, showing a social setting with a \textbf{happy} expression.''
    \vspace{0.5 em}
    \item \textbf{Factual Prompt:} ``A video in a weak-interactive show scenario, showing a speech setting with a \textbf{surprise} expression.''
    \item \textbf{Counterfactual Prompt:} ``A video in a weak-interactive show scenario, showing a speech setting with a \textbf{sad} expression.''
\end{itemize}

\subsection{EchoNet-Dynamics}
For the EchoNet-Dynamics dataset~\cite{ouyang2020video}, each video is associated with a continuous left ventricular ejection fraction (LVEF) value.
To provide physiologically meaningful conditioning to the diffusion model, each video is described using a text prompt that embeds its numerical LVEF measurement.
During diffusion-model finetuning, the following textual template is used:
\begin{quote}
``A cardiac ultrasound video of left ventricular ejection fraction of \textless LVEF\textgreater.''
\end{quote}
Here, \textless LVEF\textgreater{} corresponds to the ground-truth ejection fraction provided in the dataset.
This formulation supplies the diffusion model with clinically interpretable conditioning information tied to cardiac function.

For generating counterfactual explanations, the original LVEF value is replaced with a target LVEF that represents the opposite physiological state.
Following standard clinical guidelines, an LVEF value of $\text{LVEF} \geq 50\%$ is interpreted as normal systolic function, while values below this threshold indicate reduced systolic function associated with impaired cardiac performance.
For counterfactual generation, each video is assigned a target LVEF corresponding to the opposite functional state:

\begin{itemize}
\item Normal systolic function ($\text{LVEF} \geq 50\%$) $\rightarrow$ Reduced function: target value set to 35.0
\item Reduced systolic function ($\text{LVEF} \leq 50\%$) $\rightarrow$ Normal function: target value set to 60.0
\end{itemize}
These target values are used to construct counterfactual prompts of the form:

\begin{quote}
``A cardiac ultrasound video of left ventricular ejection fraction of \textless target-LVEF\textgreater.''
\end{quote}

Two examples are shown below:
\begin{itemize}
    \item \textbf{Factual Prompt:} A cardiac ultrasound video of left ventricular ejection fraction of \textbf{62.0}.''
    \item \textbf{Counterfactual Prompt:} A cardiac ultrasound video of left ventricular ejection fraction of \textbf{35.0}.''
    \vspace{0.5em}
    \item \textbf{Factual Prompt:} ``A cardiac ultrasound video of left ventricular ejection fraction of \textbf{38.0}.''  
    \item \textbf{Counterfactual Prompt:} ``A cardiac ultrasound video of left ventricular ejection fraction of \textbf{60.0}.''  
\end{itemize}
% These text prompts ensure consistent language-based conditioning for both finetuning and counterfactual synthesis, while preserving clinical interpretability.

\subsection{Something-Something V2}

The Something-Something V2 dataset~\cite{goyal2017something} provides human-annotated action descriptions such as ``Moving something from left to right'' or ``Pouring something into something.'' 
Each action category is associated with a unique integer identifier, and the official dataset includes a mapping that converts each category into its corresponding natural language action description.

During diffusion-model finetuning, the textual prompt for each video is obtained by directly applying this mapping. 
Thus, every video is conditioned using its ground-truth action description in the form:
\begin{quote}
``\textless action-description\textgreater{}''
\end{quote}
where \textless action-description\textgreater{} denotes the natural-language action description provided by the dataset.

For counterfactual generation, the action label is replaced with the target action category. 
The target label is converted to text using the same mapping, producing a counterfactual prompt of the form:
\begin{quote}
``\textless target-action-description\textgreater{}''
\end{quote}
Finetuning and counterfactual synthesis rely on the dataset’s human-interpretable action descriptions.
Example prompts are shown below:
\begin{itemize}
    \item \textbf{Factual Prompt:} ``Bending something so that it deforms''
    \item \textbf{Counterfactual Prompt:} ``Bending something until it breaks''
    \vspace{0.5em}
    \item \textbf{Factual Prompt:} ``Pouring something into something''
    \item \textbf{Counterfactual Prompt:} ``Pouring something into something until it overflows''
\end{itemize}

The text-guidance is applied with a small scale of $1.0$ during counterfactual synthesis, ensuring that the generated counterfactuals remain meaningfully influenced by the target model guidance.
In the unconditional setting for counterfactual explanations, the diffusion model is provided with an empty text prompt, allowing generation to proceed without explicit text conditioning.
It is also observed that the unconditional model performs comparably to the text-conditioned model in producing counterfactual explanations, suggesting that the target model guidance is sufficient to modify class-relevant attributes even in the absence of textual input.
Overall, text-based conditioning serves as an optional semantic signal during counterfactual generation, allowing datasets with rich label descriptions to guide the diffusion process while still enabling the framework to operate effectively when such textual information is limited or omitted.

\section{Evaluation Metrics}\label{sec:evaluation}

This section formally defines the metrics used to evaluate the generated counterfactual video explanations. 
A diverse set of metrics is used to assess the predictive performance and the perceptual quality of the counterfactuals.
Predictive performance is evaluated using regression and classification accuracy metrics. 
For regression tasks, the coefficient of determination ($R^2$), root mean square error (RMSE), and mean absolute error (MAE) are adopted to quantify the validity of the counterfactuals.
The Flip Ratio (FR) is used in classification tasks to quantify the success rate of the generated counterfactuals.
In addition to predictive performance, the quality of the generated counterfactuals is assessed using perceptual and distributional similarity metrics.
The structural similarity index measure (SSIM) and the learned perceptual image patch similarity (LPIPS) are used to capture visual similarity between factual and counterfactual instances.
The Fréchet inception distance (FID) measures the distributional alignment of generated images with the real data distribution, while the Fréchet video distance (FVD) extends this evaluation to the temporal domain of videos. Together, these metrics provide a comprehensive assessment of the effectiveness, realism, and temporal coherence of the generated counterfactuals.

\subsection{Coefficient of Determination (\texorpdfstring{$\mathbf{R^2}$}{R2})}

The coefficient of determination $R^2$~\cite{nagelkerke1991note} is computed as:
\begin{equation}
    R^2 = 1 - \frac{\sum_{i=1}^{N} (y'_i - \hat{y}_i)^2}{\sum_{i=1}^{N} (y'_i - \bar{y})^2}
\end{equation}
where $y'_i$ and $\hat{y}_i$ denote the target and predicted values, $\bar{y}$ is the mean of the target values, and $N$ is the number of samples.  
$R^2$ is chosen to quantify how well the regression model explains the variance in the target variable, which is important for determining whether counterfactual generation degrades predictive performance.

\subsection{Root Mean Square Error (RMSE)}

The RMSE is calculated as:
\begin{equation}
    \mathrm{RMSE} = \sqrt{\frac{1}{N} \sum_{i=1}^{N} (y'_i - \hat{y}_i)^2}
\end{equation}
RMSE is selected to detect substantial deviations between the predicted and target continuous outputs for counterfactual videos. Large RMSE values may indicate that the generated counterfactuals deviate excessively from the target distribution.

\subsection{Mean Absolute Error (MAE)}

The MAE is computed as:
\begin{equation}
    \mathrm{MAE} = \frac{1}{N} \sum_{i=1}^{N} |y'_i - \hat{y}_i|
\end{equation}
MAE is included to provide an interpretable measure of the average deviation between the predicted and target values. Together with RMSE, it ensures that both typical and extreme prediction errors are monitored during counterfactual evaluation~\cite{chicco2021coefficient}.

\subsection{Flip Ratio (FR)}

For classification tasks, the Flip Ratio is used to measure the validity of counterfactuals. It is the proportion of counterfactual instances that successfully alter the model's prediction to the target class $y'$:
\begin{equation}
    \mathrm{FR} = \frac{\sum_{i=1}^N \mathbf{I}\!\left[f_{\phi}(x'_i) = y_i' \right]}{N}
\end{equation}
where $f_{\phi}(\cdot)$ denotes the classifier, $x'_i$ the generated counterfactual, and $\mathbf{I}[\cdot]$ the indicator function.  
FR assesses the primary goal of counterfactual explanations, which is to change the model's decision while ensuring the modified instance remains realistic.

\subsection{Structural Similarity Index Measure (SSIM)}

The SSIM~\cite{wang2004image} is computed as:
\begin{equation}
    \mathrm{SSIM}(x, x') = \frac{(2\mu_x\mu_{x'} + C_1)(2\sigma_{xx'} + C_2)}{(\mu_x^2 + \mu_{x'}^2 + C_1)(\sigma_x^2 + \sigma_{x'}^2 + C_2)}
\end{equation}
where $\mu_x$ and $\mu_{x'}$ are mean intensities of the factual and counterfactual videos, $\sigma_x^2$ and $\sigma_{x'}^2$ are variances, $\sigma_{xx'}$ is the covariance, and $C_1, C_2$ are small constants.  
SSIM is chosen to evaluate the preservation of structural and luminance information, ensuring that counterfactuals maintain the spatial coherence of the original input.

\subsection{Learned Perceptual Image Patch Similarity (LPIPS)}

The LPIPS~\cite{zhang2018unreasonable} metric is calculated as:
\begin{equation}
    \mathrm{LPIPS}(x, x') = \sum_l \frac{1}{H_l W_l} \sum_{h,w} \| \hat{x}^l_{hw} - \hat{x'}^l_{hw} \|_2^2
\end{equation}
where $\hat{x}^l$ and $\hat{x'}^l$ are normalized deep features from layer $l$ of a pretrained VGG network~\cite{simonyan2014very}, and $H_l$ and $W_l$ denote the spatial dimensions at that layer. LPIPS is computed per frame and then averaged across all frames for videos.
LPIPS is adopted to capture perceptual differences between factual and counterfactual instances in a way that aligns with human visual similarity judgments, beyond simple pixel-wise comparisons.

\subsection{Fréchet Inception Distance (FID)}
The FID~\cite{heusel2017gans} is computed as:
\begin{equation}
    \mathrm{FID}(x, x') = \| \mu_{x} - \mu_{x'} \|_2^2 + \mathrm{Tr} \left( \Sigma_{x} + \Sigma_{x'} - 2(\Sigma_{x} \Sigma_{x'})^{1/2} \right)
\end{equation}
where $(\mu_{x}, \Sigma_{x})$ and $(\mu_{x'}, \Sigma_{x'})$ denote the means and covariances of deep features extracted from the factual and counterfactual samples, respectively.  
FID\footnote{\small The implementation of FID and FVD metrics is taken from \url{https://github.com/universome/stylegan-v?tab=readme-ov-file}.}  is selected to quantify the distributional similarity between factual and counterfactual in a deep feature space, ensuring that counterfactuals remain close to the manifold of realistic samples.

\subsection{Fréchet Video Distance (FVD)}
The FVD~\cite{unterthiner2018towards} is computed analogously to FID, but the features are extracted from an Inflated 3D ConvNet (I3D) pretrained on Kinetics-400. 
FVD is included to extend distributional similarity assessment to the temporal domain, ensuring that generated video counterfactuals are visually plausible and temporally coherent.

\subsection{Temporal Consistency}
Temporal consistency (Temp.) measures the smoothness of visual changes between consecutive frames in a video. 
Following prior works \cite{ceylan2023pix2video,esser2023structure}, temporal consistency is computed by averaging the cosine similarity between the CLIP~\cite{radford2021learning} image embeddings of adjacent frames.

Let $\{f_1, f_2, \dots, f_T\}$ denote a sequence of $T$ frames and let $\phi(\cdot)$ represent the CLIP image encoder. 
The temporal consistency score is defined as
\begin{equation}
    \mathrm{TC} = \frac{1}{T-1} 
    \sum_{t=1}^{T-1}
    \frac{ \phi(f_t)^\top \phi(f_{t+1}) }
    { \lVert \phi(f_t) \rVert_2 \, \lVert \phi(f_{t+1}) \rVert_2 }
\end{equation}
A higher value of temporal consistency indicates stronger temporal coherence and smoother transitions across frames.

\begin{table*}[t]
    \centering
    \small
    \caption{
        Comparison of \frameworkAcronym{} performance on the EchoNet-Dynamic regression task for varying denoising inference steps (\(T\)) and gradient loss scales (\(\lambda_c\)). Results are reported on 1,288 videos from the validation set.
        Best values are highlighted in bold. 
        Higher values of both parameters consistently improved regression accuracy while maintaining comparable perceptual quality. 
        Temporal consistency is preserved across all hyperparameter settings.         
        Based on these results, \(\lambda_c = 0.10\) and \(T = 15\) are selected, as they improve regression accuracy with only a slight degradation in perceptual quality.
    }
    \begin{tabular}{lccccccccc}
    \toprule
        \textbf{Method}  & \textbf{$\mathbf{T}$} & $\boldsymbol{\lambda_{c}}$ & \textbf{$\mathbf{R^2}$~$\uparrow$} & \textbf{MAE~$\downarrow$} & \textbf{RMSE~$\downarrow$} & \textbf{SSIM~$\uparrow$} & \textbf{LPIPS~$\downarrow$}  & \textbf{Temp.~$\uparrow$}\\
    \midrule
        \frameworkAcronym  & 5 & 0.08 & $0.81\pm0.00$ & $3.09\pm0.06$  & $4.87\pm0.06$ &  $\mathbf{0.78\pm0.00}$   &   $\mathbf{0.15\pm0.00}$   & $\mathbf{0.95\pm0.00}$ \\
        \frameworkAcronym  & 5 & 0.09 & $0.83\pm0.00$ & $2.79\pm0.05$  & $4.50\pm0.04$ &   $\mathbf{0.78\pm0.00}$  &   $\mathbf{0.15\pm0.00}$ & $\mathbf{0.95\pm0.00}$  \\
        \frameworkAcronym  & 5 & 0.10 & $0.86\pm0.00$ & $2.54\pm0.03$  & $4.18\pm0.03$ &  $\mathbf{0.78\pm0.00}$   &   $\mathbf{0.15\pm0.00}$ & $\mathbf{0.95\pm0.00}$ \\
    \midrule
        \frameworkAcronym  & 10 & 0.08 & $0.98\pm0.00$ & $0.79\pm0.02$  & $1.71\pm0.03$ &  $0.76\pm0.00$    &   $0.16\pm0.00$ & $\mathbf{0.95\pm0.00}$ \\
        \frameworkAcronym  & 10 & 0.09 & $0.98\pm0.00$ & $0.66\pm0.01$  & $1.49\pm0.03$ &  $0.76\pm0.00$    &   $0.16\pm0.00$ & $\mathbf{0.95\pm0.00}$ \\
        \frameworkAcronym  & 10 & 0.10 & $0.99\pm0.00$ & $0.57\pm0.02$  & $1.31\pm0.04$ &  $0.76\pm0.00$    &   $0.17\pm0.00$ & $\mathbf{0.95\pm0.00}$ \\
    \midrule
        \frameworkAcronym  & 15 & 0.08 & $\mathbf{1.00\pm0.00}$ & $0.37\pm0.01$  & $0.73\pm0.06$ &  $0.75\pm0.00$    &   $0.17\pm0.00$  & $\mathbf{0.95\pm0.00}$ \\
        \frameworkAcronym  & 15 & 0.09 & $\mathbf{1.00\pm0.00}$ & $0.33\pm0.01$  & $0.63\pm0.06$ &  $0.75\pm0.00$    &   $0.17\pm0.00$ & $\mathbf{0.95\pm0.00}$ \\
        \frameworkAcronym  & \textbf{15} & \textbf{0.10} & $\mathbf{1.00\pm0.00}$ & $\mathbf{0.29\pm0.01}$  & $\mathbf{0.55\pm0.06}$ &  $0.75\pm0.00$    &  $0.18\pm0.00$ & $\mathbf{0.95\pm0.00}$ \\
    \bottomrule
    \end{tabular}
    \label{tab:echonet_steps_gls}
\end{table*}

\begin{table*}[t]
    \centering
    \small
    \caption{
        FID and FVD comparison for \frameworkAcronym{} on the EchoNet-Dynamic dataset across varying denoising steps (\(T\)) and gradient loss scales (\(\lambda_c\)). 
        % FID and FVD are computed by padding each \(112\times112\) frame to \(128\times128\) with a black border to match the input resolution of the evaluation models. 
        FID\textsuperscript{*} and FVD\textsuperscript{*} are computed against the reconstructed videos instead of the original inputs, ensuring that the scores capture only the changes introduced by the counterfactual generation and not artifacts from the VAE reconstruction.
        % Lower values indicate better performance. 
        Best results are highlighted in bold. 
        The low FVD\textsuperscript{*}/FID\textsuperscript{*} values indicate that, once reconstruction artifacts are removed, the counterfactual modifications remain small, temporally consistent, and close to the real video distribution.
        Based on these results, \(\lambda_c = 0.10\) and \(T = 15\) are selected, as they improve regression accuracy with a slight increase in FVD and FID.
    }
    \begin{tabular}{lcccccc}
    \toprule
        \textbf{Method}  & \textbf{$\mathbf{T}$} & $\boldsymbol{\lambda_{c}}$ & \textbf{FVD~$\downarrow$} & \textbf{FVD\textsuperscript{*}~$\downarrow$} & \textbf{FID~$\downarrow$} & \textbf{FID\textsuperscript{*}~$\downarrow$}\\
    \midrule
        \frameworkAcronym  & 5 & 0.08 & $\mathbf{376\pm7}$ &  $\mathbf{19.1\pm0.71}$  & $\mathbf{31.5\pm0.61}$& $\mathbf{3.51\pm0.32}$\\
        \frameworkAcronym  & 5 & 0.09 &  $381\pm6$ & $22.7\pm0.85$  & $32.3\pm0.65$  & $4.33\pm0.34$\\
        \frameworkAcronym  & 5 & 0.10 & $386\pm6$ &  $27.0\pm1.06$ & $33.2\pm0.68$ & $5.39\pm0.41$\\
    \midrule
        \frameworkAcronym  & 10 & 0.08 & $383\pm5$ & $25.4\pm1.15$ & $34.6\pm0.68$ & $5.48\pm0.54$\\
        \frameworkAcronym  & 10 & 0.09 & $390\pm4$ &  $30.6\pm1.59$ & $35.7\pm0.70$ & $6.77\pm0.67$\\
        \frameworkAcronym  & 10 & 0.10 & $397\pm5$ &  $36.7\pm2.01$ & $37.1\pm0.69$ & $8.34\pm0.84$\\
    \midrule
        \frameworkAcronym  & 15 & 0.08 & $392\pm9$ & $31.7\pm1.54$ & $35.9\pm0.60$ & $7.09\pm0.72$\\
        \frameworkAcronym  & 15 & 0.09 & $400\pm9$ &  $39.0\pm1.69$ & $37.5\pm0.65$  & $8.91\pm0.84$\\
        \frameworkAcronym  & \textbf{15} & \textbf{0.10} & $412\pm7$ &  $47.3\pm0.52$ & $39.4\pm0.82$ & $11.8\pm0.59$\\
    \bottomrule
    \end{tabular}
    
    \label{tab:echonet_fvd}
\end{table*}

\section{Ablation Studies}\label{sec:ablation_Study}
In \frameworkAcronym{}, counterfactual generation is influenced by three key hyperparameters: the gradient loss scale \((\lambda_c)\), the number of denoising inference steps \((T)\), and the artifact suppression threshold \((t_{\mathrm{sup}})\).  
Systematic ablation studies are conducted to assess their impact on the quality of generated counterfactuals.  
Initially, \(\lambda_c\) and \(T\) are varied, as they directly control the strength of the classifier guidance and quality of generated sequences.
After identifying optimal values of \(\lambda_c\) and \(T\), the artifact suppression threshold \((t_{\mathrm{sup}})\) is tuned using the RA variant to improve visual quality further and reduce denoising artifacts.  
Ablation experiments are performed independently on the EchoNet-Dynamic and FERV39K datasets to account for domain-specific differences in video characteristics and complexity.

\subsection{Hyperparameters for the EchoNet-Dynamic Dataset}\label{sec:ab_echonet}

The performance of \frameworkAcronym{} is evaluated using regression accuracy metrics (\(R^2\), MAE, RMSE) to quantify the effectiveness of prediction shifts, and perceptual similarity metrics (SSIM, LPIPS) to assess structural preservation and visual fidelity of the generated counterfactuals. 
Temporal consistency is measured to ensure coherent spatiotemporal transformations across frames. 
In addition, generative quality is assessed using FID and FVD, along with their reconstruction-based variants (FID\textsuperscript{*}, FVD\textsuperscript{*}) to disentangle causal changes from VAE reconstruction artifacts.

\subsubsection{Gradient Loss Scale and Denoising Steps}

\begin{table*}[h]
    \centering
    \caption{
        Comparison of \frameworkAcronym{}-RA performance on the EchoNet-Dynamic regression task for different artifact suppression thresholds used to reduce denoising artifacts. Results are reported on 1288 videos from the validation set. Best values are highlighted in bold. An intermediate threshold (\(t_{\mathrm{sup}} = 0.02\)) is selected as a trade-off between regression accuracy and perceptual quality.
    }
    \begin{tabular}{lccccccc}
    \toprule
        \textbf{Method}  & \textbf{$\mathbf{t_{sup}}$} & \textbf{$\mathbf{R^2}$~$\uparrow$} & \textbf{MAE~$\downarrow$} & \textbf{RMSE~$\downarrow$} & \textbf{SSIM~$\uparrow$} & \textbf{LPIPS~$\downarrow$} & \textbf{Temp.~$\uparrow$}\\
    \midrule
        \frameworkAcronym-RA & 0.01 & $\mathbf{0.95\pm0.00}$ & $\mathbf{1.75\pm0.04}$ & $\mathbf{2.36\pm0.05}$  & $0.84\pm0.00$ &  $0.12\pm0.00$   & $0.93\pm0.00$ \\
        \frameworkAcronym-RA & \textbf{0.02} & $0.86\pm0.00$ & $3.07\pm0.05$  & $4.14\pm0.06$ &  $0.89\pm0.00$    &   $0.09\pm0.00$ & $\mathbf{0.94\pm0.00}$ \\
        \frameworkAcronym-RA & 0.03 & $0.73\pm0.01$ & $4.34\pm0.10$ & $5.76\pm0.12$  & $0.91\pm0.00$ &  $0.08\pm0.00$  & $\mathbf{0.94\pm0.00}$ \\
        \frameworkAcronym-RA & 0.04 & $0.57\pm0.02$ & $5.57\pm0.13$ & $7.23\pm0.14$  & $\mathbf{0.93\pm0.00}$ &  $\mathbf{0.06\pm0.00}$  & $\mathbf{0.94\pm0.00}$ \\
    \bottomrule
    \end{tabular}
    \label{tab:echonet_thres}
\end{table*}

\begin{table*}[t]
    \centering
    \caption{
        FID and FVD comparison for \frameworkAcronym{}-RA on the EchoNet-Dynamic dataset for different artifact suppression thresholds \((t_{\mathrm{sup}})\). 
        FID and FVD are computed with respect to the original videos, while FID\textsuperscript{*} and FVD\textsuperscript{*} are computed against VAE-reconstructed videos. 
        Best values are highlighted in bold.
        Increasing \(t_{\mathrm{sup}}\) progressively lowers FID and FVD, as more of the original video is preserved and diffusion-induced artifacts are suppressed, whereas FID\textsuperscript{*} and FVD\textsuperscript{*} increase because the refined counterfactuals deviate from the smoothed VAE reconstructions.
         An intermediate threshold (\(t_{\mathrm{sup}} = 0.02\)) is selected as a trade-off between regression accuracy and distributional alignment.
    }
    
    \begin{tabular}{lcccccc}
    \toprule
        \textbf{Method}  & \textbf{$\mathbf{t_{sup}}$} & \textbf{FVD~$\downarrow$} & \textbf{FVD\textsuperscript{*}~$\downarrow$} & \textbf{FID~$\downarrow$} & \textbf{FID\textsuperscript{*}~$\downarrow$}\\
    \midrule
        \frameworkAcronym-RA & 0.01 & $157\pm8.11$ & $\mathbf{163\pm6.38}$ & $23.09\pm0.92$ & $\mathbf{15.2\pm0.84}$ \\
        \frameworkAcronym-RA & \textbf{ 0.02} & $101\pm5.98$ & $221\pm4.98$ & $15.6\pm0.79$ & $17.7\pm0.60$ \\
        \frameworkAcronym-RA & 0.03 &  $75.5\pm4.25$ & $250\pm4.55$ &$11.7\pm0.63$ & $19.7\pm0.47$ \\
        \frameworkAcronym-RA & 0.04 &  $\mathbf{59.9\pm3.46}$ & $269\pm3.65$ &$\mathbf{9.38\pm0.53}$ & $21.2\pm0.43$ \\
    \bottomrule
    \end{tabular}
    \label{tab:echonet_fvd_thres}
\end{table*}

In counterfactual generation, increasing the number of denoising inference steps \((T)\) provides the model with more opportunities to iteratively refine the generated video and incorporate guidance of the target model. 
The number of denoising steps corresponds to the same number of steps for which noise is added initially.  
Although, higher \(T\) values can enhance prediction shifts, excessive values may also lead to over-modification of the input, potentially affecting perceptual quality. 
Similarly, increasing the gradient loss scale \((\lambda_c)\) strengthens the influence of gradients during generation and can improve counterfactual success by steering the generation towards the desired prediction change more effectively.
However, excessively large values may introduce out-of-distribution noise, which can appear as unrealistic structures or subtle artifacts in the generated frames.

\cref{tab:echonet_steps_gls} shows that regression accuracy consistently improved with higher values of \(T\) and \(\lambda_c\).  
Perceptual metrics (SSIM and LPIPS) exhibited minor variations, indicating that the accuracy gains are achieved with only slight degradation in visual quality. 
Notably, temporal consistency remains high across all settings (\(\text{Temp.} \approx 0.95\)), indicating that neither guidance scale nor denoising disrupts motion coherence.
Based on these results, \(\lambda_c = 0.10\) and \(T = 15\) are selected as the optimal values for subsequent experiments on the EchoNet-Dynamic dataset.

\cref{tab:echonet_fvd} reports FID and FVD values, along with their reconstruction-based variants (FID\textsuperscript{*}, FVD\textsuperscript{*}). 
FID\textsuperscript{*} and FVD\textsuperscript{*} denote scores computed against the reconstructed videos obtained from the 3D causal VAE of CogVideoX, rather than the original inputs.
These scores are computed by padding each \(112\times112\) frame to \(128\times128\) with a black border to match the input resolution of the evaluation models. 
The quality of the generated counterfactuals depends on the reconstruction of the 3D causal VAE, which is not fine-tuned on the medical domain and removes fine-grained details, leading to increased FVD scores. This reflects a limitation of the pretrained VAE rather than the counterfactuals themselves, and FVD/FID should therefore not be interpreted in the same way as for pixel-space approaches.
The substantially lower FVD\textsuperscript{*}/FID\textsuperscript{*} scores indicate that, once reconstruction artifacts are removed, the counterfactual changes introduced by \frameworkAcronym{} remain temporally consistent, and close to the real video distribution. 
These findings indicate that while FVD should be interpreted cautiously on EchoNet-Dynamic, FVD\textsuperscript{*}/FID\textsuperscript{*} provides a more precise measure of the actual generative changes induced by the counterfactual process.

\begin{table*}[h]
    \centering
    \caption{
        Quantitative evaluation of \frameworkAcronym{} on the FERV39K classification task for varying denoising inference steps (\(T\)) and gradient loss scales (\(\lambda_c\)).
        Results are reported on 7,847 samples from the test set. Best values are highlighted in bold. 
        % \(\lambda_c = 15\) and \(T = 20\) are selected as optimal values for subsequent experiments.
        Higher values of both parameters generally improve classification success, with a slight trade-off in perceptual quality.
        A balance between classification performance and perceptual quality is achieved with \(\lambda_c = 20\) and \(T = 15\), which are selected as the optimal settings for subsequent experiments on the FERV39K dataset.
    }
    \begin{tabular}{lcccccccc}
    \toprule
        \textbf{Method}  & \textbf{$\mathbf{T}$} & $\boldsymbol{\lambda_{c}}$ & \textbf{FR~$\uparrow$} & \textbf{SSIM~$\uparrow$} & \textbf{LPIPS~$\downarrow$} & \textbf{FID~$\downarrow$} & \textbf{FVD~$\downarrow$} & \textbf{Temp.~$\uparrow$}\\
    \midrule
        \frameworkAcronym  & 5 & 20 & $0.88\pm0.00$ & $\mathbf{0.86\pm0.00}$  & $\mathbf{0.16\pm0.00}$ &  $\mathbf{3.78\pm0.03}$    &  $\mathbf{25.3\pm0.68}$   & $\mathbf{0.96\pm0.00}$   \\
        \frameworkAcronym  & 5 & 25 &  $0.92\pm0.00$ & $0.85\pm0.00$  & $\mathbf{0.16\pm0.00}$ &  $3.80\pm0.03$    & $26.0\pm0.67$   & $\mathbf{0.96\pm0.00}$  \\
        \frameworkAcronym  & 5 & 30 & $0.94\pm0.00$ &  $0.85\pm0.00$ & $\mathbf{0.16\pm0.00}$ &   $3.83\pm0.03$    & $26.6\pm0.70$    & $\mathbf{0.96\pm0.00}$   \\
    \midrule
        \frameworkAcronym  & 10 & 20  & $0.98\pm0.00$ &  $0.84\pm0.00$ & $0.17\pm0.00$ &  $4.01\pm0.05$   & $27.0\pm0.91$   & $\mathbf{0.96\pm0.00}$    \\
        \frameworkAcronym  & 10 & 25  & $0.99\pm0.00$ & $0.84\pm0.00$  & $0.17\pm0.00$ &   $4.05\pm0.05$   & $27.6\pm0.92$   & $\mathbf{0.96\pm0.00}$      \\
        \frameworkAcronym  & 10 & 30  & $0.99\pm0.00$ &  $0.84\pm0.00$ & $0.17\pm0.00$ &   $4.09\pm0.05$   & $28.3\pm0.95$    & $\mathbf{0.96\pm0.00}$     \\
    \midrule
        \frameworkAcronym  & \textbf{15} & \textbf{20}  & $\mathbf{1.00\pm0.00} $ & $0.83\pm0.00$  & $0.18\pm0.00$ &  $4.19\pm0.02$     & $28.1\pm0.69$   & $\mathbf{0.96\pm0.00}$   \\
        \frameworkAcronym  & 15 & 25  & $\mathbf{1.00\pm0.00} $ & $0.83\pm0.00$  & $0.18\pm0.00$ &  $4.24\pm0.03$     & $28.6\pm0.71$   & $\mathbf{0.96\pm0.00}$      \\
        \frameworkAcronym  & 15 & 30  & $\mathbf{1.00\pm0.00} $ & $0.83\pm0.00$  & $0.18\pm0.00$ &   $4.29\pm0.03$   & $29.3\pm0.73$   & $\mathbf{0.96\pm0.00}$ \\
    \bottomrule
    \end{tabular}    
    \label{tab:ferv39k_steps_gls}
\end{table*}

\begin{table*}
    \centering
    \caption{
        Quantitative evaluation of \frameworkAcronym{} on the FERV39K classification task for varying artifact suppression thresholds (\(t_{\mathrm{sup}}\)) applied to reduce denoising artifacts. Results are reported on 7,847 samples from the test set. Best values are highlighted in bold. A threshold (\(t_{\mathrm{sup}} = 0.10\)) provided the best trade-off, effectively reducing artifacts while maintaining classification success and perceptual quality.
    }
    \begin{tabular}{lcccccccc}
    \toprule
        Method  & \textbf{$\mathbf{t_{sup}}$} & \textbf{FR~$\uparrow$} & \textbf{SSIM~$\uparrow$} & \textbf{LPIPS~$\downarrow$} & \textbf{FID~$\downarrow$} & \textbf{FVD~$\downarrow$} & \textbf{Temp.~$\uparrow$}\\
    \midrule
        \frameworkAcronym-RA  & 0.01 & $\mathbf{1.00\pm0.00} $ & $0.84\pm0.00$  & $0.21\pm0.00$ &  $7.86\pm0.12$     & $47.3\pm0.79$   & $\mathbf{0.94\pm0.00}$   \\
        \frameworkAcronym-RA  & 0.03 & $0.99\pm0.00 $ & $0.88\pm0.00$  & $0.20\pm0.00$ &  $11.8\pm0.19$     & $57.4\pm0.82$   & $0.93\pm0.00$   \\
        \frameworkAcronym-RA  & 0.05 & $0.96\pm0.00 $ & $0.91\pm0.00$  & $0.16\pm0.00$ &  $10.5\pm0.14$     & $44.7\pm0.49$   & $0.93\pm0.00$   \\
        \frameworkAcronym-RA & 0.07 & $0.90\pm0.00 $ & $0.93\pm0.00$  & $0.13\pm0.00$ &  $8.61\pm0.09$     & $32.5\pm0.33$   & $0.93\pm0.00$   \\
        \frameworkAcronym-RA & \textbf{0.10} & $0.79\pm0.00$ & $\mathbf{0.95\pm0.00}$  & $\mathbf{0.10\pm0.00}$ &  $\mathbf{6.45\pm0.05}$     & $\mathbf{19.8\pm0.31}$   & $\mathbf{0.94\pm0.00}$   \\
    \bottomrule
    \end{tabular}
    \label{tab:ferv39k_thres}
\end{table*}

\subsubsection{Artifact Suppression Threshold}

The artifact suppression threshold \((t_{\mathrm{sup}})\) mitigates denoising-related artifacts that may appear during counterfactual generation.  
Such artifacts typically arise from the denoising process and may include high-frequency distortions, unnatural textures, or subtle structural inconsistencies in the generated frames. 
By applying this threshold, minor changes below \(t_{\mathrm{sup}}\) are suppressed, reducing the visual impact of residual denoising errors while preserving the overall structural coherence of the video.

\cref{tab:echonet_thres} shows the effect of varying \(t_{\mathrm{sup}}\) by using the previously selected optimal values of \(\lambda_c\) and \(T\).  
Lower thresholds failed to entirely suppress denoising artifacts, resulting in a reduced perceptual quality. 
Conversely, higher thresholds significantly reduced artifacts but at the cost of oversmoothing and reduced regression accuracy.
Temporal consistency (Temp.) remains high and nearly constant across all thresholds (\(\approx 0.93{-}0.94\)), suggesting that the refinement step does not compromise motion coherence. 
An intermediate threshold value of \(t_{\mathrm{sup}} = 0.02\) offers a trade-off, yielding substantially improved perceptual quality while maintaining competitive regression accuracy, and is therefore adopted for all subsequent experiments on the EchoNet-Dynamic dataset.

The behavior of FID and FVD score across different \(t_{\mathrm{sup}}\) values is summarized in \cref{tab:echonet_fvd_thres}.
When FID and FVD are computed against the original videos, higher thresholds lead to lower scores, as more factual content is preserved and diffusion-induced artifacts are reduced.
In contrast, the FID\textsuperscript{*}/FVD\textsuperscript{*} evaluated against VAE-reconstructed videos increases as the threshold grows. This increase occurs because preserving more of the original structure causes the refined counterfactuals to diverge from the smoother, lower-frequency VAE reconstructions.
This pattern is consistent with earlier observation that the 3D causal VAE removes fine-grained details, higher \(t_{\mathrm{sup}}\) restores sharper, more realistic structures that are closer to the original data but farther from the VAE reconstruction manifold.

%%%%%%%%%%%%%%%%%%%%%%%%%%%%%%%%%%%%%%%%%%%%%%%%%%%%%%%%%%%%%%%%%%%%%%%%%%%%%%%%%%%%%%%%%%%%%%%%%%%%%%%%%%%%%%%%%%%%%%%%%%%%%%%%%%%%%%%%%%%%%%%%%%%%%%%%%%%%%%%%%%%%%%%%%%%%%%%%%%%%%%%%%%%%%%%%%%%%%%%%%%%%%%%%%%%%%%%%%%%%%%%%%%%%%%%%%%%%%%%%%%%%%%%%%%%%%%%%%%%%%%%%%%%%%%%%%%%%%%%%%%%%%%%%%%%%%%%%%%%%%%%%%%%%%%%%%%%%%%%%%%%%%%%%%%%%%%%%%%%%%%%%%%%%%%%%%%%%%%%%%%%%%%%%%%%%%%%%%%%%%%%%%%%%%%%%%%%%%%%%%%%%%%%%%%%%%%%%%%%%%%%%%%%%%%%%%%%%%%%%%%%%%%%%%%%%%%%%%%%%%%%%%%%%%%%%%%%%%%%%%%%%%%%%%%%%%%%%%%%%%%%%%%%%%%%%%%%%%%%%%%%%%%%%%%%%%%%%%%%%%%%%%%%%%%%%%%%%%%%%%%%%%%%%%%%%%%%%%%%%
\subsection{Hyperparameters for the FERV39K Dataset}

The performance of \frameworkAcronym{} on the FERV39K dataset is evaluated using the classification success rate (FR), perceptual similarity metrics (SSIM and LPIPS), and distributional alignment measures (FID and FVD). 
FR quantifies the effectiveness of counterfactuals in flipping the classifier prediction, while SSIM and LPIPS assess visual similarity to the factual video. 
FID and FVD capture realism and temporal coherence at the dataset level. 
Temporal consistency is additionally measured to ensure that the generated counterfactuals maintain smooth and coherent motion across frames. 
For each video, the target class is randomly sampled from the non-predicted classes using a fixed random seed to ensure reproducibility across runs.

\subsubsection{Gradient Loss Scale and Denoising Steps}

A similar impact of the denoising inference steps \((T)\) and gradient loss scale \((\lambda_c)\) is observed on the FERV39K dataset as in the EchoNet‑Dynamic experiments.
Higher values of these hyperparameters improved classification performance but slightly reduced perceptual quality. 
As shown in \cref{tab:ferv39k_steps_gls}, FR improved consistently with higher values of the denoising inference steps \((T)\) and the gradient loss scale \((\lambda_c)\).  
In particular, increasing \(T\) enhanced the incorporation of classifier guidance, while larger \(\lambda_c\) values strengthened the influence of classifier gradients, leading to more decisive counterfactual modifications.  
However, perceptual metrics (SSIM and LPIPS) declined slightly at higher parameter settings, and distributional metrics (FID and FVD) increased moderately, indicating a small trade‑off between classification accuracy and perceptual realism.
Temporal consistency remains high and stable across all settings (\(\text{Temp.} \approx 0.96\)), demonstrating that the diffusion process preserves coherent motion independently of parameter choices.
A balance between classification performance and perceptual quality is achieved with \(\lambda_c = 20\) and \(T = 15\), which are selected as the optimal settings for subsequent experiments on the FERV39K dataset.

\subsubsection{Artifact Suppression Threshold}

The artifact suppression threshold \((t_{\mathrm{sup}})\) controls the extent to which denoising artifacts are removed during the refinement stage.  
Higher thresholds reduced denoising artifacts and improved perceptual and distributional alignment metrics (SSIM, LPIPS, FID, FVD) but lowered classification success (FR), while lower thresholds preserved accuracy at the cost of leaving residual artifacts.
\cref{tab:ferv39k_thres} indicates that increasing \(t_{\mathrm{sup}}\) progressively improved visual and distributional quality but led to decreased FR.
This comes with the expected trade-off of reduced classification success, since a higher suppression threshold also removes some classifier-relevant changes.
Temporal consistency remains stable at \(\approx 0.93{-}0.94\), indicating that refinement affects spatial quality more than temporal coherence.  
A threshold of \(t_{\mathrm{sup}} = 0.10\) provides the best perceptual quality and distributional alignment, and is therefore adopted for subsequent experiments on the FERV39K dataset.

%%%%%%%%%%%%%%%%%%%%%%%%%%%%%%%%%%%%%%%%%%%%%%%%%%%%%%%%%%%%%%%%%%%%%%%%%%%%%%%%%%%%%%%%%%%%%%%%%%%%%%%%%%%%%%%%%%%%%%%%%%%%%%%%%%%%%%%%%%%%%%%%%%%%%%%%%%%%%%%%%%%%%%%%%%%%%%%%%%%%%%%%%%%%%%%%%%%%%%%%%%%%%%%%%%%%%%%%%%%%%%%%%%%%%%%%%%%%%%%%%%%%%%%%%%%%%%%%%%%%%%%%%%%%%%%%%%%%%%

\section{EchoNet-Dynamic Dataset Counterfactual Results}\label{sec:qual_echonet}

\begin{figure*}[h]
    \centering
    \includegraphics[width=\linewidth]{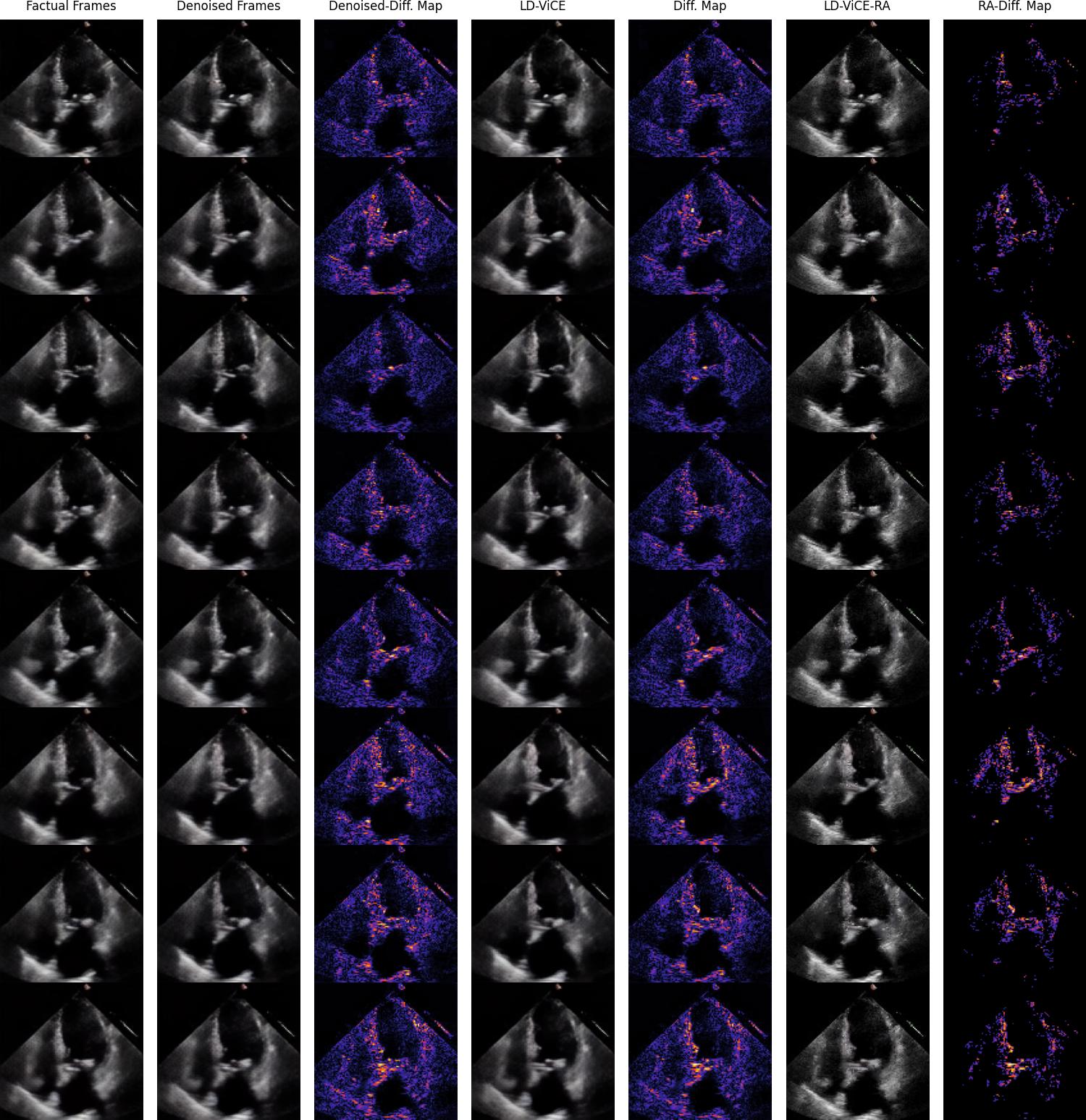} 
    \caption{
        Qualitative counterfactual results on the EchoNet-Dynamic dataset. 
        Eight frames of a video are displayed.
        Factual frames and their guidance-free denoised versions are shown with difference maps (Denoised-Diff.\ Map) visualizing diffusion-induced changes. 
        Classifier-guided counterfactuals generated by \frameworkAcronym{} introduce targeted adjustments, highlighted in the difference maps (Diff.\ Map). 
        The RA variant suppresses high-frequency artifacts, yielding cleaner counterfactuals, with difference maps (RA-Diff.\ Map) showing only the salient, causal changes.
    }
    \label{fig:echo_video_0_cmp}
\end{figure*}

\begin{figure*}[h]
    \centering
    \includegraphics[width=\linewidth]{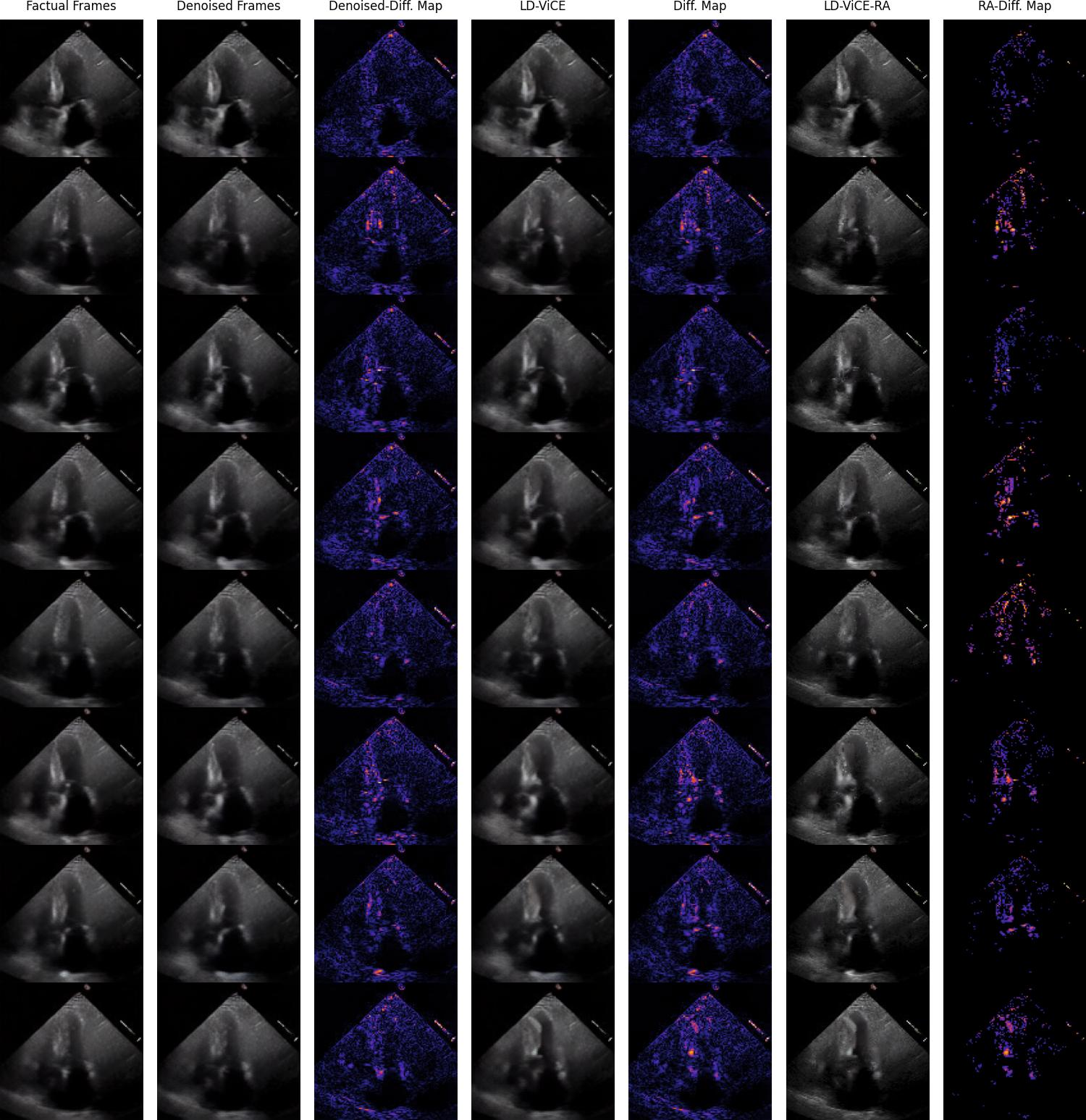} 
    \caption{
        Qualitative counterfactual results on the EchoNet-Dynamic dataset. 
        Eight frames of a video are displayed.
        Factual frames and their guidance-free denoised versions are shown with difference maps (Denoised-Diff.\ Map) visualizing diffusion-induced changes. 
        Classifier-guided counterfactuals generated by \frameworkAcronym{} introduce targeted adjustments, highlighted in the difference maps (Diff.\ Map). 
        The RA variant suppresses high-frequency artifacts, yielding cleaner counterfactuals, with difference maps (RA-Diff.\ Map) showing only the salient, causal changes.
    }
    \label{fig:echo_video_2_cmp}
\end{figure*}

\cref{fig:echo_video_0_cmp,fig:echo_video_2_cmp} present qualitative examples of counterfactual video explanations generated by \frameworkAcronym{} and its RA variant.
The difference maps highlight the regions altered between the factual and counterfactual frames.
The factual frames are first shown alongside their denoised counterparts, which are obtained through a guidance-free denoising process.  
The corresponding difference maps (Denoised Difference Maps) visualize the changes introduced solely by the diffusion denoising dynamics, revealing fine-grained variations that arise even without the influence of the classifier.
The subsequent columns display counterfactual frames produced by \frameworkAcronym{} under classifier guidance.
These counterfactuals incorporate targeted adjustments driven by the regression model’s gradients, along with diffusion-induced changes and the difference maps (Denoised Difference Maps) highlight these combined changes.
Finally, the counterfactuals obtained through the RA refinement variant are shown.
By suppressing changes below the artifact-suppression threshold, the RA variant reduces high-frequency denoising artifacts that appear in the gradient-guided outputs.  
This results in smoother, more stable, and visually cleaner counterfactual sequences.  
The difference maps (RA–Difference Maps) further demonstrate that only salient and target model-induced changes are preserved, while spurious diffusion-induced noise is effectively removed.

\begin{figure*}[h]
    \centering
    \includegraphics[width=0.98\linewidth]{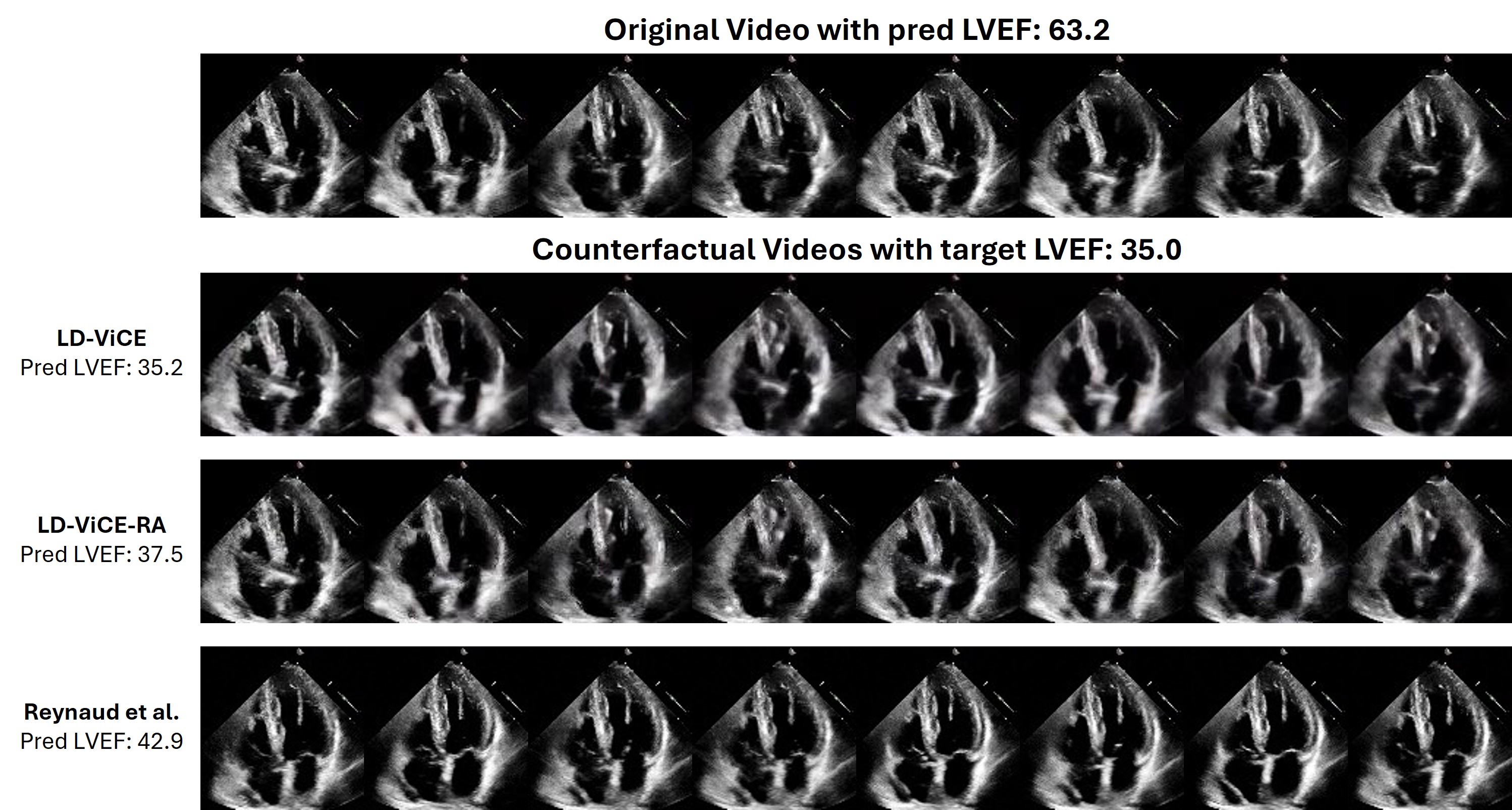} 
    \caption{
        Qualitative comparison of counterfactual explanations on the EchoNet-Dynamics dataset, demonstrating \frameworkAcronym{}’s improved alignment between counterfactual explanations and target regression values. The first row displays eight frames from the original video, while the following rows show counterfactuals generated using \frameworkAcronym, \frameworkAcronym-RA, and 1SCM~\cite{reynaud2023feature}, respectively. The predicted and LVEF values are given on the left.
    }
    \label{fig:echo_sample2}
\end{figure*}

\begin{figure*}[h]
    \centering
    \includegraphics[width=0.98\linewidth]{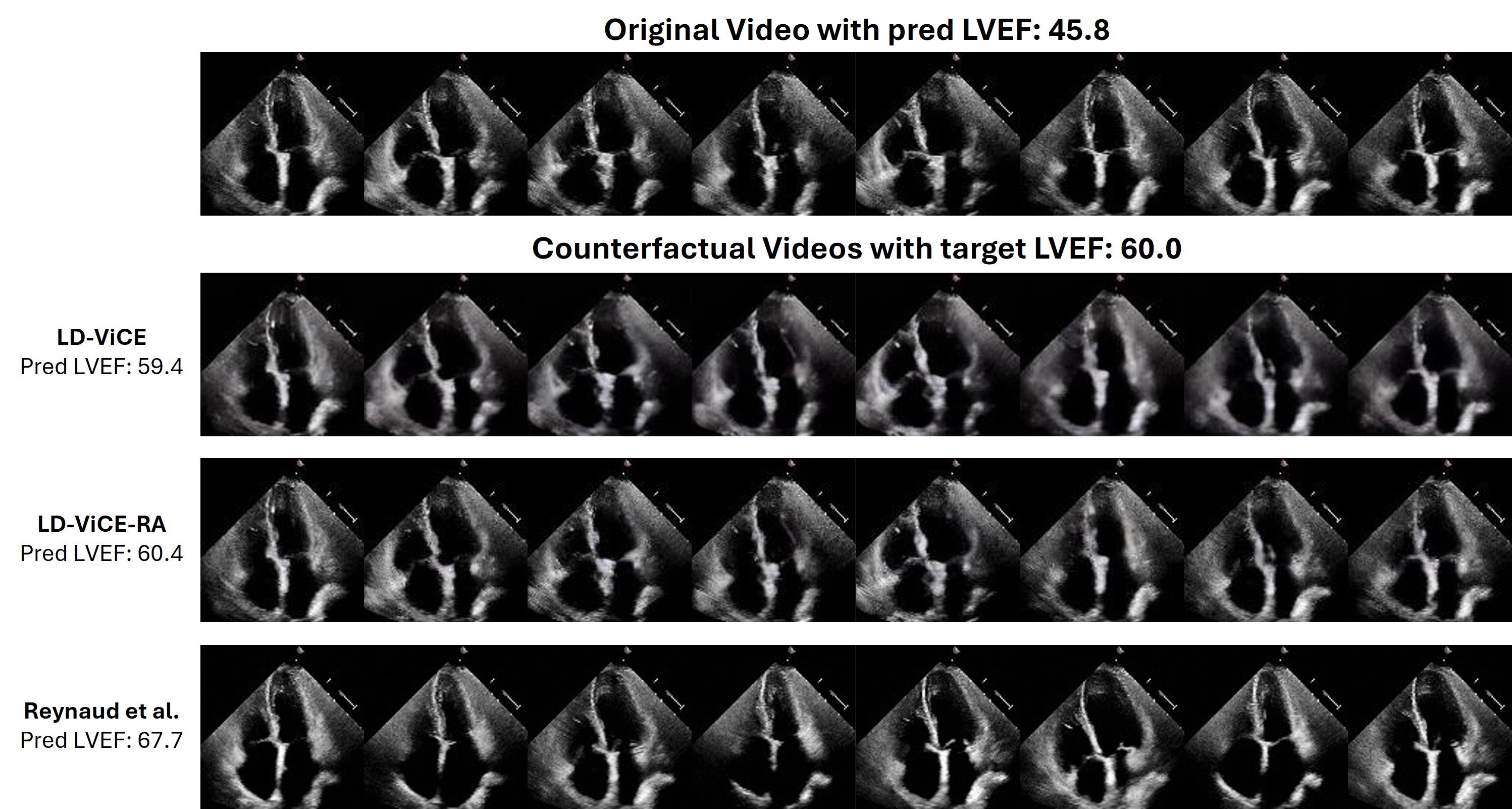} 
    \caption{
        Qualitative comparison of counterfactual explanations on the EchoNet-Dynamics dataset, demonstrating \frameworkAcronym{}’s improved alignment between counterfactual explanations and target regression values. The first row displays eight frames from the original video, while the following rows show counterfactuals generated using \frameworkAcronym, \frameworkAcronym-RA, and 1SCM~\cite{reynaud2023feature}, respectively. The predicted and LVEF values are given on the left.
    }
    \label{fig:echo_sample6}
\end{figure*}

\cref{fig:echo_sample2,fig:echo_sample6} provide a qualitative comparison with 1SCM~\cite{reynaud2023feature} on the EchoNet-Dynamic dataset, highlighting \frameworkAcronym{}’s ability to generate counterfactual explanations that are better aligned with the target regression values compared to baseline methods.
The results illustrate both the effectiveness of classifier-guided generation and the visual fidelity of the produced counterfactuals.

\clearpage
\twocolumn

\section{FERV39K Dataset Counterfactual Results}\label{sec:qual_ferv39k}

\begin{figure*}[h]
    \centering
    \includegraphics[width=\linewidth]{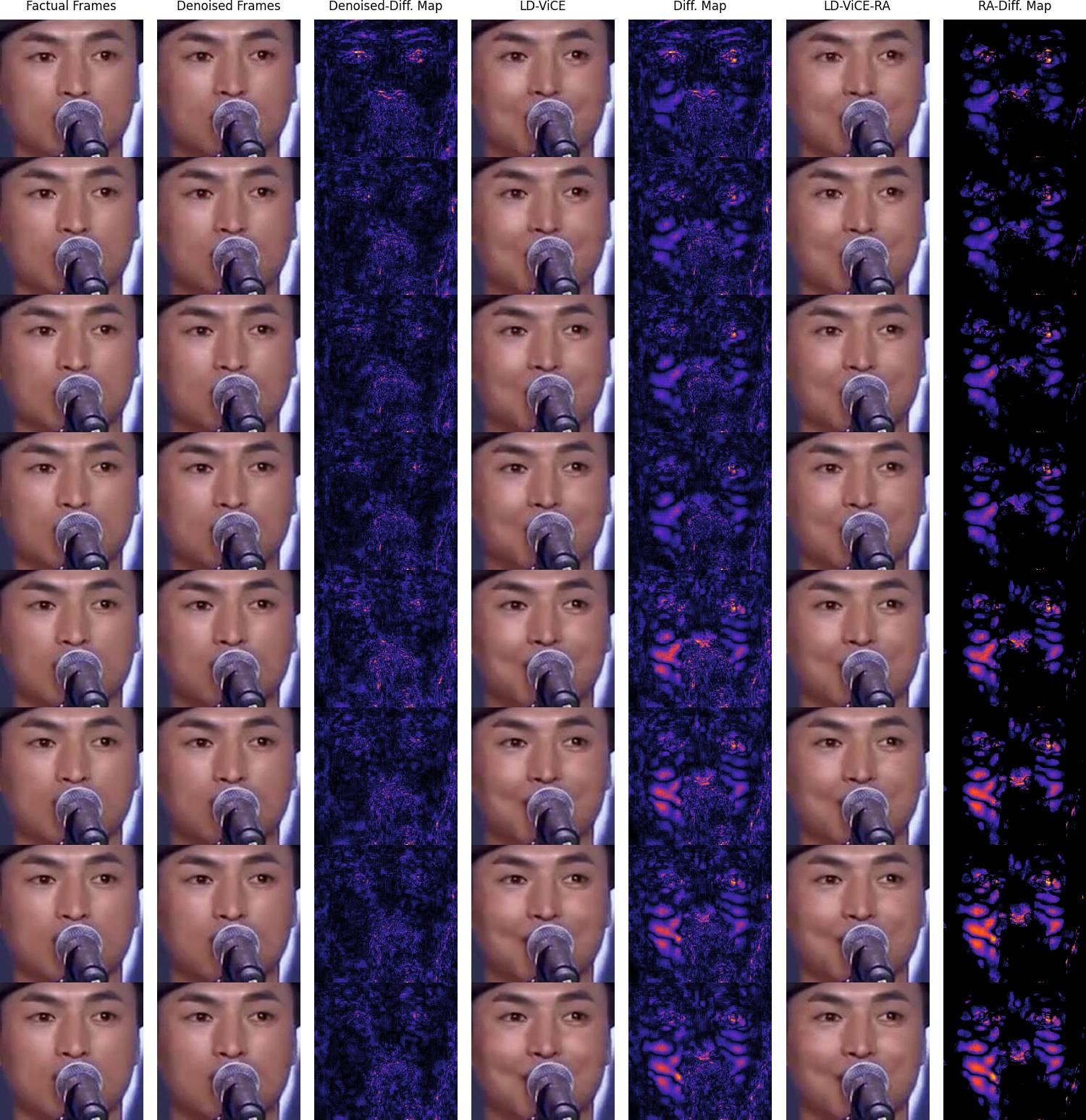} 
    \caption{
        Qualitative counterfactual results on the FERV39K dataset. 
        Factual and denoised frames are shown with difference maps (Denoised-Diff.\ Map) visualizing diffusion-induced changes. 
        Classifier-guided counterfactuals generated by \frameworkAcronym{} introduce facial adjustments, such as cheek lifting and softening of the mouth curvature aligned with the \textit{Happy} emotion, highlighted in the difference maps (Diff.\ Map). 
        The RA variant suppresses high-frequency artifacts and yields cleaner counterfactuals, with difference maps (RA-Diff.\ Map) showing only the salient, class-relevant modifications.
    }
    \label{fig:video_12_cmp}
\end{figure*}

\begin{figure*}[h]
    \centering
    \includegraphics[width=\linewidth]{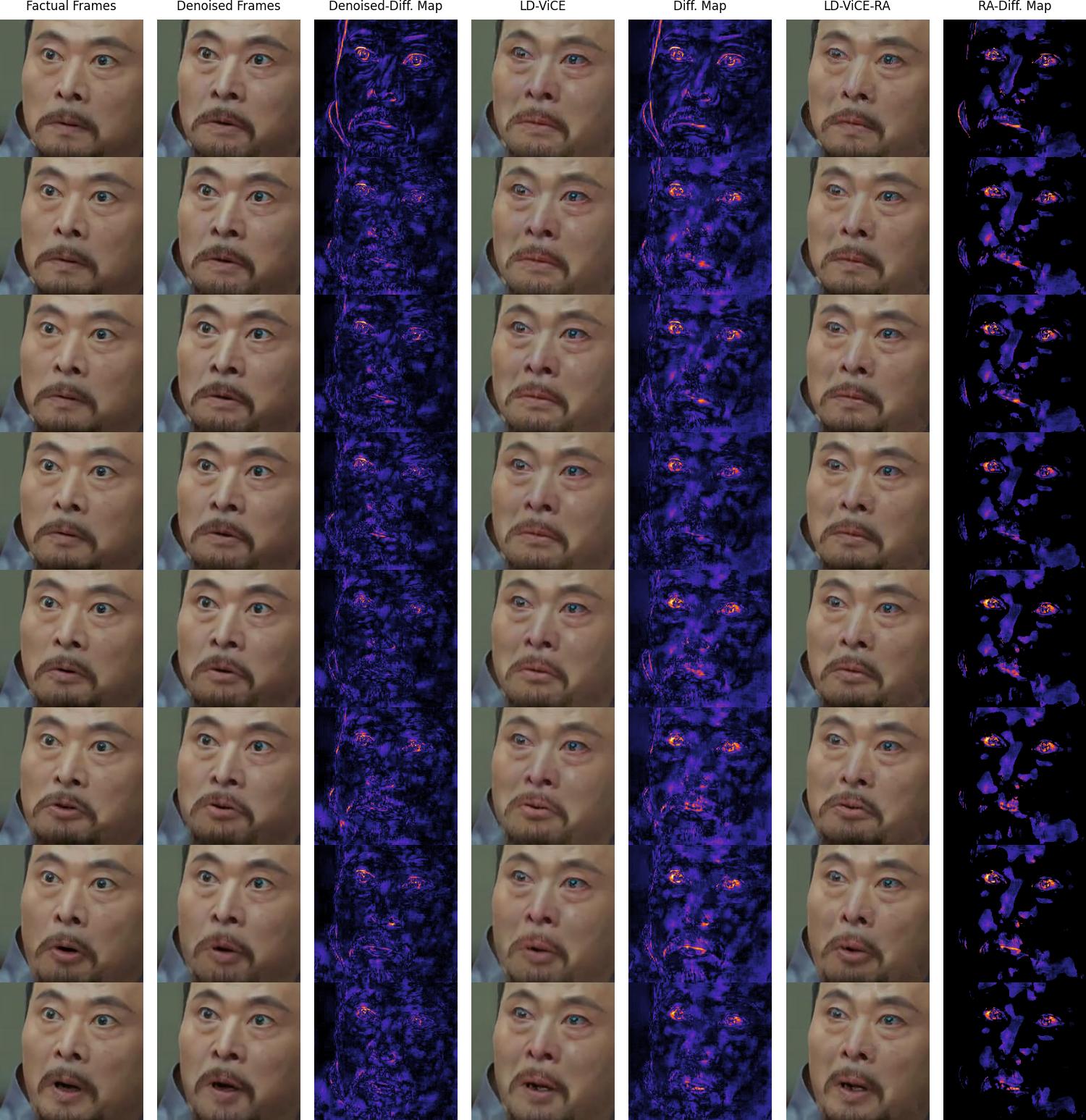} 
    \caption{
        Qualitative counterfactual results on the FERV39K dataset. 
        Factual and denoised frames are shown with difference maps (Denoised-Diff.\ Map) visualizing diffusion-induced changes. 
        Classifier-guided counterfactuals generated by \frameworkAcronym{} introduce facial adjustments, such as reduced brow tension, lowered mouth corners, and tearful eye aligned with the \textit{Sad} emotion, highlighted in the difference maps (Diff.\ Map). 
        The RA variant suppresses high-frequency artifacts and yields cleaner counterfactuals, with difference maps (RA-Diff.\ Map) showing only the salient, class-relevant modifications.
    }
    \label{fig:video_182_cmp}
\end{figure*}

\begin{figure*}[h]
    \centering
    \includegraphics[width=\linewidth]{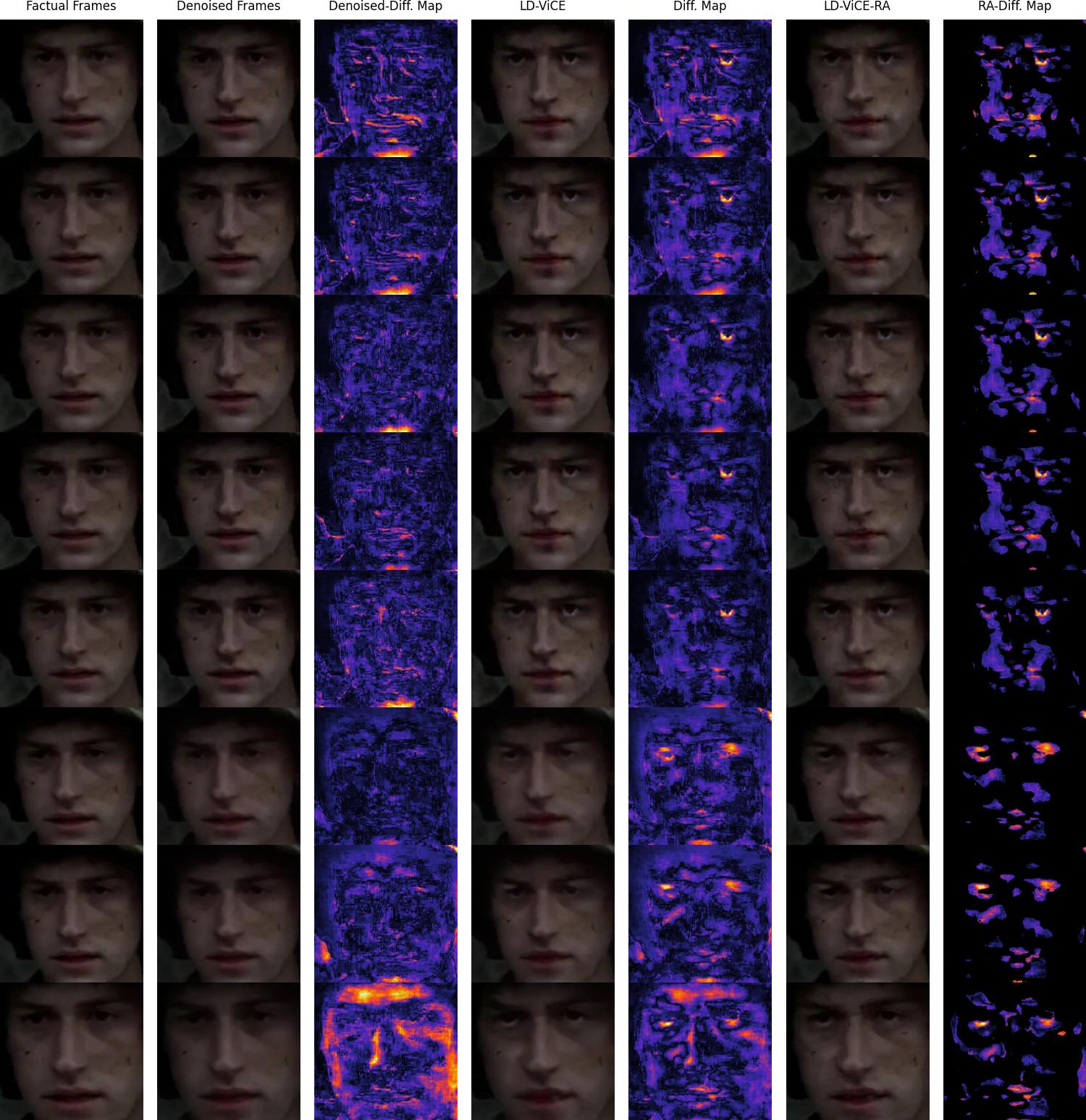} 
    \caption{
        Qualitative counterfactual results on the FERV39K dataset. 
        Factual and denoised frames are shown with difference maps (Denoised-Diff.\ Map) visualizing diffusion-induced changes. 
        Classifier-guided counterfactuals generated by \frameworkAcronym{} introduce facial adjustments, such as widened, more intense eyes, stronger brow contraction, and increased tightening around the mouth aligned with the \textit{Angry} emotion, highlighted in the difference maps (Diff.\ Map). 
        The RA variant suppresses high-frequency artifacts and yields cleaner counterfactuals, with difference maps (RA-Diff.\ Map) showing only the salient, class-relevant modifications.
    }
    \label{fig:video_723_cmp}
\end{figure*}

\cref{fig:video_12_cmp,fig:video_182_cmp,fig:video_723_cmp} illustrate qualitative counterfactual examples generated by \frameworkAcronym{} and its RA variant on the FERV39K dataset. 
The first two columns show the factual frames and their guidance-free denoised versions, followed by difference maps (Denoised-Difference Map) that highlight changes introduced purely by the diffusion denoising process. 
These maps reveal minor, spatially diffuse variations that arise even without classifier guidance.
The subsequent columns display counterfactual frames produced by \frameworkAcronym{}. 
The output frames introduce facial modifications consistent with the target emotion, such as changes around the eyes, eyebrows, and mouth region. 
These targeted adjustments are highlighted in the difference maps (Difference Map), which show coherent, localized changes rather than widespread noise.
The final two columns show counterfactuals refined using the RA variant and its corresponding difference map. 
By suppressing low-magnitude updates below the artifact-suppression threshold, the RA variant removes high-frequency denoising artifacts while preserving expression-relevant structure. 
The difference maps (RA-Difference Map) confirm that only salient, class-discriminative modifications remain, with irrelevant diffusion-induced variations effectively eliminated.

\cref{fig:video_12_cmp} shows the counterfactual transformation shifts the predicted emotion from \textit{Neutral} to \textit{happy}.
The generated samples exhibit class‑specific facial modifications, such as a subtle upward lift of the cheeks and a softening of the mouth curvature, consistent with features typically associated with happiness.
\cref{fig:video_182_cmp} illustrates another transformation set that changes the predicted emotion from \textit{angry} to \textit{sad}.
Here, the modifications reduce the intensity of the brow furrow, relax the tension in the upper eyelid, and lower the corners of the mouth to convey a sense of sadness.
\cref{fig:video_723_cmp} illustrates a transformation from the emotion \textit{Fear} to \textit{Angry}. 
The counterfactuals introduce features characteristic of anger, such as widened, more intense eyes, stronger brow contraction, and increased tightening around the mouth. 
Overall, the results demonstrate that \frameworkAcronym{} generates realistic class-specific transformations, and that the RA refinement produces visually cleaner and more interpretable counterfactuals.

\begin{figure*}[h]
    \centering
    \includegraphics[width=\linewidth]{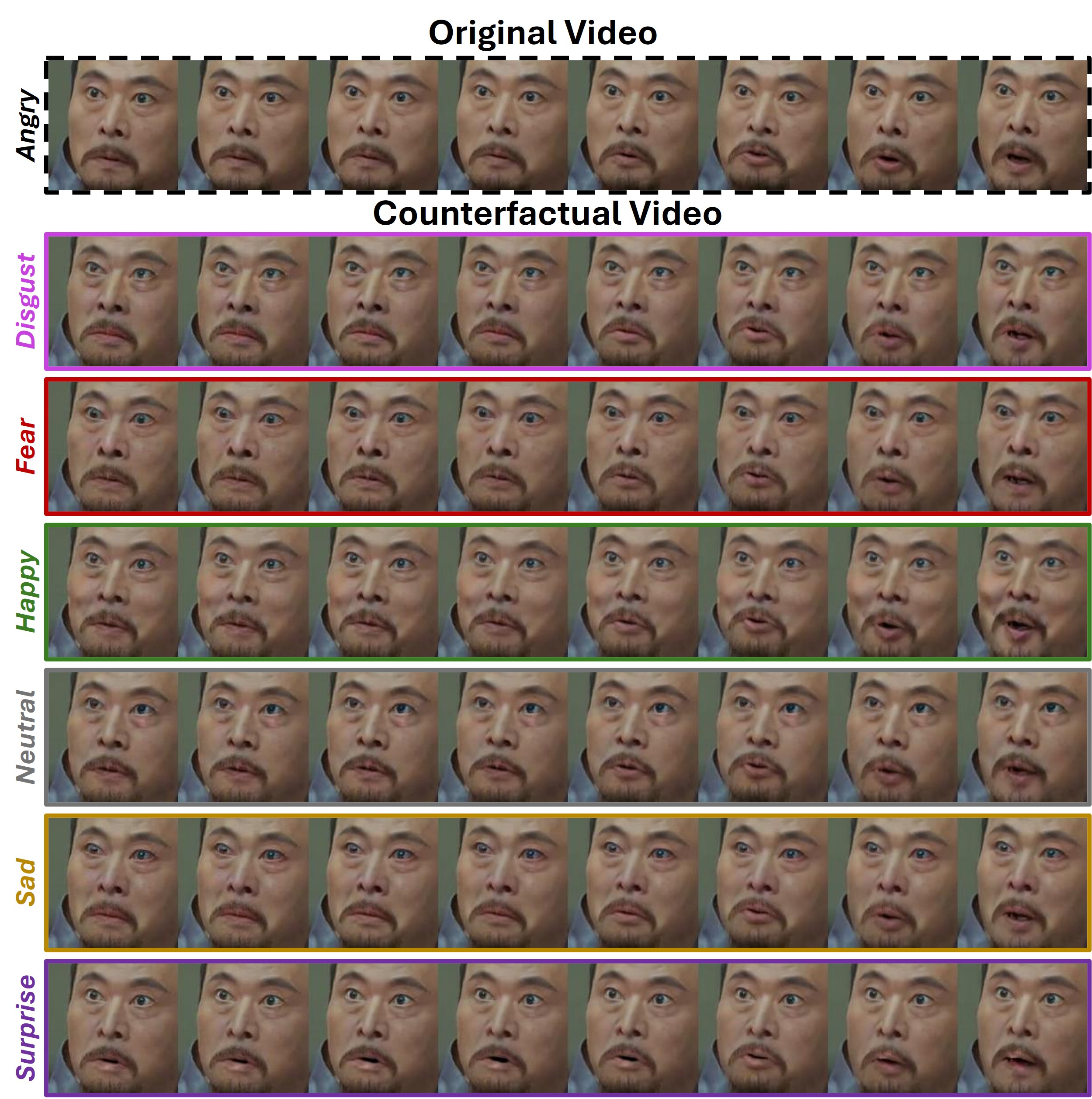} 
    \caption{
        Qualitative counterfactual results generated by \frameworkAcronym{} on the FERV39K dataset, transforming the predicted class emotion \textit{Angry} into each of the other emotion classes. The generated counterfactuals display distinct, class‑consistent facial dynamics that align with the target emotional categories.
    }    
    \label{fig:ferv39k_sample182}
\end{figure*}

\begin{figure*}[h]
    \centering
    \includegraphics[width=\linewidth]{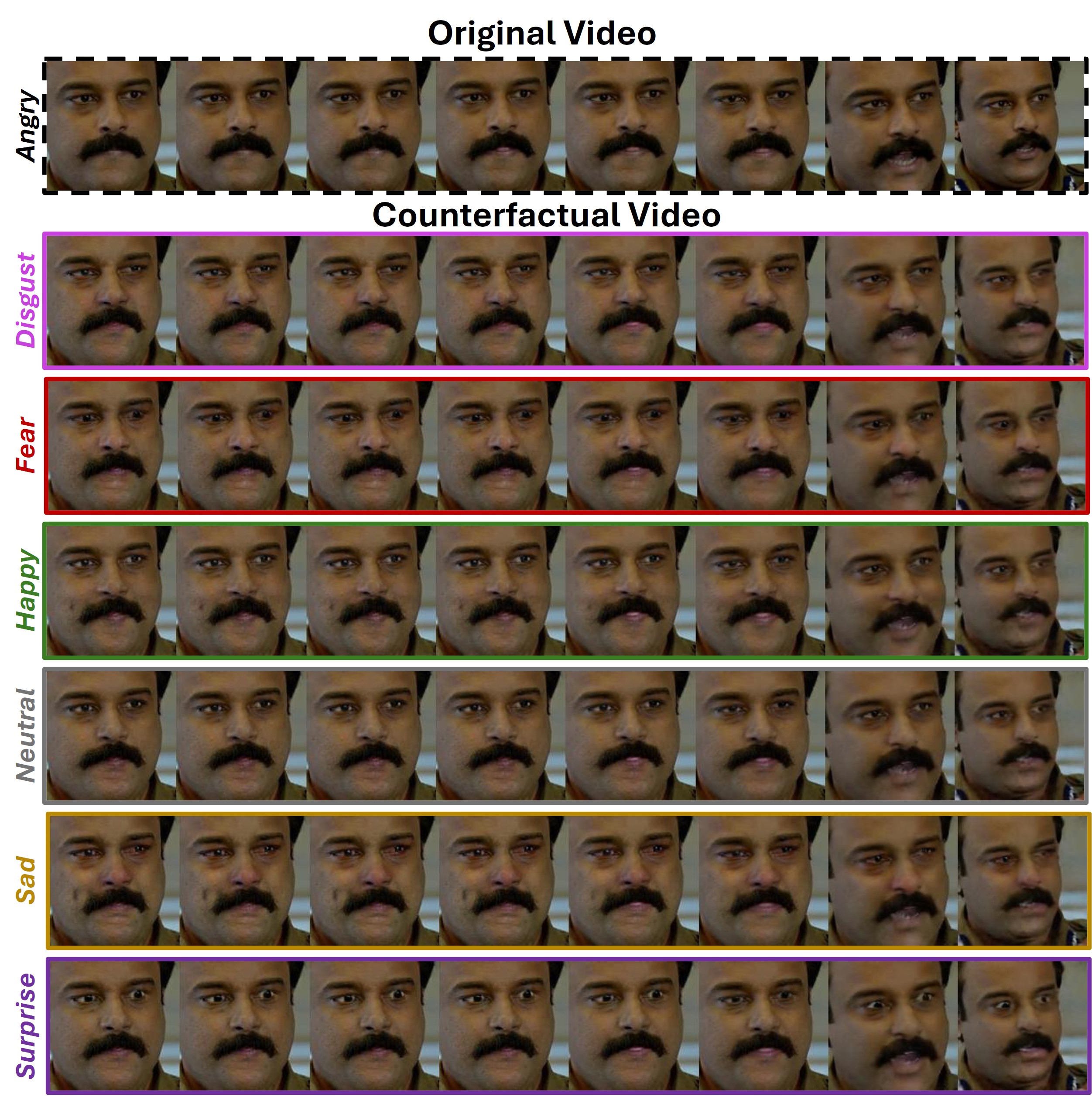} %ferv39k_sample370
    \caption{
        Qualitative counterfactual results generated by \frameworkAcronym\ on the FERV39k dataset, transforming the predicted class emotion \textit{Angry} into each of the other emotion classes. The generated counterfactuals exhibit distinct and class-consistent facial dynamics corresponding to the desired emotional categories. 
    }
    \label{fig:ferv39k_sample370}
\end{figure*}

\begin{figure*}[h]
    \centering
    \includegraphics[width=\linewidth]{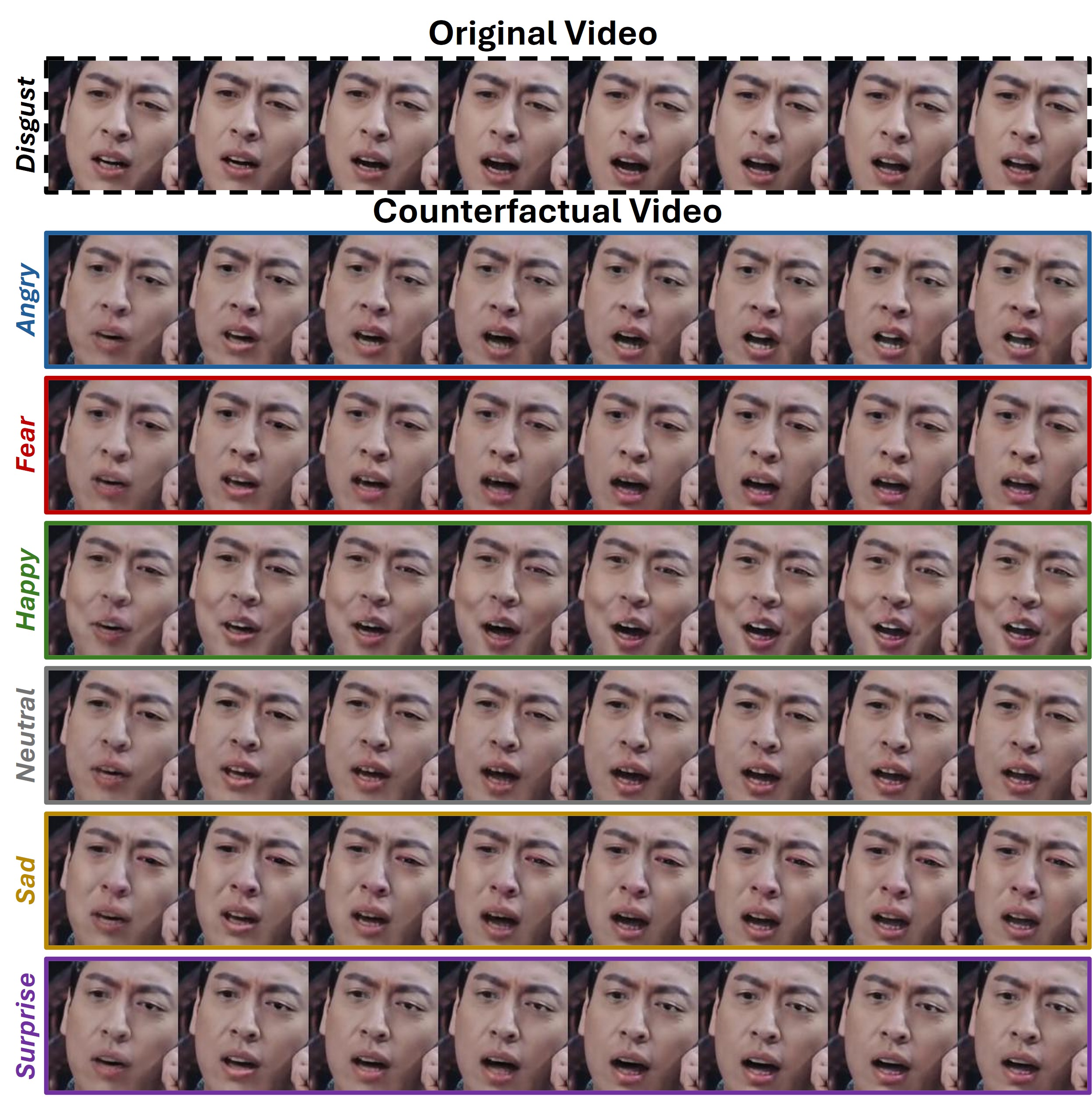} 
    \caption{
        Qualitative counterfactual results generated by \frameworkAcronym\ on the FERV39k dataset, transforming the predicted class emotion \textit{Disgust} into each of the other emotion classes. The generated counterfactuals exhibit distinct and class-consistent facial dynamics corresponding to the desired emotional categories. 
    }
    \label{fig:ferv39k_sample3}
\end{figure*}

\begin{figure*}[h]
    \centering
    \includegraphics[width=\linewidth]{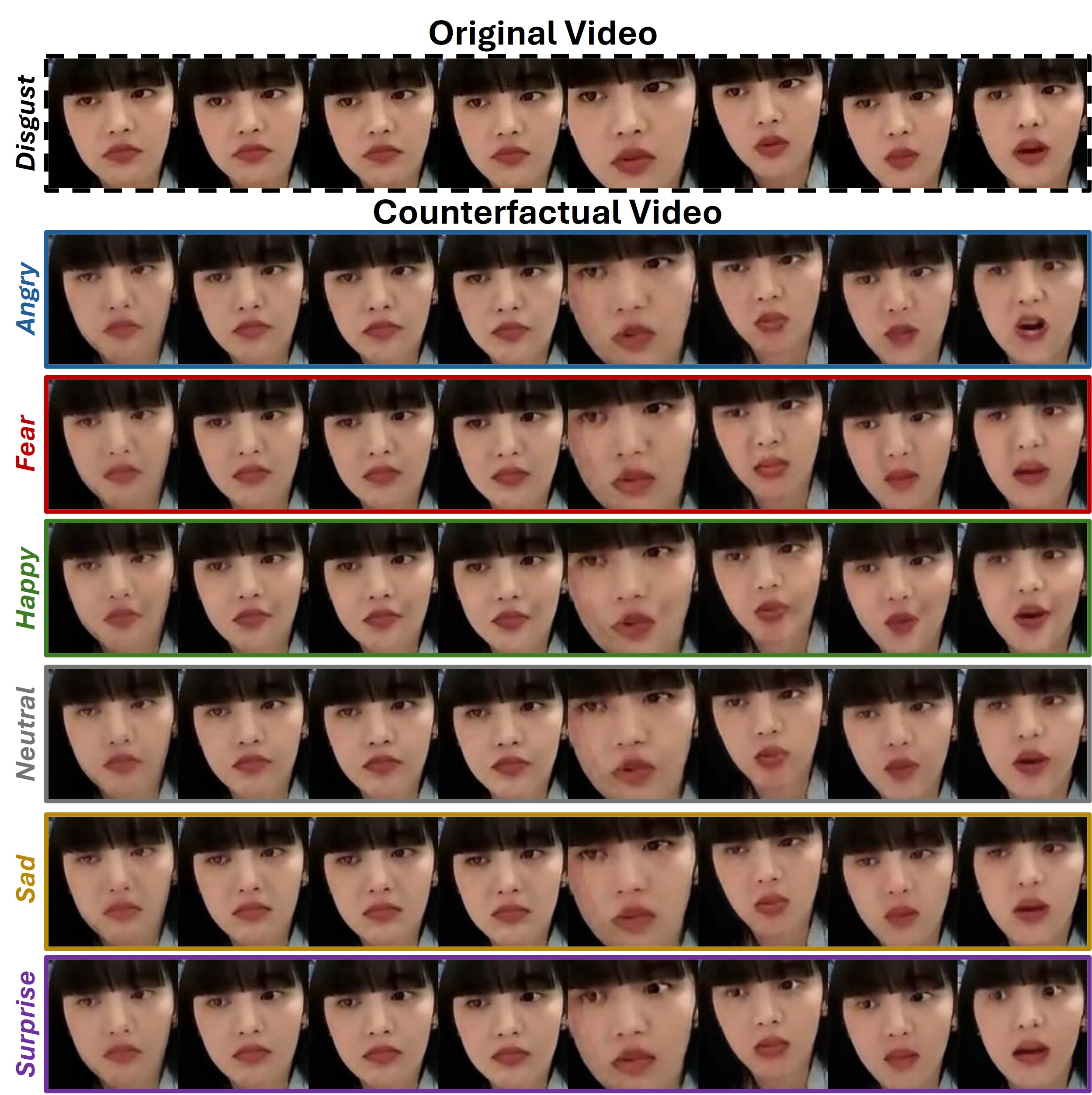} 
    \caption{
        Qualitative counterfactual results generated by \frameworkAcronym\ on the FERV39k dataset, transforming the predicted class emotion \textit{Disgust} into each of the other emotion classes. The generated counterfactuals exhibit distinct and class-consistent facial dynamics corresponding to the desired emotional categories. 
    }
    \label{fig:ferv39k_sample47}
\end{figure*}

\begin{figure*}[h]
    \centering
    \includegraphics[width=\linewidth]{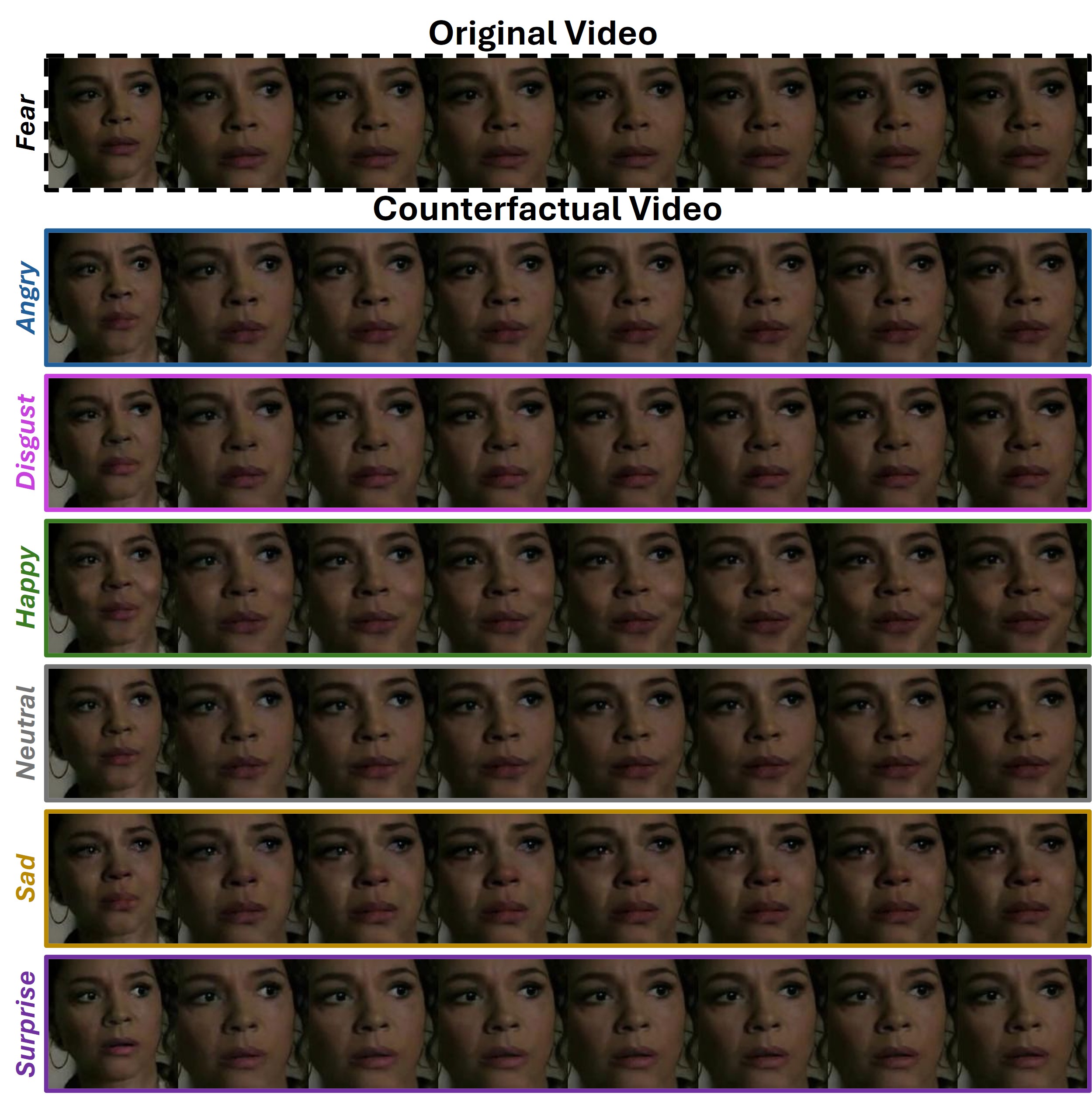} 
    \caption{
        Qualitative counterfactual results generated by \frameworkAcronym\ on the FERV39k dataset, transforming the predicted class emotion \textit{Fear} into each of the other emotion classes. The generated counterfactuals exhibit distinct and class-consistent facial dynamics corresponding to the desired emotional categories. 
    }
    \label{fig:ferv39k_sample248}
\end{figure*}

\begin{figure*}[h]
    \centering
    \includegraphics[width=\linewidth]{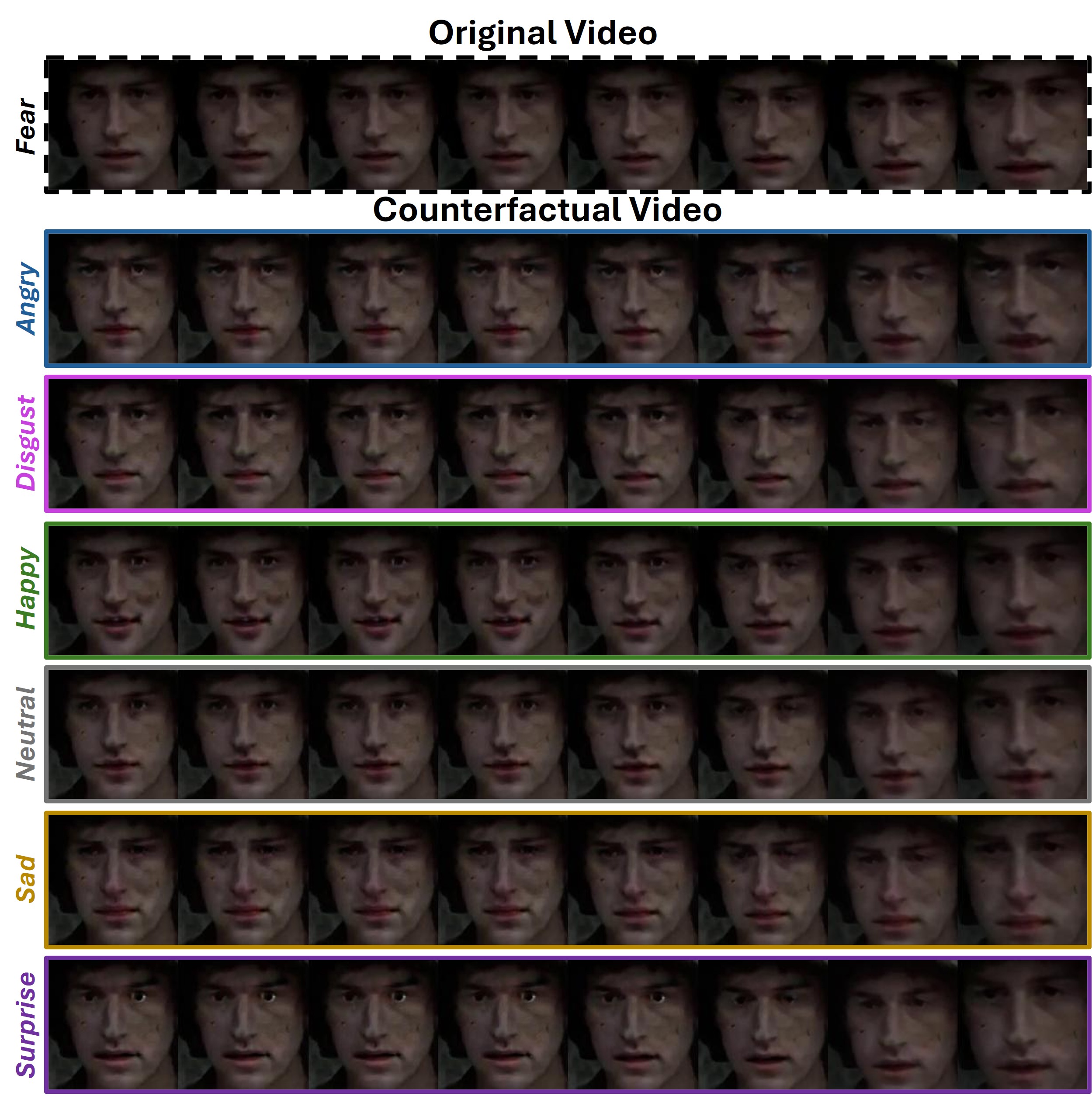} 
    \caption{
        Qualitative counterfactual results generated by \frameworkAcronym\ on the FERV39k dataset, transforming the predicted class emotion \textit{Fear} into each of the other emotion classes. The generated counterfactuals exhibit distinct and class-consistent facial dynamics corresponding to the desired emotional categories. 
    }
    \label{fig:ferv39k_sample723}
\end{figure*}

\begin{figure*}[h]
    \centering
    \includegraphics[width=\linewidth]{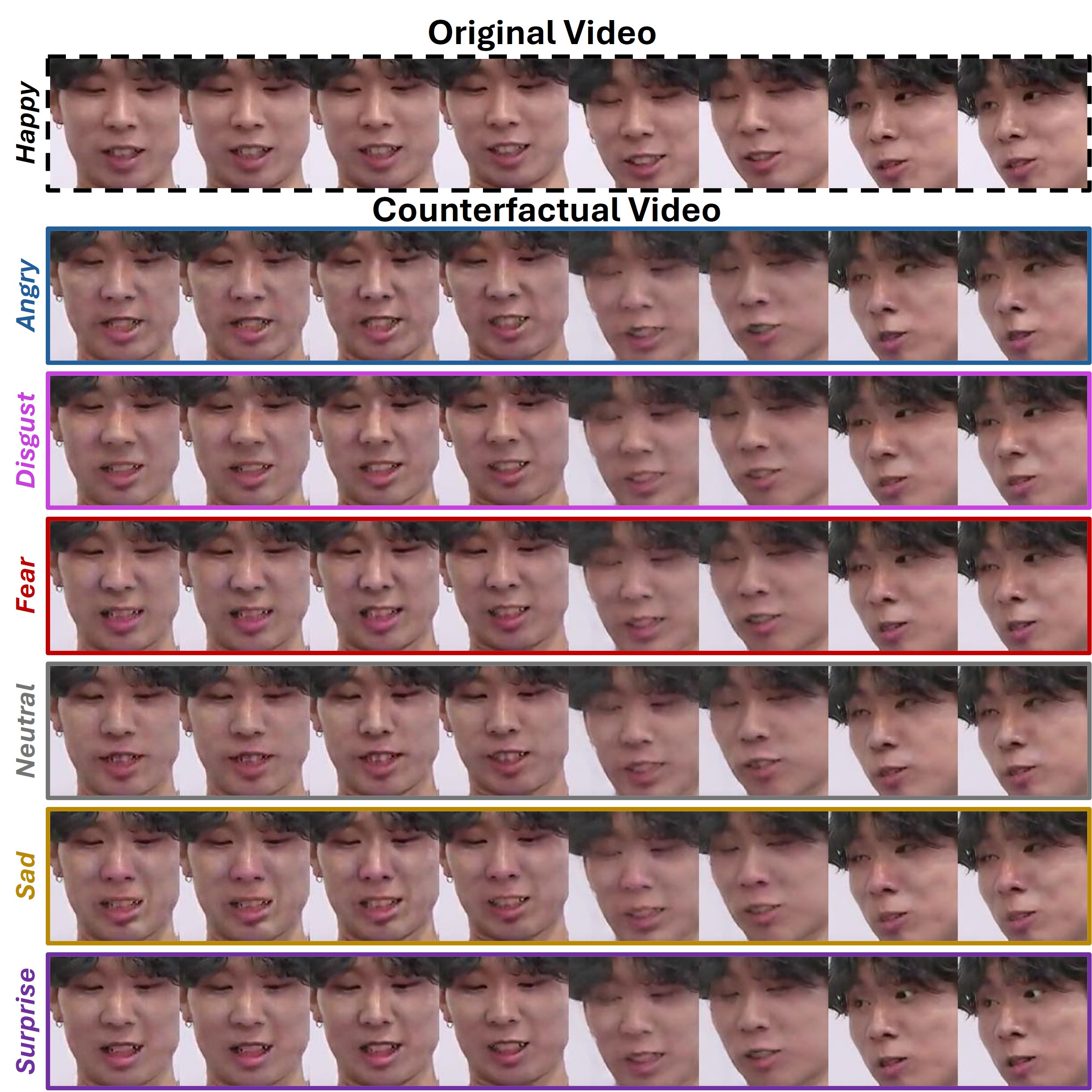} 
    \caption{
        Qualitative counterfactual results generated by \frameworkAcronym\ on the FERV39k dataset, transforming the predicted class emotion \textit{Happy} into each of the other emotion classes. The generated counterfactuals exhibit distinct and class-consistent facial dynamics corresponding to the desired emotional categories. 
    }
    \label{fig:ferv39k_sample11}
\end{figure*}

\begin{figure*}[h]
    \centering
    \includegraphics[width=\linewidth]{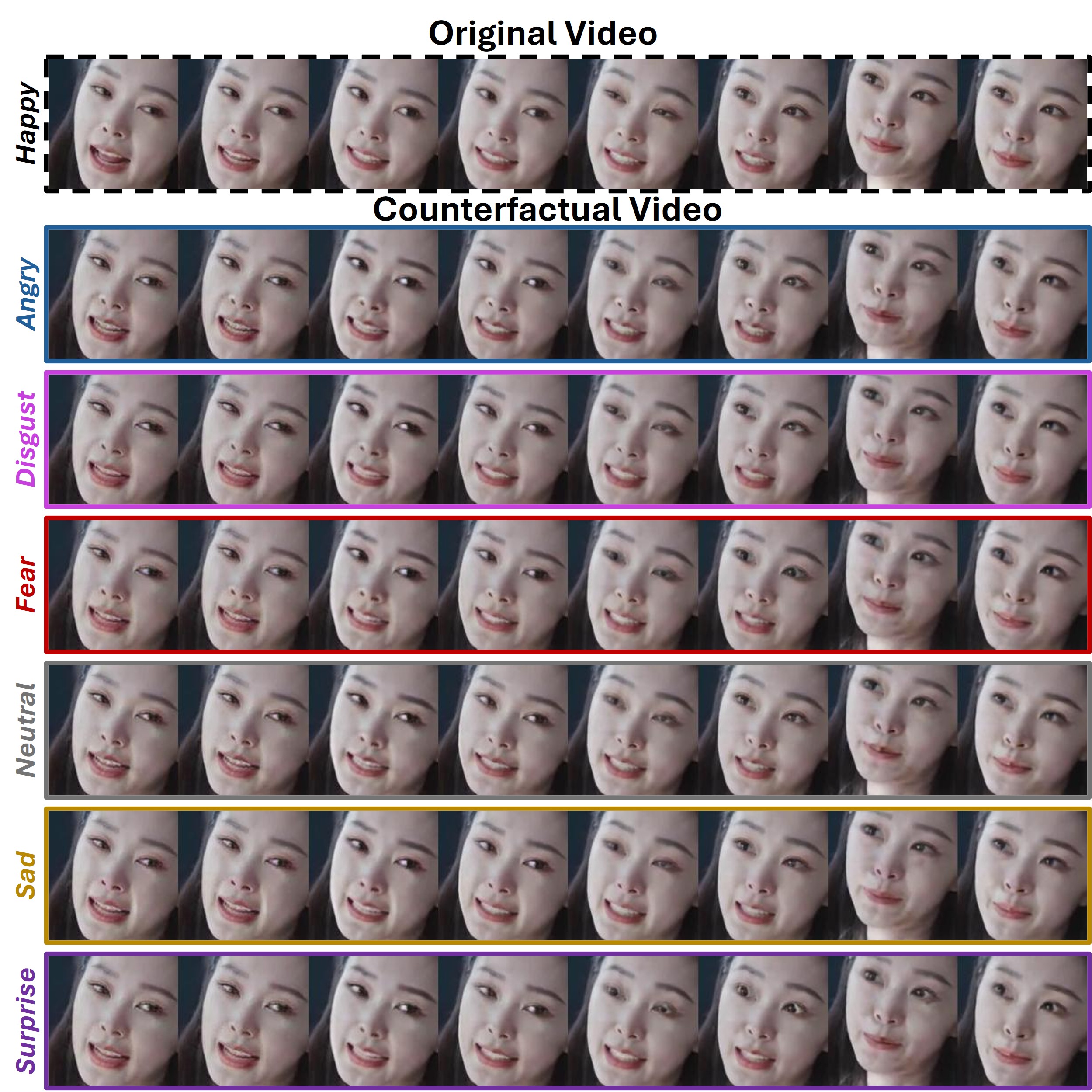} 
    \caption{
        Qualitative counterfactual results generated by \frameworkAcronym\ on the FERV39k dataset, transforming the predicted class emotion \textit{Happy} into each of the other emotion classes. The generated counterfactuals exhibit distinct and class-consistent facial dynamics corresponding to the desired emotional categories. 
    }
    \label{fig:ferv39k_sample82}
\end{figure*}

\begin{figure*}[h]
    \centering
    \includegraphics[width=\linewidth]{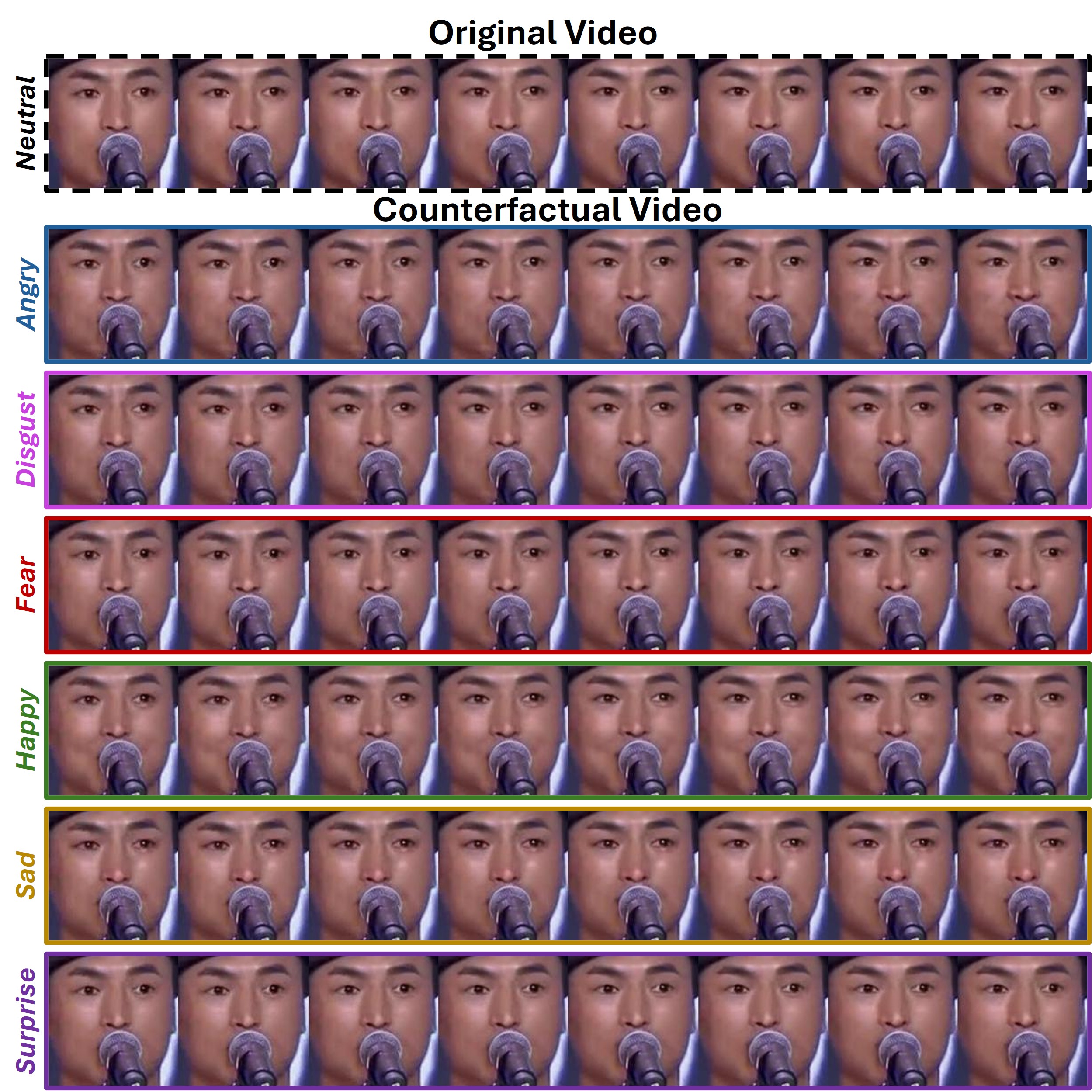} 
    \caption{
        Qualitative counterfactual results generated by \frameworkAcronym\ on the FERV39k dataset, transforming the predicted class emotion \textit{Neutral} into each of the other emotion classes. The generated counterfactuals exhibit distinct and class-consistent facial dynamics corresponding to the desired emotional categories. 
    }
    \label{fig:ferv39k_sample12}
\end{figure*}

\begin{figure*}[h]
    \centering
    \includegraphics[width=\linewidth]{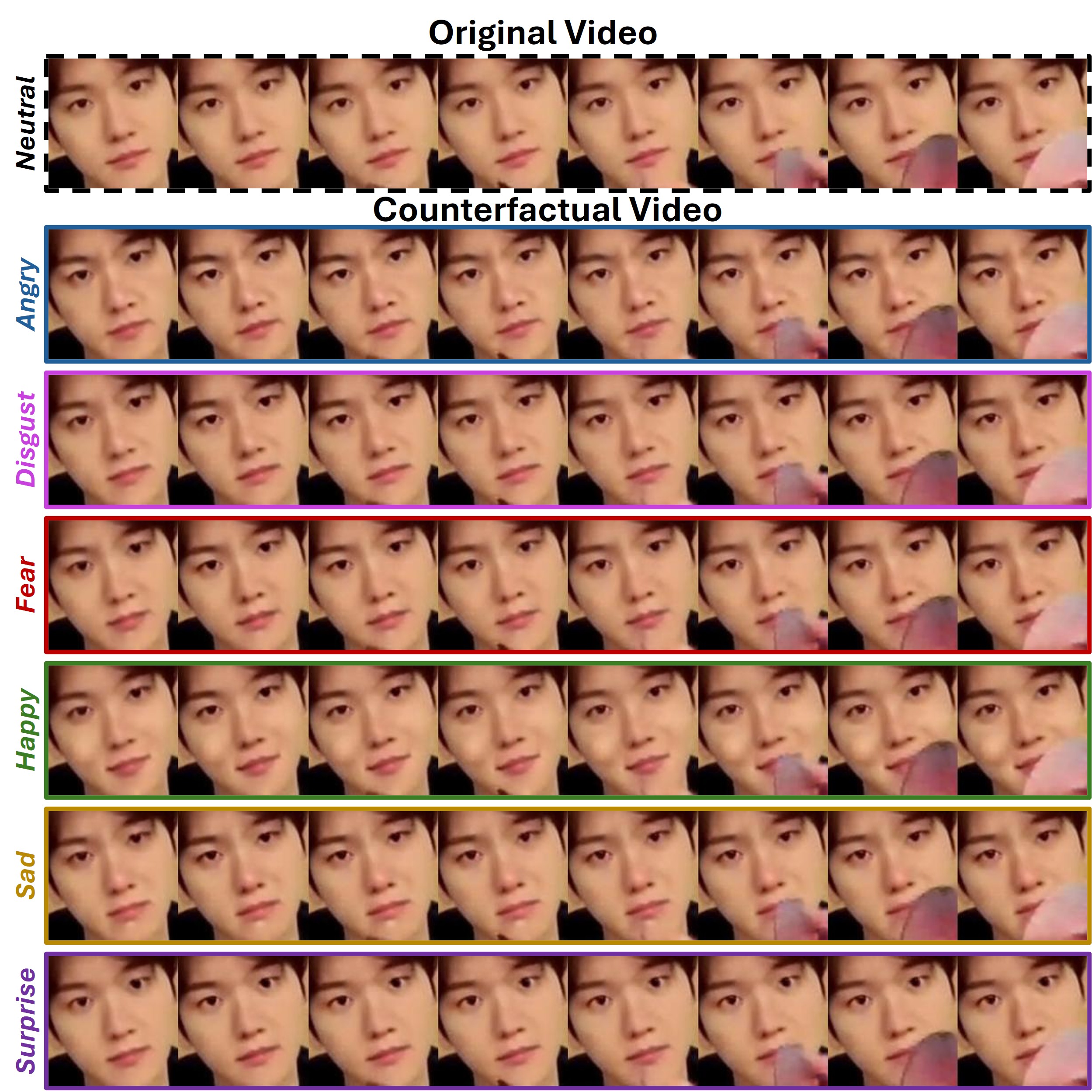} 
    \caption{
        Qualitative counterfactual results generated by \frameworkAcronym\ on the FERV39k dataset, transforming the predicted class emotion \textit{Neutral} into each of the other emotion classes. The generated counterfactuals exhibit distinct and class-consistent facial dynamics corresponding to the desired emotional categories. 
    }
    \label{fig:ferv39k_sample83}
\end{figure*}

\begin{figure*}[h]
    \centering
    \includegraphics[width=\linewidth]{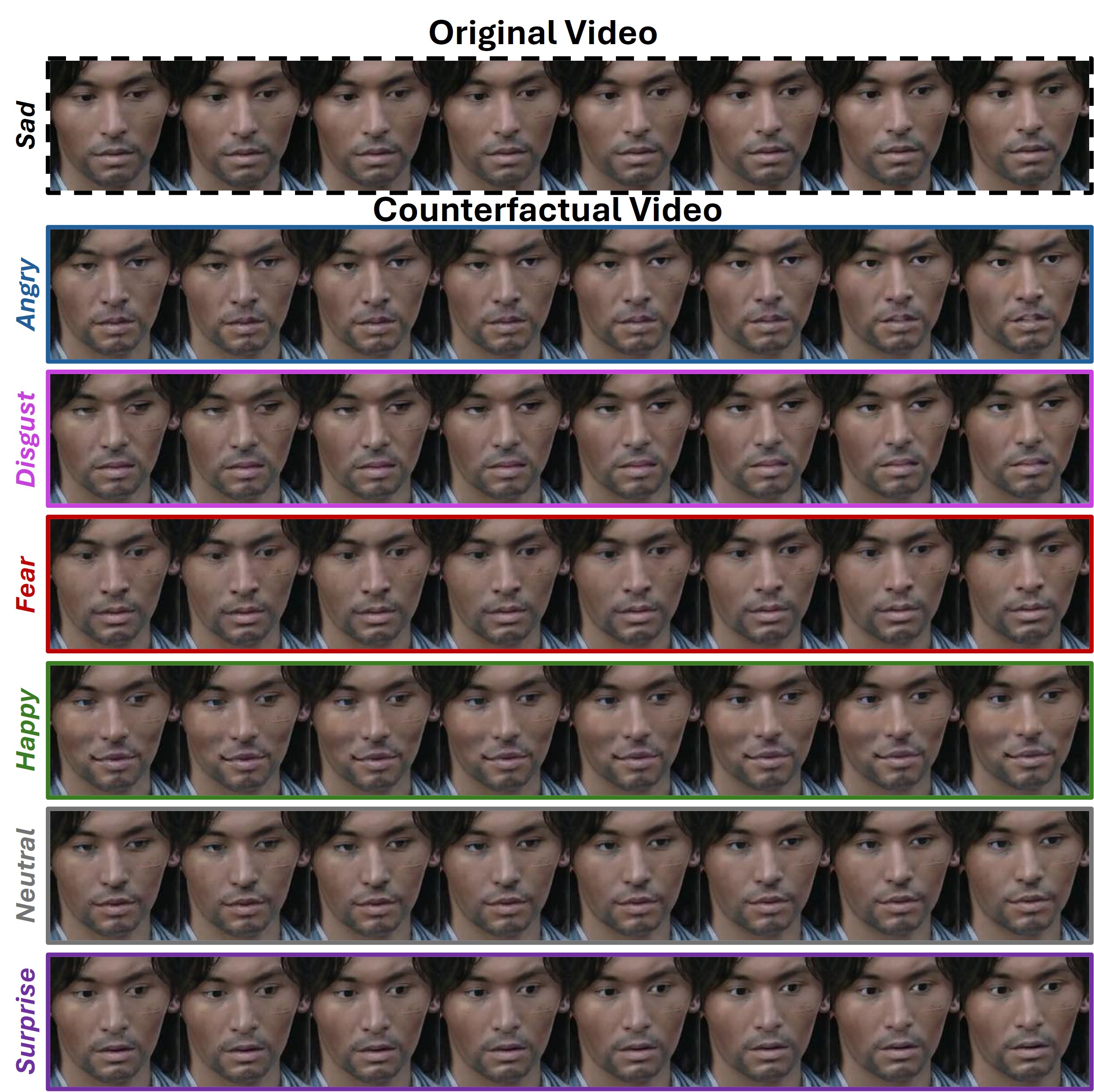} 
    \caption{
        Qualitative counterfactual results generated by \frameworkAcronym\ on the FERV39k dataset, transforming the predicted class emotion \textit{Sad} into each of the other emotion classes. The generated counterfactuals exhibit distinct and class-consistent facial dynamics corresponding to the desired emotional categories. 
    }
    \label{fig:ferv39k_sample5}
\end{figure*}

\begin{figure*}[h]
    \centering
    \includegraphics[width=\linewidth]{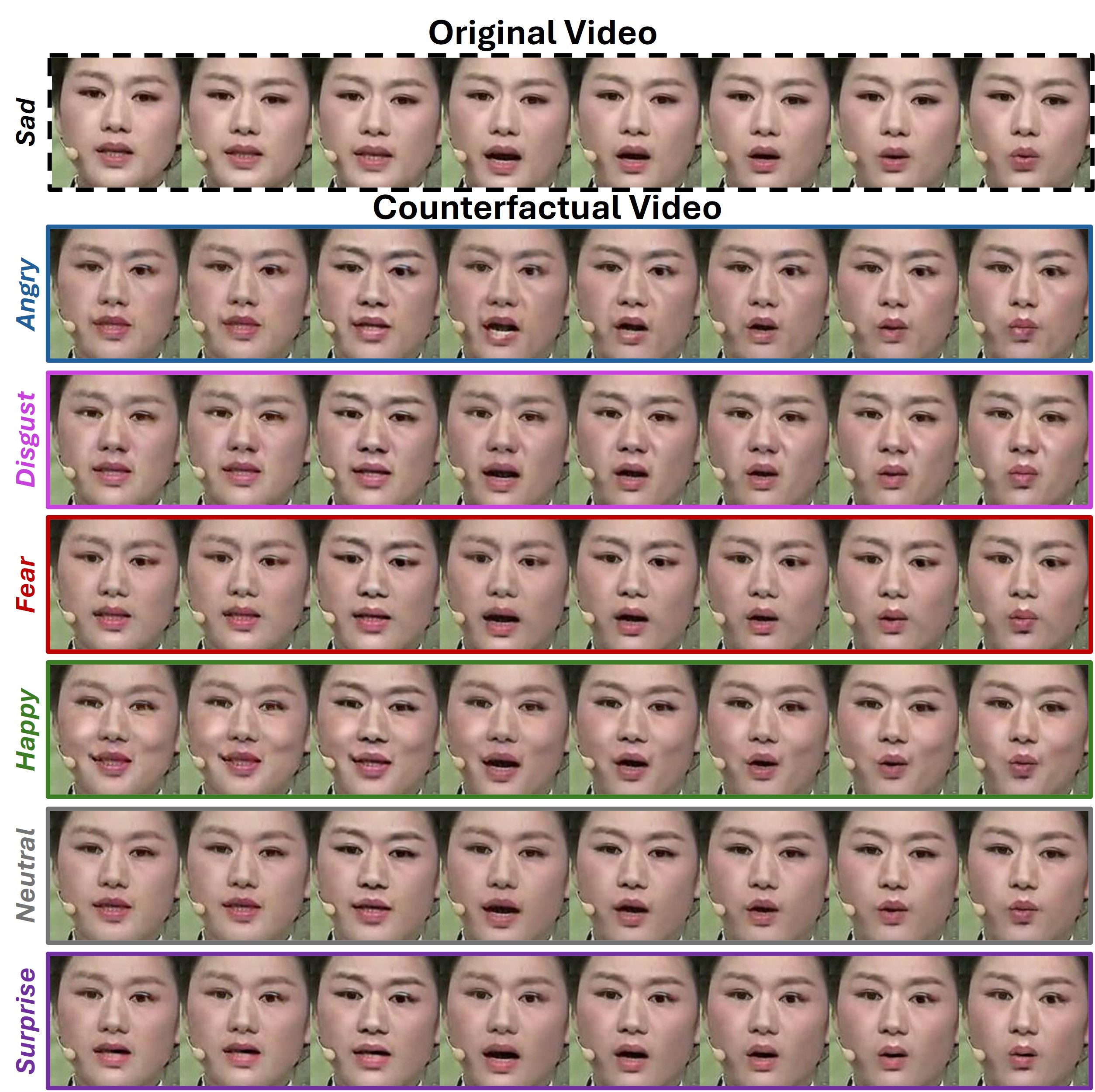} 
    \caption{
        Qualitative counterfactual results generated by \frameworkAcronym\ on the FERV39k dataset, transforming the predicted class emotion \textit{Sad} into each of the other emotion classes. The generated counterfactuals exhibit distinct and class-consistent facial dynamics corresponding to the desired emotional categories. 
    }
    \label{fig:ferv39k_sample63}
\end{figure*}

\begin{figure*}[h]
    \centering
    \includegraphics[width=\linewidth]{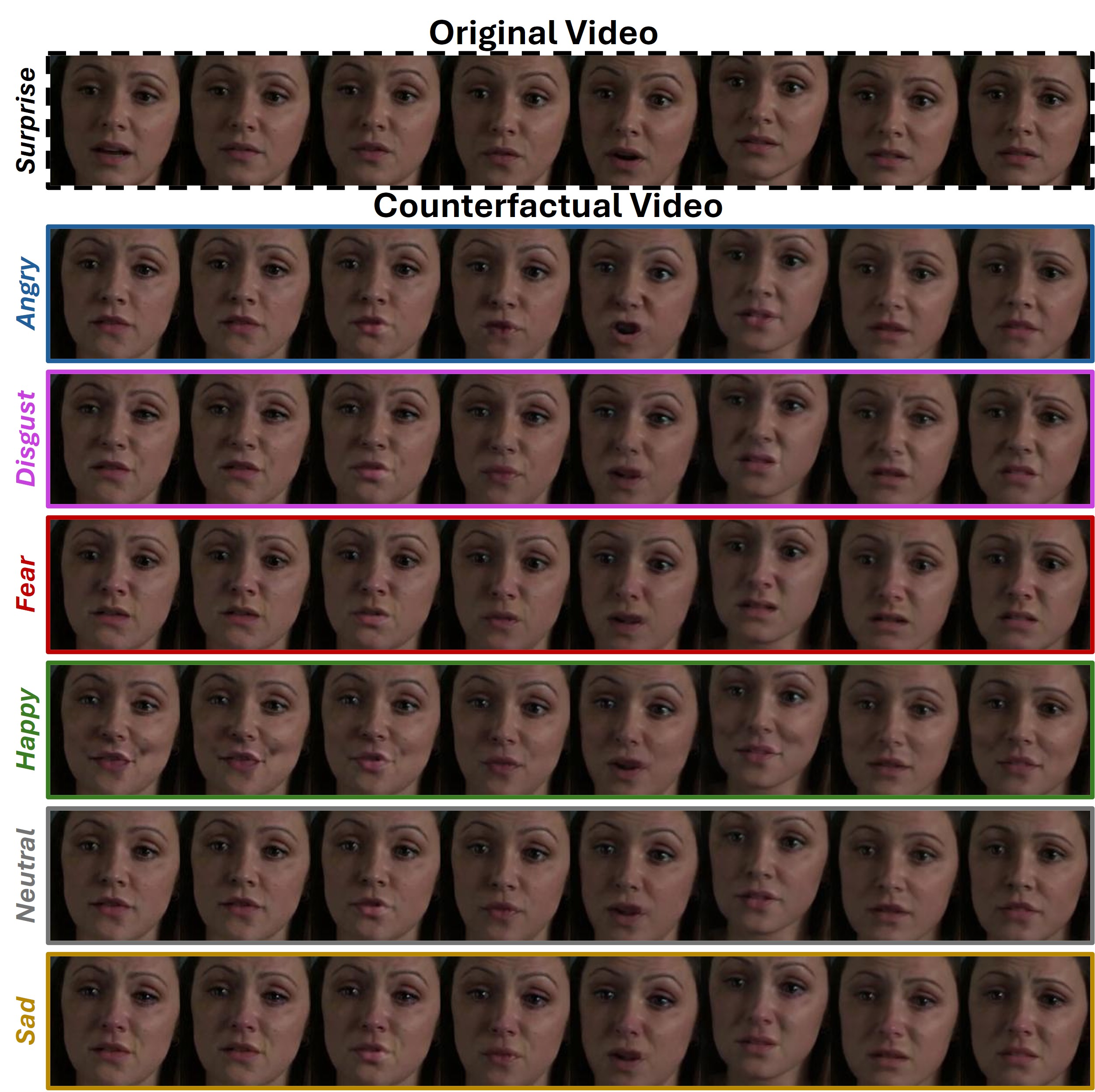} 
    \caption{
        Qualitative counterfactual results generated by \frameworkAcronym\ on the FERV39k dataset, transforming the predicted class emotion \textit{Surprise} into each of the other emotion classes. The generated counterfactuals exhibit distinct and class-consistent facial dynamics corresponding to the desired emotional categories. 
    }
    \label{fig:ferv39k_sample1}
\end{figure*}

\begin{figure*}[h]
    \centering
    \includegraphics[width=\linewidth]{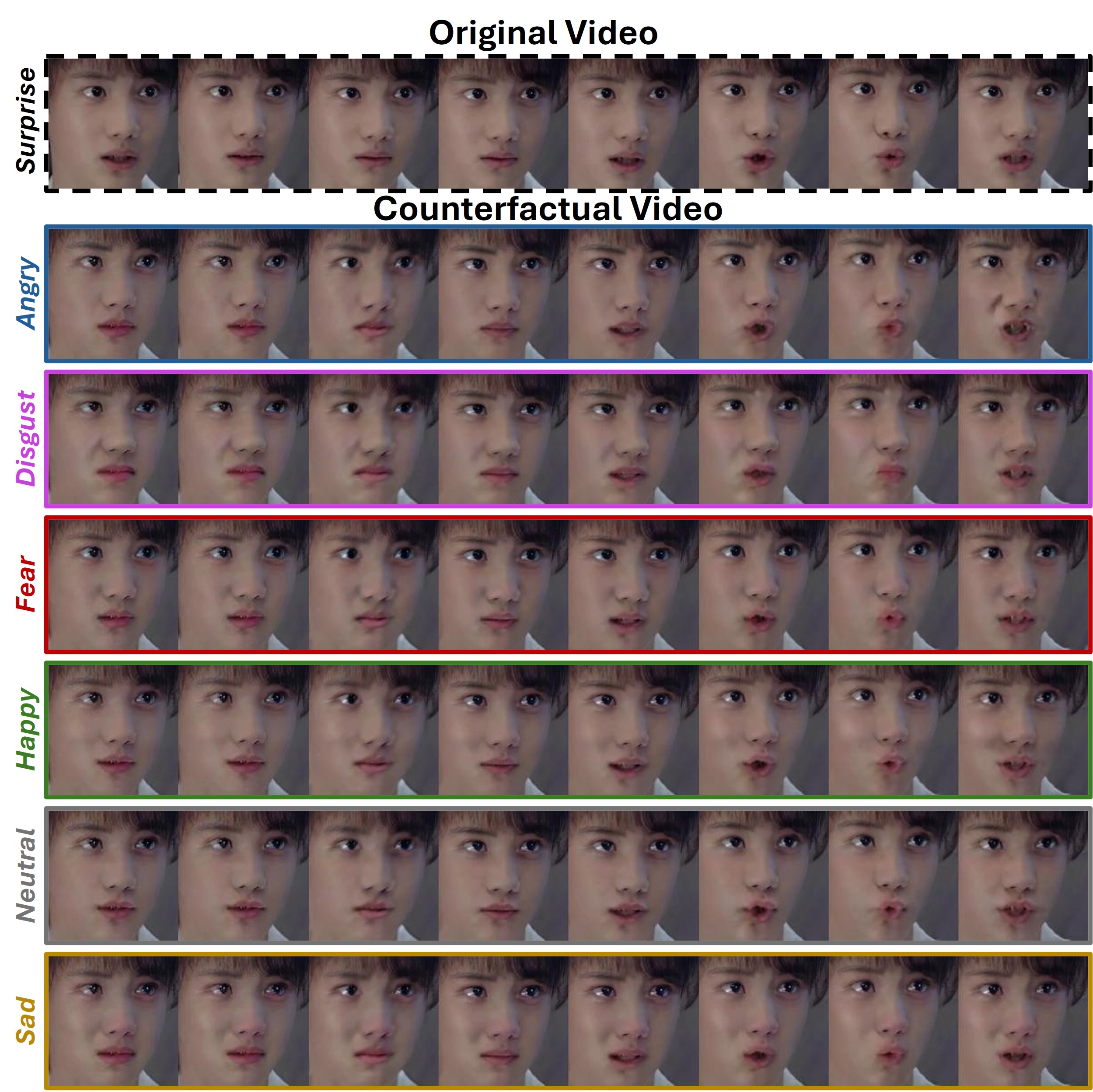} 
    \caption{
        Qualitative counterfactual results generated by \frameworkAcronym\ on the FERV39k dataset, transforming the predicted class emotion \textit{Surprise} into each of the other emotion classes. The generated counterfactuals exhibit distinct and class-consistent facial dynamics corresponding to the desired emotional categories.
    }
    \label{fig:ferv39k_sample41}
\end{figure*}

In addition to these examples, counterfactual results are generated for two representative samples from each of the seven emotion classes, with each sample transformed into all other non‑predicted target classes, as shown in \cref{fig:ferv39k_sample182} to \ref{fig:ferv39k_sample41}. The subsequent figures demonstrate \frameworkAcronym{}’s ability to produce semantically meaningful modifications across a diverse range of source–target emotion transitions.

%%%%%%%%%%%%%%%%%%%%%%%%%%%%%%%%%%%%%%%%%%%%%%%%%%%%%%%%%%%%%%%%%%%%%%%%%%%%%%%%%%%%%%%%%%%%%%%%%%%%%%%%%%%%%%%%%%%%%%%%%%%%%%%%%%%%%%%%%%%%%%%%%%%%%%%%%%%%%%%%%%%%%%%%%%%%%%%%%%%%%%%%%%%%%%%%%%%%%%%%%%%%%%%%%%%%%%%%%%%%%%%%%%%%%%%%%%%%%%%%%%%%%%%%%%%%%%%%%%%%%%%%%%%%%%%%%%%%%%
% class 0, Angry

%%%%%%%%%%%%%%%%%%%%%%%%%%%%%%%%%%%%%%%%%%%%%%%%%%%%%%%%%%%%%%%%%%%%%%%%%%%%%%%%%%%%%%%%%%%%%%%%%%%%%%%%%%%%%%%%%%%%%%%%%%%%%%%%%%%%%%%%%%%%%%%%%%%%%%%%%%%%%%%%%%%%%%%%%%%%%%%%%%%%%%%%%%%%%%%%%%%%%%%%%%%%%%%%%%%%%%%%%%%%%%%%%%%%%%%%%%%%%%%%%%%%%%%%%%%%%%%%%%%%%%%%%%%%%%%%%%%%%%

% class 1, Disgust

%%%%%%%%%%%%%%%%%%%%%%%%%%%%%%%%%%%%%%%%%%%%%%%%%%%%%%%%%%%%%%%%%%%%%%%%%%%%%%%%%%%%%%%%%%%%%%%%%%%%%%%%%%%%%%%%%%%%%%%%%%%%%%%%%%%%%%%%%%%%%%%%%%%%%%%%%%%%%%%%%%%%%%%%%%%%%%%%%%%%%%%%%%%%%%%%%%%%%%%%%%%%%%%%%%%%%%%%%%%%%%%%%%%%%%%%%%%%%%%%%%%%%%%%%%%%%%%%%%%%%%%%%%%%%%%%%%%%%%
% class 2, Fear

%%%%%%%%%%%%%%%%%%%%%%%%%%%%%%%%%%%%%%%%%%%%%%%%%%%%%%%%%%%%%%%%%%%%%%%%%%%%%%%%%%%%%%%%%%%%%%%%%%%%%%%%%%%%%%%%%%%%%%%%%%%%%%%%%%%%%%%%%%%%%%%%%%%%%%%%%%%%%%%%%%%%%%%%%%%%%%%%%%%%%%%%%%%%%%%%%%%%%%%%%%%%%%%%%%%%%%%%%%%%%%%%%%%%%%%%%%%%%%%%%%%%%%%%%%%%%%%%%%%%%%%%%%%%%%%%%%%%%%
% % class 3, Happy

%%%%%%%%%%%%%%%%%%%%%%%%%%%%%%%%%%%%%%%%%%%%%%%%%%%%%%%%%%%%%%%%%%%%%%%%%%%%%%%%%%%%%%%%%%%%%%%%%%%%%%%%%%%%%%%%%%%%%%%%%%%%%%%%%%%%%%%%%%%%%%%%%%%%%%%%%%%%%%%%%%%%%%%%%%%%%%%%%%%%%%%%%%%%%%%%%%%%%%%%%%%%%%%%%%%%%%%%%%%%%%%%%%%%%%%%%%%%%%%%%%%%%%%%%%%%%%%%%%%%%%%%%%%%%%%%%%%%%%

% % class 4, Neutral

%%%%%%%%%%%%%%%%%%%%%%%%%%%%%%%%%%%%%%%%%%%%%%%%%%%%%%%%%%%%%%%%%%%%%%%%%%%%%%%%%%%%%%%%%%%%%%%%%%%%%%%%%%%%%%%%%%%%%%%%%%%%%%%%%%%%%%%%%%%%%%%%%%%%%%%%%%%%%%%%%%%%%%%%%%%%%%%%%%%%%%%%%%%%%%%%%%%%%%%%%%%%%%%%%%%%%%%%%%%%%%%%%%%%%%%%%%%%%%%%%%%%%%%%%%%%%%%%%%%%%%%%%%%%%%%%%%%%%%
% % class 5, Sad

%%%%%%%%%%%%%%%%%%%%%%%%%%%%%%%%%%%%%%%%%%%%%%%%%%%%%%%%%%%%%%%%%%%%%%%%%%%%%%%%%%%%%%%%%%%%%%%%%%%%%%%%%%%%%%%%%%%%%%%%%%%%%%%%%%%%%%%%%%%%%%%%%%%%%%%%%%%%%%%%%%%%%%%%%%%%%%%%%%%%%%%%%%%%%%%%%%%%%%%%%%%%%%%%%%%%%%%%%%%%%%%%%%%%%%%%%%%%%%%%%%%%%%%%%%%%%%%%%%%%%%%%%%%%%%%%%%%%%%
% % class 6, surprise

%%%%%%%%%%%%%%%%%%%%%%%%%%%%%%%%%%%%%%%%%%%%%%%%%%%%%%%%%%%%%%%%%%%%%%%%%%%%%%%%%%%%%%%%%%%%%%%%%%%%%%%%%%%%%%%%%%%%%%%%%%%%%%%%%%%%%%%%%%%%%%%%%%%%%%%%%%%%%%%%%%%%%%%%%%%%%%%%%%%%%%%%%%%%%%%%%%%%%%%%%%%%%%%%%%%%%%%%%%%%%%%%%%%%%%%%%%%%%%%%%%%%%%%%%%%%%%%%%%%%%%%%%%%%%%%%%%%%%%
%%%%%%%%%%%%%%%%%%%%%%%%%%%%%%%%%%%%%%%%%%%%%%%%%%%%%%%%%%%%%%%%%%%%%%%%%%%%%%%%%%%%%%%%%%%%%%%%%%%%%%%%%%%%%%%%%%%%%%%%%%%%%%%%%%%%%%%%%%%%%%%%%%%%%%%%%%%%%%%%%%%%%%%%%%%%%%%%%%%%%%%%%%%%%%%%%%%%%%%%%%%%%%%%%%%%%%%%%%%%%%%%%%%%%%%%%%%%%%%%%%%%%%%%%%%%%%%%%%%%%%%%%%%%%%%%%%%%%%%%%%%%%%%%%%%%%%%%%%%%%%%%%%%%%%%%%%%%%%%%%%%%%%%%%%%%%%%%%%%%%%%%%%%%%%%%%%%%%%%%%%%%%%%%%%%%%%%%%%%%%%%%%%%%%%%%%%%%%%%%%%%%%%%%%%%%%%%%%%%%%%%%%%%%%%%%%%%%%%%%%%%%%%%%%%%%%%%%%%%%%%%%%%%%%%%%%%%%%%%%%%%%%%%%%%%%%%%%%%%%%%%%%%%%%%%%%%%%%%%%%%%%%%%%%%%%%%%%%%%%%%%%%%%%%%%%%%%%
\clearpage
\section{Something- Something V2 Dataset Counterfactual Results}\label{sec:ssv2}

\begin{figure*}[h]
    \centering
    \includegraphics[width=\linewidth]{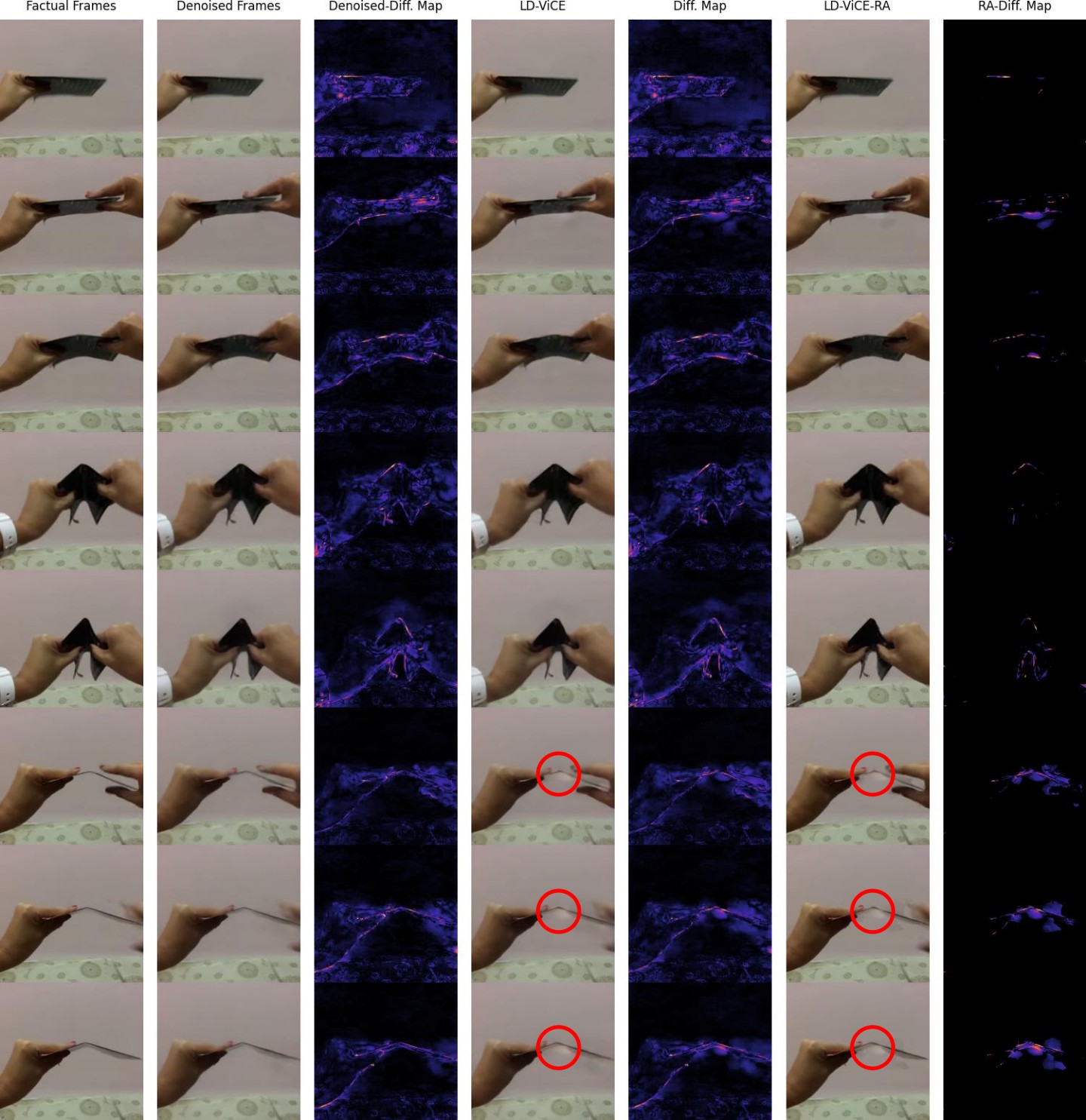} 
    \caption{
            Qualitative counterfactual results for transforming the predicted action from \textit{Bending something so that it deforms} to \textit{Bending something until it breaks} on the SSv2 dataset using \frameworkAcronym{}.
            Factual frames and denoised reconstructions (guidance-free) are shown alongside their denoised difference maps (Denoised–Diff.\ Map).
            \frameworkAcronym{} counterfactuals introduce classifier-driven changes in the interaction region, while the RA variant suppresses high-frequency diffusion artifacts.
            Changes are primarily applied to the expected interaction region, and in the final frames, localized brightening and deformation in the folding area represent a potential break in the object (highlighted in the red circle).
            }    
    \label{fig:ssv2_video_730_cmp}
\end{figure*}

\begin{figure*}[h]
    \centering
    \includegraphics[width=\linewidth]{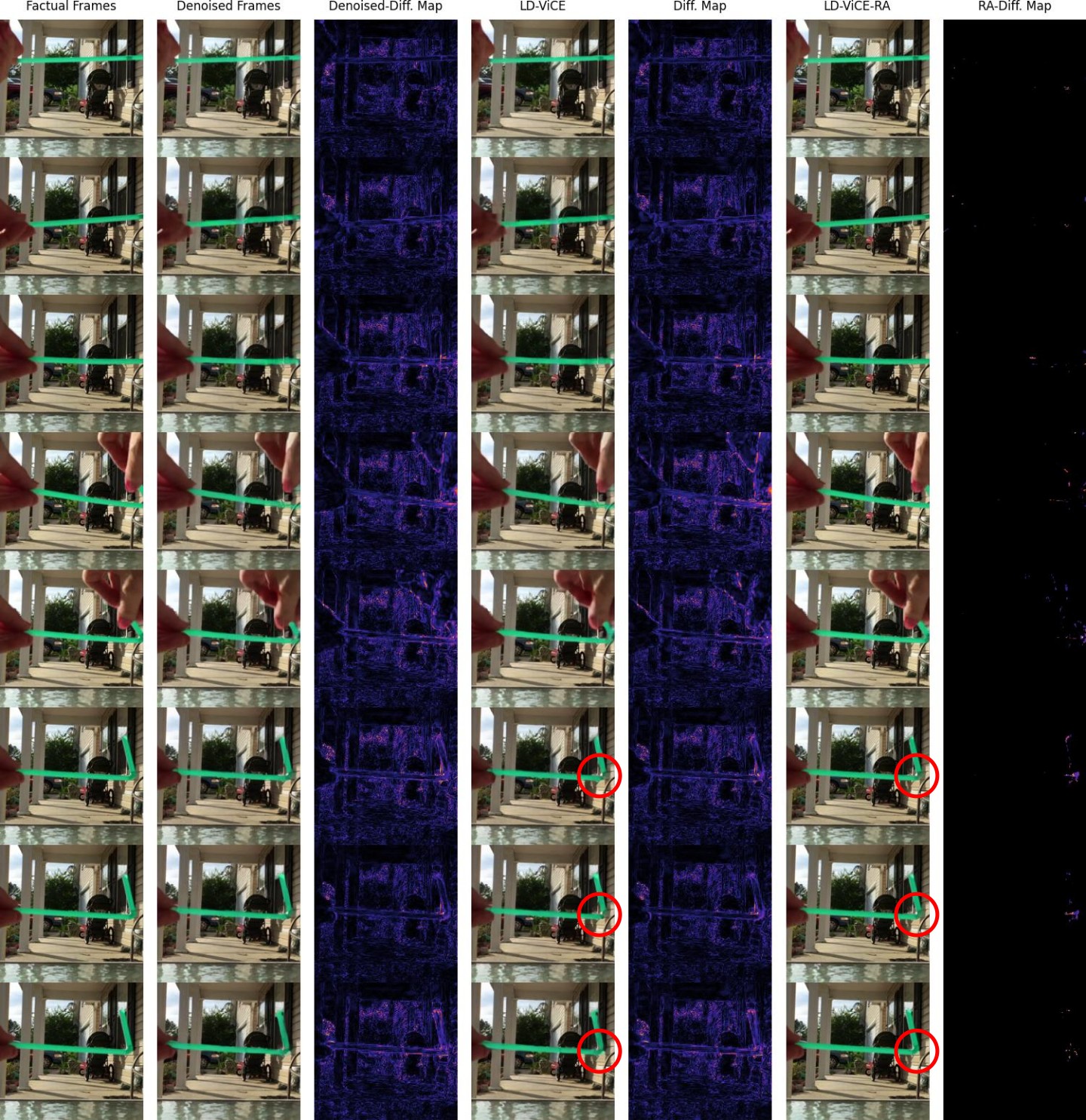} 
    \caption{
            Qualitative counterfactual results for transforming the predicted action from \textit{Bending something so that it deforms} to \textit{Bending something until it breaks} on the SSv2 dataset, using \frameworkAcronym{}.
            Factual frames and denoised reconstructions (guidance-free) are shown alongside their denoised difference maps (Denoised–Diff.\ Map).
            \frameworkAcronym{} counterfactuals introduce classifier-driven changes in the interaction region, while the RA variant suppresses high-frequency diffusion artifacts.
            Modifications are concentrated in the expected interaction region, and in the final frames, a visible break emerges in the middle of the straw (highlighted in the red circle).
            }
    \label{fig:ssv2_video_2942_cmp}
\end{figure*}

The Something‑Something V2 (SSv2) dataset is a highly complex benchmark for counterfactual generation.
Its large number of action classes spans a wide range of object types, interactions, and motion patterns, requiring substantial and perceptually consistent changes across multiple frames to transform one class into another.
Generating such counterfactuals is challenging, as the modifications must convincingly alter spatial appearance and temporal dynamics to reflect the target class.
For this study, only qualitative evaluation is performed by selecting representative source–target class pairs to examine the ability of \frameworkAcronym{} to produce semantically meaningful counterfactuals.
\cref{fig:ssv2_video_730_cmp,fig:ssv2_video_2942_cmp} show transformations from source class \textit{Bending something so that it deforms} to target class \textit{Bending something until it breaks}.
The factual frames are first shown alongside denoised reconstructions obtained through guidance-free denoising.
The corresponding difference maps (Denoised–Difference Map) capture changes introduced purely by the diffusion process, revealing minor structural variations unrelated to classifier guidance.
The subsequent \frameworkAcronym{} counterfactuals show targeted modifications in the interaction region; however, the model does not reliably introduce the visual signature of an actual break event, failing to form the distinct deformation or separation typical of breakage.
The RA variant reduces high-frequency denoising artifacts and produces more stable and spatially localized edits, as seen in the difference maps (RA–Difference Map), but the semantic cue of “breaking” remains weak.

\begin{figure*}[h]
    \centering
    \includegraphics[width=\linewidth]{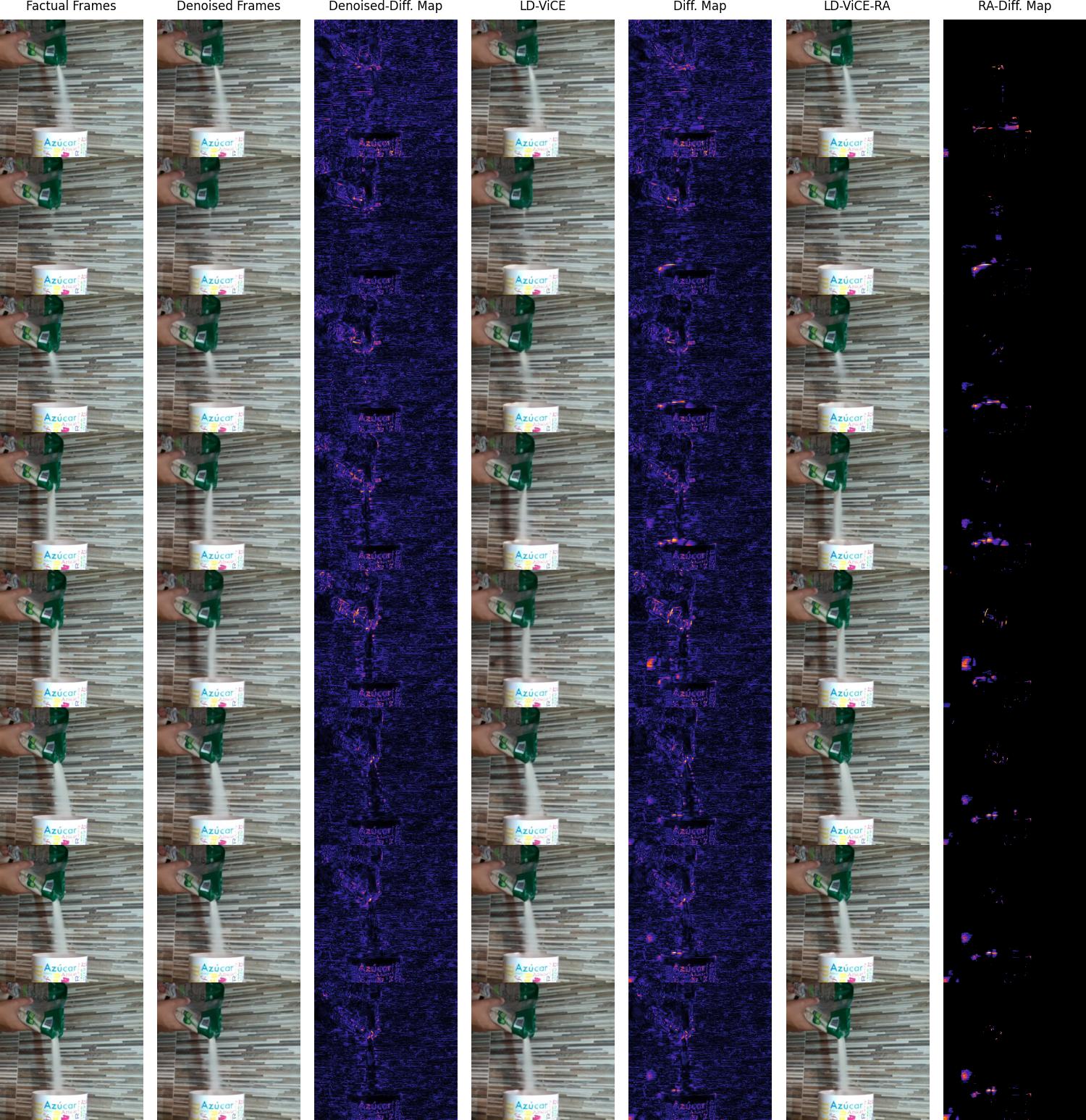} 
    \caption{
        Qualitative counterfactual results for transforming the predicted action from \textit{Pouring something into something} to \textit{Pouring something into something until it overflows} on the SSv2 dataset, using \frameworkAcronym{}.
        Factual frames and denoised reconstructions (guidance-free) are shown alongside their denoised difference maps (Denoised–Diff.\ Map).
        \frameworkAcronym{} counterfactuals introduce classifier-driven changes in the interaction region, while the RA variant suppresses high-frequency diffusion artifacts.
        Modifications occur near the pouring region, creating the effect of a filled container without clear visual evidence of overflowing liquid.
        }
    \label{fig:ssv2_video_7_cmp}
\end{figure*}

\begin{figure*}[h]
    \centering
    \includegraphics[width=\linewidth]{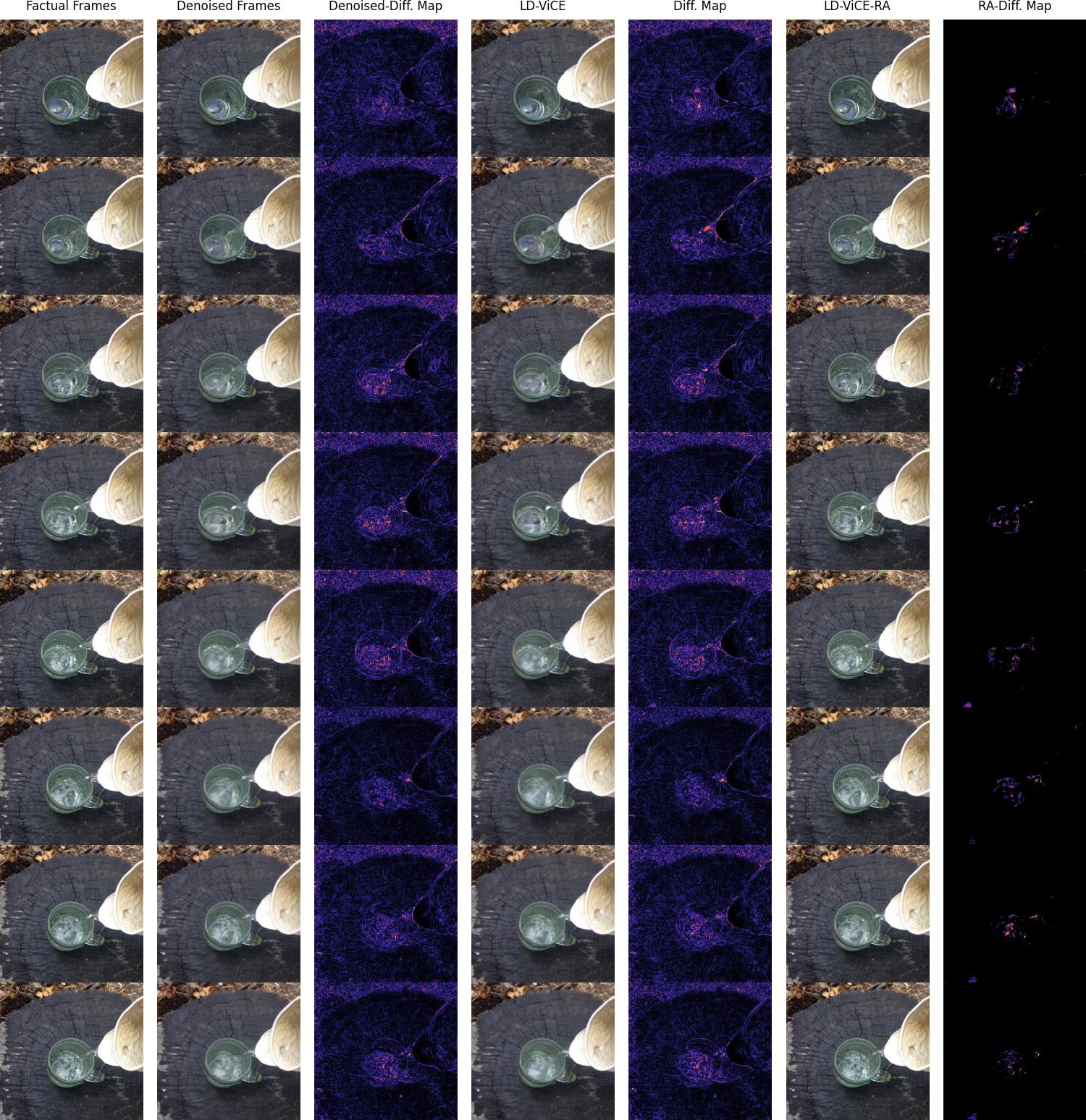} 
    \caption{
            Qualitative counterfactual results for transforming the predicted action from \textit{Pouring something into something} to \textit{Pouring something into something until it overflows} on the SSv2 dataset, using \frameworkAcronym{}.
            Factual frames and denoised reconstructions (guidance-free) are shown alongside their denoised difference maps (Denoised–Diff.\ Map).
            \frameworkAcronym{} counterfactuals introduce classifier-driven changes in the interaction region, while the RA variant suppresses high-frequency diffusion artifacts.
            Modifications occur near the pouring region, creating the effect of a filled container without clear visual evidence of overflowing liquid.
            }
    \label{fig:ssv2_video_75_cmp}
\end{figure*}

Similarly, \cref{fig:ssv2_video_7_cmp,fig:ssv2_video_75_cmp} illustrate transformations from source class \textit{Pouring something into something} to target class \textit{Pouring something into something until it overflows}.
While modifications on the upper surface of the container create the appearance of increasing volume, the generated videos lack any explicit depiction of spillage or overflow. As a result, the counterfactual only partially reflects the semantic requirements of the target class, indicating a limitation in capturing the full class-specific visual evidence.

However, the generated counterfactuals, while frequently classified as the target class with high confidence, do not always convey semantically faithful visual evidence for the intended action change.
This mismatch suggests that the model may rely on dataset biases or non‑discriminative background/context cues to trigger the classifier, rather than learning robust, class‑specific visual–temporal features.
The high diversity of SSv2 classes, variation in object appearance, and complex temporal dependencies contribute to this difficulty.

%%%%%%%%%%%%%%%%%%%%%%%%%%%%%%%%%%%%%%%%%%%%%%%%%%%%%%%%%%%%%%%%%%%%%%%%%%%%%%%%%%%%%%%%%%%%%%%%%%%%%%%%%%%%%%%%%%%%%%%%%%%%%%%%%%%%%%%%%%%%%%%%%%%%%%%%%%%%%%%%%%%%%%%%%%%%%%%%%%%%%%%%%%%%%%%%%%%%%%%%%%%%%%%%%%%%%%%%%%%%%%%%%%%%%%%%%%%%%%%%%%%%%%%%%%%%%%%%%%%%%%%%%%%%%%%%%%%%%%

\clearpage
{
    \small
    \bibliographystyle{ieeenat_fullname}
    \bibliography{main}

@inproceedings{ceylan2023pix2video,
  title={Pix2video: Video editing using image diffusion},
  author={Ceylan, Duygu and Huang, Chun-Hao P and Mitra, Niloy J},
  booktitle={Proceedings of the IEEE/CVF International Conference on Computer Vision},
  pages={23206--23217},
  year={2023}
}

@inproceedings{esser2023structure,
  title={Structure and content-guided video synthesis with diffusion models},
  author={Esser, Patrick and Chiu, Johnathan and Atighehchian, Parmida and Granskog, Jonathan and Germanidis, Anastasis},
  booktitle={Proceedings of the IEEE/CVF international conference on computer vision},
  pages={7346--7356},
  year={2023}
}

@article{unterthiner2018towards,
  title={Towards accurate generative models of video: A new metric \& challenges},
  author={Unterthiner, Thomas and Van Steenkiste, Sjoerd and Kurach, Karol and Marinier, Raphael and Michalski, Marcin and Gelly, Sylvain},
  journal={arXiv preprint arXiv:1812.01717},
  year={2018}
}

@inproceedings{reynaud2023feature,
  title={Feature-conditioned cascaded video diffusion models for precise echocardiogram synthesis},
  author={Reynaud, Hadrien and Qiao, Mengyun and Dombrowski, Mischa and Day, Thomas and Razavi, Reza and Gomez, Alberto and Leeson, Paul and Kainz, Bernhard},
  booktitle={International Conference on Medical Image Computing and Computer-Assisted Intervention},
  pages={142--152},
  year={2023},
  organization={Springer}
}

@inproceedings{radford2021learning,
  title={Learning transferable visual models from natural language supervision},
  author={Radford, Alec and Kim, Jong Wook and Hallacy, Chris and Ramesh, Aditya and Goh, Gabriel and Agarwal, Sandhini and Sastry, Girish and Askell, Amanda and Mishkin, Pamela and Clark, Jack and others},
  booktitle={International conference on machine learning},
  pages={8748--8763},
  year={2021},
  organization={PmLR}
}

@article{ouyang2020video,
  title={Video-based AI for beat-to-beat assessment of cardiac function},
  author={Ouyang, David and He, Bryan and Ghorbani, Amirata and Yuan, Neal and Ebinger, Joseph and Langlotz, Curtis P and Heidenreich, Paul A and Harrington, Robert A and Liang, David H and Ashley, Euan A and others},
  journal={Nature},
  volume={580},
  number={7802},
  pages={252--256},
  year={2020},
  publisher={Nature Publishing Group}
}

@inproceedings{goyal2017something,
  title={The" something something" video database for learning and evaluating visual common sense},
  author={Goyal, Raghav and Ebrahimi Kahou, Samira and Michalski, Vincent and Materzynska, Joanna and Westphal, Susanne and Kim, Heuna and Haenel, Valentin and Fruend, Ingo and Yianilos, Peter and Mueller-Freitag, Moritz and others},
  booktitle={Proceedings of the IEEE international conference on computer vision},
  pages={5842--5850},
  year={2017}
}

@article{nagelkerke1991note,
  title={A note on a general definition of the coefficient of determination},
  author={Nagelkerke, Nico JD and others},
  journal={biometrika},
  volume={78},
  number={3},
  pages={691--692},
  year={1991},
  publisher={Oxford University Press}
}

@article{chicco2021coefficient,
  title={The coefficient of determination R-squared is more informative than SMAPE, MAE, MAPE, MSE and RMSE in regression analysis evaluation},
  author={Chicco, Davide and Warrens, Matthijs J and Jurman, Giuseppe},
  journal={Peerj computer science},
  volume={7},
  pages={e623},
  year={2021},
  publisher={PeerJ Inc.}
}

@article{wang2004image,
  title={Image quality assessment: from error visibility to structural similarity},
  author={Wang, Zhou and Bovik, Alan C and Sheikh, Hamid R and Simoncelli, Eero P},
  journal={IEEE transactions on image processing},
  volume={13},
  number={4},
  pages={600--612},
  year={2004},
  publisher={IEEE}
}

@inproceedings{zhang2018unreasonable,
  title={The unreasonable effectiveness of deep features as a perceptual metric},
  author={Zhang, Richard and Isola, Phillip and Efros, Alexei A and Shechtman, Eli and Wang, Oliver},
  booktitle={Proceedings of the IEEE conference on computer vision and pattern recognition},
  pages={586--595},
  year={2018}
}

@article{simonyan2014very,
  title={Very deep convolutional networks for large-scale image recognition},
  author={Simonyan, Karen and Zisserman, Andrew},
  journal={arXiv preprint arXiv:1409.1556},
  year={2014}
}

@article{heusel2017gans,
  title={Gans trained by a two time-scale update rule converge to a local nash equilibrium},
  author={Heusel, Martin and Ramsauer, Hubert and Unterthiner, Thomas and Nessler, Bernhard and Hochreiter, Sepp},
  journal={Advances in neural information processing systems},
  volume={30},
  year={2017}
}

@inproceedings{wang2022ferv39k,
  title={Ferv39k: A large-scale multi-scene dataset for facial expression recognition in videos},
  author={Wang, Yan and Sun, Yixuan and Huang, Yiwen and Liu, Zhongying and Gao, Shuyong and Zhang, Wei and Ge, Weifeng and Zhang, Wenqiang},
  booktitle={Proceedings of the IEEE/CVF conference on computer vision and pattern recognition},
  pages={20922--20931},
  year={2022}
}

@String(ICIP = {IEEE Int. Conf. Image Process.})

@String(AAAI = {AAAI})

@String(ICIP  = {ICIP})

@article{yurtsever2020survey,
  title={A survey of autonomous driving: Common practices and emerging technologies},
  author={Yurtsever, Ekim and Lambert, Jacob and Carballo, Alexander and Takeda, Kazuya},
  journal={IEEE access},
  volume={8},
  pages={58443--58469},
  year={2020},
  publisher={IEEE}
}

@article{ullah2023comprehensive,
  title={A comprehensive review on vision-based violence detection in surveillance videos},
  author={Ullah, Fath U Min and Obaidat, Mohammad S and Ullah, Amin and Muhammad, Khan and Hijji, Mohammad and Baik, Sung Wook},
  journal={ACM Computing Surveys},
  volume={55},
  number={10},
  pages={1--44},
  year={2023},
  publisher={ACM New York, NY}
}

@article{farid2023latent,
  title={Latent diffusion counterfactual explanations},
  author={Farid, Karim and Schrodi, Simon and Argus, Max and Brox, Thomas},
  journal={arXiv preprint arXiv:2310.06668},
  year={2023}
}

@article{farhad2023review,
  title={A review of medical diagnostic video analysis using deep learning techniques},
  author={Farhad, Moomal and Masud, Mohammad Mehedy and Beg, Azam and Ahmad, Amir and Ahmed, Luai},
  journal={Applied Sciences},
  volume={13},
  number={11},
  pages={6582},
  year={2023},
  publisher={MDPI}
}

@article{tang2025video,
  title={Video understanding with large language models: A survey},
  author={Tang, Yunlong and Bi, Jing and Xu, Siting and Song, Luchuan and Liang, Susan and Wang, Teng and Zhang, Daoan and An, Jie and Lin, Jingyang and Zhu, Rongyi and others},
  journal={IEEE Transactions on Circuits and Systems for Video Technology},
  year={2025},
  publisher={IEEE}
}

@article{tong2022videomae,
  title={Videomae: Masked autoencoders are data-efficient learners for self-supervised video pre-training},
  author={Tong, Zhan and Song, Yibing and Wang, Jue and Wang, Limin},
  journal={Advances in neural information processing systems},
  volume={35},
  pages={10078--10093},
  year={2022}
}

@article{arunnehru2023deep,
  title={Deep learning-based real-world object detection and improved anomaly detection for surveillance videos},
  author={Arunnehru, J and others},
  journal={Materials Today: Proceedings},
  volume={80},
  pages={2911--2916},
  year={2023},
  publisher={Elsevier}
}

@article{adadi2018peeking,
  title={Peeking inside the black-box: a survey on explainable artificial intelligence (XAI)},
  author={Adadi, Amina and Berrada, Mohammed},
  journal={IEEE access},
  volume={6},
  pages={52138--52160},
  year={2018},
  publisher={IEEE}
}

@article{arun2021assessing,
  title={Assessing the trustworthiness of saliency maps for localizing abnormalities in medical imaging},
  author={Arun, Nishanth and Gaw, Nathan and Singh, Praveer and Chang, Ken and Aggarwal, Mehak and Chen, Bryan and Hoebel, Katharina and Gupta, Sharut and Patel, Jay and Gidwani, Mishka and others},
  journal={Radiology: Artificial Intelligence},
  volume={3},
  number={6},
  pages={e200267},
  year={2021},
  publisher={Radiological Society of North America}
}

@article{adebayo2018sanity,
  title={Sanity checks for saliency maps},
  author={Adebayo, Julius and Gilmer, Justin and Muelly, Michael and Goodfellow, Ian and Hardt, Moritz and Kim, Been},
  journal={Advances in neural information processing systems},
  volume={31},
  year={2018}
}

@inproceedings{stergiou2019saliency,
  title={Saliency tubes: Visual explanations for spatio-temporal convolutions},
  author={Stergiou, Alexandros and Kapidis, Georgios and Kalliatakis, Grigorios and Chrysoulas, Christos and Veltkamp, Remco and Poppe, Ronald},
  booktitle={2019 IEEE international conference on image processing (ICIP)},
  pages={1830--1834},
  year={2019},
  organization={IEEE}
}

@inproceedings{li2021towards,
  title={Towards visually explaining video understanding networks with perturbation},
  author={Li, Zhenqiang and Wang, Weimin and Li, Zuoyue and Huang, Yifei and Sato, Yoichi},
  booktitle={Proceedings of the IEEE/CVF Winter Conference on Applications of Computer Vision},
  pages={1120--1129},
  year={2021}
}

@inproceedings{ji2023spatial,
  title={Spatial-temporal concept based explanation of 3d convnets},
  author={Ji, Ying and Wang, Yu and Kato, Jien},
  booktitle={Proceedings of the IEEE/CVF Conference on Computer Vision and Pattern Recognition},
  pages={15444--15453},
  year={2023}
}

@inproceedings{saha2024exploring,
  title={Exploring explainability in video action recognition},
  author={Saha, Avinab and Gupta, Shashank and Ankireddy, Sravan Kumar and Chahine, Karl and Ghosh, Joydeep},
  booktitle={Proceedings of the IEEE/CVF Conference on Computer Vision and Pattern Recognition},
  pages={8176--8181},
  year={2024}
}

@article{wachter2017counterfactual,
  title={Counterfactual explanations without opening the black box: Automated decisions and the GDPR},
  author={Wachter, Sandra and Mittelstadt, Brent and Russell, Chris},
  journal={Harv. JL \& Tech.},
  volume={31},
  pages={841},
  year={2017},
  publisher={HeinOnline}
}

@inproceedings{reynaud2022d,
  title={D’artagnan: Counterfactual video generation},
  author={Reynaud, Hadrien and Vlontzos, Athanasios and Dombrowski, Mischa and Gilligan Lee, Ciar{\'a}n and Beqiri, Arian and Leeson, Paul and Kainz, Bernhard},
  booktitle={International Conference on Medical Image Computing and Computer-Assisted Intervention},
  pages={599--609},
  year={2022},
  organization={Springer}
}

@inproceedings{zong2025text,
  title={Text-Guided Fine-grained Counterfactual Inference for Short Video Fake News Detection},
  author={Zong, Linlin and Lin, Wenmin and Zhou, Jiahui and Liu, Xinyue and Zhang, Xianchao and Xu, Bo and Wu, Shimin},
  booktitle={Proceedings of the AAAI Conference on Artificial Intelligence},
  volume={39},
  number={1},
  pages={1237--1245},
  year={2025}
}

@article{kolarik2023explainability,
  title={Explainability of deep learning models in medical video analysis: a survey},
  author={Kolarik, Michal and Sarnovsky, Martin and Paralic, Jan and Babic, Frantisek},
  journal={PeerJ Computer Science},
  volume={9},
  pages={e1253},
  year={2023},
  publisher={PeerJ Inc.}
}

@inproceedings{kanehira2019multimodal,
  title={Multimodal explanations by predicting counterfactuality in videos},
  author={Kanehira, Atsushi and Takemoto, Kentaro and Inayoshi, Sho and Harada, Tatsuya},
  booktitle={Proceedings of the IEEE/CVF Conference on Computer Vision and Pattern Recognition},
  pages={8594--8602},
  year={2019}
}

@inproceedings{jeanneret2022diffusion,
  title={Diffusion models for counterfactual explanations},
  author={Jeanneret, Guillaume and Simon, Lo{\"\i}c and Jurie, Fr{\'e}d{\'e}ric},
  booktitle={Proceedings of the Asian conference on computer vision},
  pages={858--876},
  year={2022}
}

@article{augustin2022diffusion,
  title={Diffusion visual counterfactual explanations},
  author={Augustin, Maximilian and Boreiko, Valentyn and Croce, Francesco and Hein, Matthias},
  journal={Advances in Neural Information Processing Systems},
  volume={35},
  pages={364--377},
  year={2022}
}

@inproceedings{varshney2024generating,
  title={Generating counterfactual trajectories with latent diffusion models for concept discovery},
  author={Varshney, Payal and Lucieri, Adriano and Balada, Christoph and Dengel, Andreas and Ahmed, Sheraz},
  booktitle={International Conference on Pattern Recognition},
  pages={138--153},
  year={2024},
  organization={Springer}
}

@article{song2020denoising,
  title={Denoising diffusion implicit models},
  author={Song, Jiaming and Meng, Chenlin and Ermon, Stefano},
  journal={arXiv preprint arXiv:2010.02502},
  year={2020}
}

@article{chen2024static,
  title={From static to dynamic: Adapting landmark-aware image models for facial expression recognition in videos},
  author={Chen, Yin and Li, Jia and Shan, Shiguang and Wang, Meng and Hong, Richang},
  journal={IEEE Transactions on Affective Computing},
  year={2024},
  publisher={IEEE}
}

@article{yang2024cogvideox,
  title={Cogvideox: Text-to-video diffusion models with an expert transformer},
  author={Yang, Zhuoyi and Teng, Jiayan and Zheng, Wendi and Ding, Ming and Huang, Shiyu and Xu, Jiazheng and Yang, Yuanming and Hong, Wenyi and Zhang, Xiaohan and Feng, Guanyu and others},
  journal={arXiv preprint arXiv:2408.06072},
  year={2024}
}

@article{hiley2020explaining,
  title={Explaining motion relevance for activity recognition in video deep learning models},
  author={Hiley, Liam and Preece, Alun and Hicks, Yulia and Chakraborty, Supriyo and Gurram, Prudhvi and Tomsett, Richard},
  journal={arXiv preprint arXiv:2003.14285},
  year={2020}
}
}

% 
% % WARNING: do not forget to delete the supplementary pages from your submission 

% {   \small
%     \bibliographystyle{ieeenat_fullname}
%     \bibliography{appendix}
% }

\end{document}